\documentclass[letterpaper, 10pt, journal, twoside]{cls/IEEEtran}

\IEEEoverridecommandlockouts  

\usepackage{amsmath,amssymb,amsfonts,amsthm,amscd,dsfont} 
\usepackage{bbm}
\usepackage{mathtools}
\usepackage{accents} 

\usepackage[utf8]{inputenc}
\usepackage[T1]{fontenc}
\usepackage{xspace}
\usepackage{multirow}
\usepackage{hhline}

\usepackage{algorithm,listings}
\usepackage[noend]{algpseudocode}
\usepackage[table,usenames,dvipsnames]{xcolor}
\usepackage{graphicx,tabularx,adjustbox}
\usepackage[font={small}]{caption}
\usepackage{subcaption}
\usepackage{paralist}
\usepackage[inline]{enumitem}
\usepackage{cite}
\usepackage[normalem]{ulem}  

\usepackage[bookmarks=true, colorlinks, breaklinks=true]{hyperref}
\usepackage{pbox}
\usepackage{makecell}

\usepackage[titletoc,title]{appendix}
\usepackage{balance}

\DeclareMathOperator*{\diag}{diag}

\newcommand{\tigtto}{\mathrel{\mkern-5mu\to\mkern-5mu}}

\theoremstyle{definition}

\theoremstyle{remark}

\newcommand{\bfc}{\mathbf{c}}

\newcommand{\bfg}{\mathbf{g}}

\newcommand{\bfk}{\mathbf{k}}

\newcommand{\bfn}{\mathbf{n}}
\newcommand{\bfo}{\mathbf{o}}
\newcommand{\bfp}{\mathbf{p}}
\newcommand{\bfq}{\mathbf{q}}

\newcommand{\bfs}{\mathbf{s}}

\newcommand{\bfv}{\mathbf{v}}
\newcommand{\bfw}{\mathbf{w}}
\newcommand{\bfx}{\mathbf{x}}
\newcommand{\bfy}{\mathbf{y}}
\newcommand{\bfz}{\mathbf{z}}

\newcommand{\bfmu}{\boldsymbol{\mu}}

\newcommand{\bfphi}{\boldsymbol{\phi}}

\newcommand{\bfG}{\mathbf{G}}

\newcommand{\bfI}{\mathbf{I}}

\newcommand{\bfK}{\mathbf{K}}

\newcommand{\bfR}{\mathbf{R}}

\newcommand{\bfV}{\mathbf{V}}

\newcommand{\bfX}{\mathbf{X}}

\newcommand{\bfSigma}{\boldsymbol{\Sigma}}

\newcommand{\bbE}{\mathbb{E}}

\newcommand{\bbR}{\mathbb{R}}

\newcommand{\bbV}{\mathbb{V}}

\newcommand{\calB}{\mathcal{B}}

\newcommand{\calD}{\mathcal{D}}

\newcommand{\calL}{\mathcal{L}}

\newcommand{\calN}{\mathcal{N}}
\newcommand{\calO}{\mathcal{O}}
\newcommand{\calP}{\mathcal{P}}
\newcommand{\calQ}{\mathcal{Q}}

\usepackage{bibunits}
\defaultbibliographystyle{cls/IEEEtran}
\defaultbibliography{bib/main.bib}

\newcommand{\edit}[1]{{#1}}
\newcommand{\NA}[1]{}
\newcommand{\ZD}[1]{}
\newcommand{\BL}[1]{}
\newcommand{\JM}[1]{}
\newcommand{\MA}[2][]{}

\newcommand{\mainref}[1]{\ref{#1}}
\newcommand{\maineqref}[1]{\eqref{#1}}

\def\methodname{Kernel-SDF\xspace}
\title{\bf \methodname: An Open-Source Library for Real-Time Signed Distance Function Estimation using Kernel Regression}

\newcommand{\linebreakand}{%
\end{@IEEEauthorhalign}
\hfill\mbox{}\par
\mbox{}\hfill
\begin{@IEEEauthorhalign}
}

\ifdefined\anonymous
\author{Anonymous Authors}
\else
\author{
    Zhirui Dai\textsuperscript{1,*} \qquad Tianxing Fan\textsuperscript{1,*} \qquad Mani Amani\textsuperscript{1} \qquad Jaemin Seo\textsuperscript{2} \\
    Ki Myung Brian Lee\textsuperscript{1} \qquad Hyondong Oh\textsuperscript{2} \qquad Nikolay Atanasov\textsuperscript{1}%
    \thanks{Manuscript received: March 30, 2026; Revised June 17, 2026; Accepted July 6, 2026. This paper was recommended for publication by Editor Ayoung Kim upon evaluation of the Associate Editor and Reviewers' comments.}%
    \thanks{This work was supported by ARL DCIST CRA W911NF-17-2-0181, NSF FRR CAREER 2045945, the Ministry of Trade, Industry, and Energy (MOTIE), Korea, under the Strategic Technology Development Program supervised by the Korea Institute for Advancement of Technology (KIAT) [Grant No. P0026052], and Korea Institute of Planning and Evaluation for Technology in Food, Agriculture, Forestry (IPET) through (Smart Farm Innovation Technology Development Program), funded by Ministry of Agriculture, Food and Rural Affairs (MAFRA) (RS-2025-02219411).}%
    \thanks{\textsuperscript{1}Department of Electrical and Computer Engineering, University of California San Diego, La Jolla, CA 92093, USA. E-mails: \{\mbox{zhdai},t2fan,mamai5250,kmblee,natanasov\}@ucsd.edu.}%
    \thanks{\textsuperscript{2}Department of Mechanical Engineering, Korea Advanced Institute of Science and Technology (KAIST), Daejeon 34051, Republic of Korea. E-mails: \{jam.seo,h.oh\}@kaist.ac.kr.}%
    \thanks{\textsuperscript{*}Equal contribution. Code and additional results: \url{https://github.com/ExistentialRobotics/kernel_sdf}.}
}
\fi

\begin{document}

\markboth{IEEE Robotics and Automation Letters. Preprint Version. Accepted July, 2026}%
{Dai \MakeLowercase{\textit{et al.}}: \methodname: Real-Time Signed Distance Function Estimation using Kernel Regression}

\begin{bibunit}
\nocite{IEEEexample:BSTcontrol}

\maketitle

\begin{abstract}
Accurate and efficient scene representation is crucial for robotic tasks such as motion planning, manipulation, and navigation. Signed distance functions (SDFs) have emerged as a powerful representation for encoding distance to obstacle boundaries, enabling efficient collision-checking and trajectory optimization.
However, existing methods are limited for large-scale uncertainty-aware SDF estimation from streaming sensor data: voxel-based approaches have fixed resolution and lack uncertainty quantification, neural network methods require significant training time, and Gaussian process (GP) methods struggle with scalability, sign estimation, and uncertainty calibration.
In this letter, we develop an open-source library, \methodname, using kernel regression to learn SDF with calibrated uncertainty in real-time. It combines a front-end learning a continuous occupancy field via kernel regression with a back-end that estimates accurate SDF via GP regression using samples from the front-end surface boundaries.
\methodname provides accurate SDF, gradient, uncertainty, and mesh construction in real-time. Evaluations show it achieves superior accuracy over existing methods while maintaining real-time performance, making it suitable for robotics tasks requiring reliable uncertainty-aware geometry.
\end{abstract}


\def\allSdfWorks{
newcombe_kinectfusion_2011,oleynikova_voxblox_2017,park_deepsdf_2019,han_fiesta_2019,lee2019gpis,lan2021loggpis,wang_neus_2021,vizzo_vdbfusion_2022,pan_voxfield_2022,ortiz_isdf_2022,wang_neus2_2023,jiang_h2-mapping_2023,pan_pin-slam_2024,zou_gmm_gp_2024,wu_vdb-gpdf_2025,miso2025}
\def\voxelSdfWorks{
newcombe_kinectfusion_2011,oleynikova_voxblox_2017,han_fiesta_2019,vizzo_vdbfusion_2022}
\def\vdbSdfWorks{
	vizzo_vdbfusion_2022,wu_vdb-gpdf_2025}
\def\neuralSdfWorks{
	park_deepsdf_2019,ortiz_isdf_2022,wang_neus2_2023,jiang_h2-mapping_2023,pan_pin-slam_2024,miso2025}
\def\gpSdfWorks{
	lee2019gpis,lan2021loggpis,zou_gmm_gp_2024,wu_vdb-gpdf_2025}

\section{Introduction}
\label{sec:introduction}

\IEEEPARstart{G}{eometric} scene representations play a key role for many robot autonomy tasks, including mapping, localization, motion planning, manipulation, and navigation. An ideal representation should meet several requirements:
\begin{enumerate*}[label=\arabic*)]
	\item accuracy to reflect the scene structure,
	\item efficiency to allow real-time processing of streaming sensor data,
	\item scalability to large scenes,
	\item robustness to noise,
	\item uncertainty quantification to support risk-aware decision making,
	\item differentiability to support gradient-based optimization,
	\item adaptability to dynamic scenes, and
	\item compatibility with different sensors.
\end{enumerate*}

Many effective representations have been proposed: occupancy maps \cite{hornung13octomap}, point clouds \cite{surfelslam2018}, meshes \cite{kimera2020}, and implicit functions such as SDFs \cite{\allSdfWorks} and neural radiance fields (NeRFs) \cite{nerf2020}.
Among these, SDFs have attracted attention for providing the distance to the nearest obstacles, useful for mapping \cite{newcombe_kinectfusion_2011, miso2025} and motion planning \cite{lee_safe_2024}. Moreover, differentiable SDF methods can define safety constraints in navigation \cite{long_sensor-based_2025} and manipulation \cite{li_manipulation_2024}.

\begin{figure}
	\centering
    \begin{subfigure}[t]{0.48\linewidth}
        \centering
        \includegraphics[width=\linewidth,trim={0pt 40pt 0pt 70pt},clip]{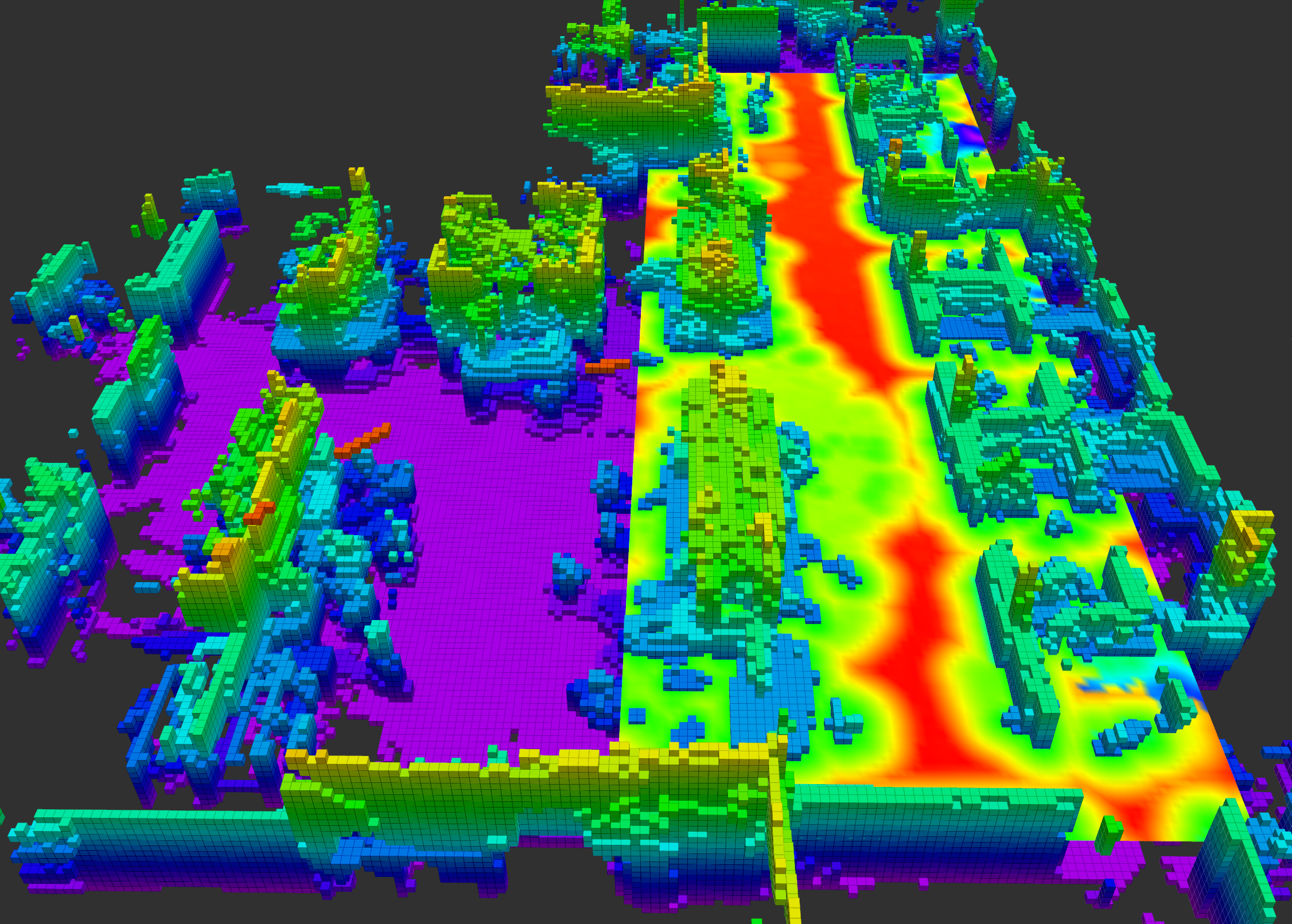}
        \caption{Signed distance function}
        \label{fig:teaser_sdf_vis}
    \end{subfigure}%
    \hfill%
    \begin{subfigure}[t]{0.48\linewidth}
        \centering
        \includegraphics[width=\linewidth,trim={0pt 0pt 0pt 40pt},clip]{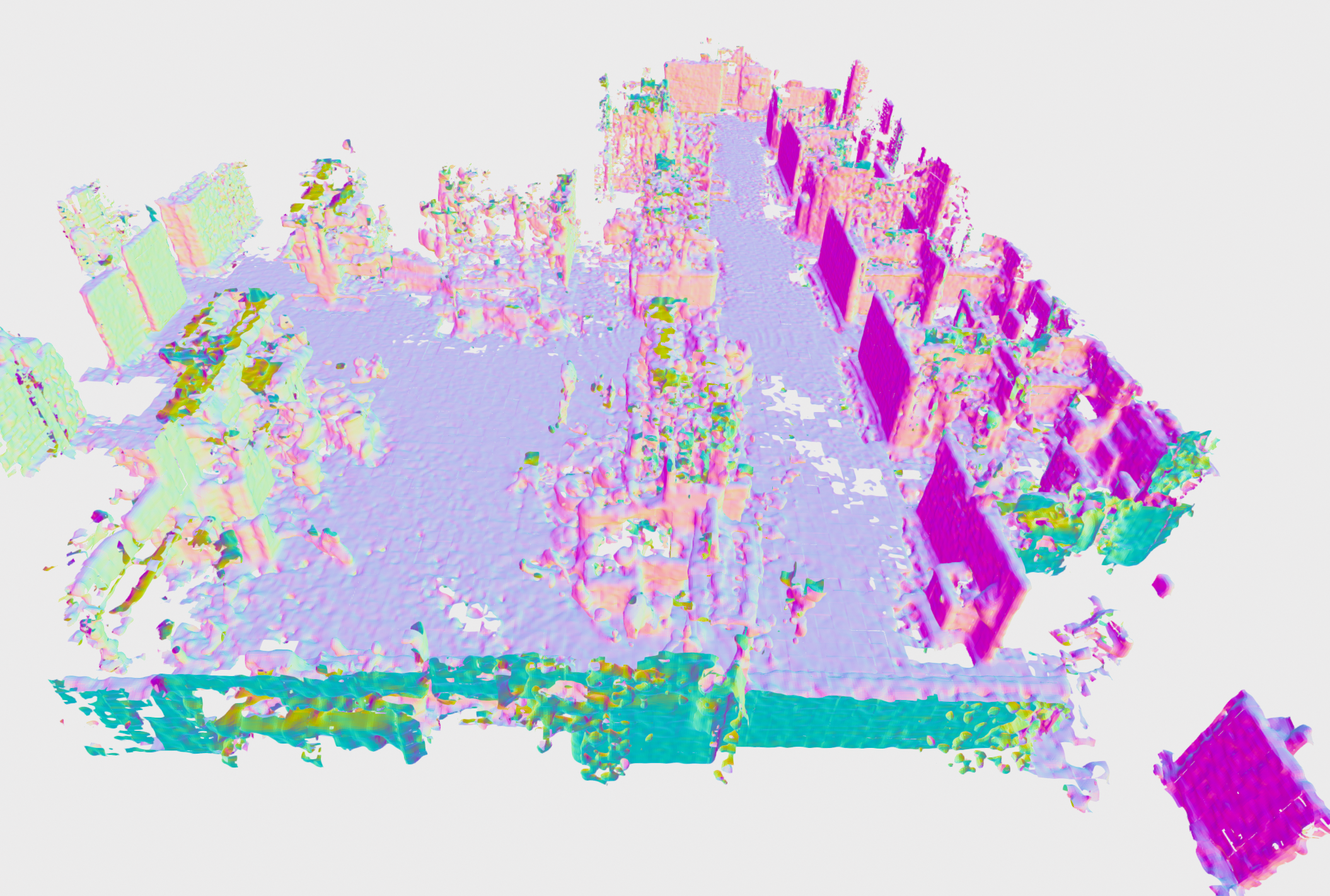}
        \caption{Reconstructed mesh}
        \label{fig:teaser_mesh}
    \end{subfigure}
	\caption{Example of SDF estimation, visualized as a heatmap in (a), and mesh reconstruction in (b) of a large-scale environment, obtained using \methodname.}
	\label{fig:teaser}
    \vspace{-0.5em}
\end{figure}

Various methods have been proposed to learn SDFs from range sensors such as LiDAR or depth cameras. Voxel-based methods \cite{\voxelSdfWorks} often focus only on truncated SDF (TSDF) estimation, though some \cite{oleynikova_voxblox_2017,han_fiesta_2019,bai_vdbblox_2023,millane_nvblox_2024} reconstruct non-truncated (Euclidean) SDF on a discrete grid.
These methods maintain a dense, fixed-resolution approximation of (T)SDF, which trades off accuracy against efficiency and prevents differentiability.
In contrast, neural network methods \cite{\neuralSdfWorks} learn SDF as a continuous function, providing accurate and differentiable approximations without an explicit grid.
However, the significant training time they require to converge makes them unsuitable for online learning.

Gaussian process (GP) methods \cite{\gpSdfWorks} are non-parametric and can estimate accurate Euclidean SDF, but scale poorly with environment size, lack robust sign estimation, and report variance inconsistent with actual SDF errors.
For scalability, octree \cite{lan2021loggpis} and OpenVDB \cite{wu_vdb-gpdf_2025} data structures split the training data to train multiple GPs.
For consistent sign and variance, GMMGP \cite{zou_gmm_gp_2024} uses an SDF prior from a hierarchical Gaussian mixture model, but this prior yields increasingly large errors away from the surface.

In this letter, we develop an open-source library, \methodname, for SDF estimation that meets the aforementioned requirements and addresses the limitations of existing GP methods. \methodname maintains an octree to split the estimation into smaller areas, estimates the occupancy/sign in each octant using a Bayesian Hilbert map (BHM) \cite{bhm2017}, and reconstructs SDF via GP regression from the BHM surface samples. The library is thus organized as a surface-estimation front-end performing BHM occupancy estimation and an SDF-prediction back-end performing GP regression. The front-end learns a continuous occupancy field for robust sign estimation and provides surface point samples via the marching cubes algorithm \cite{lorensen_marching_1987}. The back-end uses these samples to train SDF GPs in each octant with integrated uncertainty quantification. We introduce several optimizations for real-time performance and map consistency: spatial partitioning and priority-queue-based updates for efficient computation, and BHM weight synchronization for consistency across neighboring local maps.

\methodname provides accurate SDF, gradient, uncertainty, and mesh construction in real time (see Fig.~\ref{fig:teaser}). In summary, our contributions include:
\begin{enumerate}
	\item an open-source C++ implementation for 2D/3D SDF learning, with Python and ROS1/ROS2 interfaces;


    \item a highly-optimized kernel regression method for real-time SDF learning via two components: a BHM front-end for continuous occupancy learning and uncertainty-aware surface extraction, and a GP back-end for SDF and gradient prediction with uncertainty;

    \item five strategies for efficiency and robustness: octree-based partitioning for local region learning, optimized data processing for rapid training set generation, weight synchronization for map consistency, adaptive scaling for GP stability, and priority-queue-based updates to balance accuracy and responsiveness.
\end{enumerate}

\section{Related Work}
\label{sec:related_work}

SDF learning methods can be categorized as voxel-based, neural-net-based, and Gaussian-process-based.

\vspace{-1.25em}
\subsection{Voxel Methods}
\label{sec:related_work:voxel_based}

Many methods learn SDF using a grid \cite{curless_volumetric_1996}.
KinectFusion \cite{newcombe_kinectfusion_2011} pioneered real-time 3D reconstruction from depth sensors based on projective TSDF.
Follow-up works such as Voxblox \cite{oleynikova_voxblox_2017} extended this to Euclidean SDF via breadth-first-search (BFS) integration from the occupied voxels. However, projective TSDF inaccurately estimates the true SDF as distance along camera rays, and BFS integration adds further errors.
FIESTA \cite{han_fiesta_2019} integrates SDF estimation with an occupancy map, replacing the BFS-based distance with the distance to the nearest occupied voxel.
VoxField \cite{pan_voxfield_2022} formulates non-projective TSDF, though its accuracy relies on estimated surface normals and SDF gradients.
A common limitation is the use of fixed-resolution grids, which restricts scalability in larger environments.
VDBblox \cite{bai_vdbblox_2023} and nvblox \cite{millane_nvblox_2024} address this via OpenVDB \cite{museth_vdb_2013} or GPU-based voxel hashing.
However, voxel-based methods lack uncertainty quantification and are usually not differentiable.

\vspace{-1.25em}
\subsection{Neural Network Methods}

Neural network methods are differentiable by design and can achieve highly accurate SDF reconstruction. DeepSDF \cite{park_deepsdf_2019} uses an MLP with latent features to learn the SDF of objects.
NeuS \cite{wang_neus_2021} and NeuS2 \cite{wang_neus2_2023} combine SDF with NeRFs \cite{nerf2020} to jointly learn SDF and rendering. PIN-SLAM \cite{pan_pin-slam_2024} and MISO \cite{miso2025} perform simultaneous localization and mapping (SLAM) on an SDF map encoded with neural features. $\text{H}_2$-Mapping \cite{jiang_h2-mapping_2023} co-optimizes a voxel SDF prior and a neural TSDF residual with volumetric rendering.
However, most neural network methods cannot meet real-time requirements in novel environments due to their significant training time. Some works, like $\text{H}_2$-Mapping \cite{jiang_h2-mapping_2023} and PIN-SLAM \cite{pan_pin-slam_2024}, learn neural SDF online but sacrifice substantial accuracy.

\vspace{-1.25em}
\subsection{Gaussian Process Methods}
\label{sec:related_work:gp_based}

GPs are suitable for online SDF learning but face the high computation cost of matrix inversion as the environment scale increases.
To mitigate this, GPIS \cite{lee2019gpis} employs octree-based data partitioning and trains a GP per local partition. However, its SDF estimates quickly revert to the zero-mean prior away from the surface.
Log-GPIS \cite{lan2021loggpis} exploits a connection between the heat kernel and the unsigned distance function (UDF) to formulate UDF GP regression in log space.
However, the log transformation ignores sign information and distorts variance estimation, causing uncertainty to explode far from the surface. Its surface estimation is also sensitive to sensor noise and dynamic changes.
VDB-GPDF \cite{wu_vdb-gpdf_2025} improves surface estimation through OpenVDB TSDF fusion but provides no sign estimates or reliable uncertainty quantification.
To overcome these limitations, \methodname uses a continuous occupancy field front-end based on BHM \cite{bhm2017}, which provides robust sign prediction and accurate surface estimation even in dynamic or noisy environments. Our GP back-end then learns SDF from the surface samples and formulates an uncertainty estimate that respects the SDF errors, delivering precise SDF and gradient predictions with reliable uncertainty quantification.

\section{PROBLEM STATEMENT}
\label{sec:problem_statement}

Consider a robot equipped with a depth camera or LiDAR sensor operating in an $n$-dimensional environment ($n=2,3$) with occupied space $\calO \subset \bbR^n$. The sensor provides point cloud measurements $\calP_t \subset \bbR^n$ of $\calO$ from known (or estimated) sensor positions $\bfp_t \in \bbR^n$ and orientations $\bfR_t \in SO(n)$ at time steps $t$. Using these data, our objective is to learn an SDF representation of the occupied space $\calO$:
\begin{equation} \label{eq:sdf_definition}
	d(\bfx) =
	\begin{cases}
		\phantom{+}\min_{\bfy \in \partial \calO} \left\|\bfx - \bfy\right\|_2,  & \bfx \not\in \calO, \\
		-\min_{\bfy \in \partial \calO} \left\|\bfx - \bfy\right\|_2, & \bfx \in \calO,
	\end{cases}
\end{equation}
where $\bfx \in \bbR^n$ is a query point and $\bfy$ is a surface point.
The SDF $d(\bfx)$ encodes the surface $\partial\calO$ as its zero-level set. Further, the gradient $\nabla d(\bfx)$ has unit norm whenever differentiable:
\begin{equation}
	d(\bfx) = 0,\ \forall \bfx \in \partial\calO\ \text{and}\ \left\|\nabla d(\bfx)\right\|_2=1, \ \text{a.e. }\bfx \in \calO. \label{eq:sdf_properties}
\end{equation}
We focus on estimating $d(\bfx)$ using the streaming point cloud measurements $\calP_t$ and the associated sensor poses $(\bfp_t, \bfR_t)$.

\section{\methodname}
\label{sec:kernel_sdf}

\methodname consists of a surface-estimation front-end using BHM \cite{bhm2017} (Sec.~\ref{sec:frontend}) and an SDF-prediction back-end using GP regression \cite{williams2006gaussian} (Sec.~\ref{sec:backend}).
Both are kernel regression methods supporting continuous representations with uncertainty quantification.
The BHM front-end fuses free and occupied measurements into a continuous occupancy field, which provides sign and per-surface-point uncertainty information to the SDF back-end. The SDF back-end provides a differentiable SDF with closed-form gradients and calibrated variance.
However, applying kernel regression over the entire environment is computationally expensive and impractical for real-time use.
\methodname addresses this by partitioning the environment into local regions via an octree and applying kernel regression within each region with additional optimizations (Fig.~\ref{fig:method_overview}).

\vspace{-1em}
\subsection{Surface Estimation Front-End}
\label{sec:frontend}

In this section, we review Bayesian Hilbert map (BHM) \cite{bhm2017
} and describe our extensions to achieve efficient updates, sign prediction, and surface point extraction.

1) {\bf Review of Bayesian Hilbert Map}:
BHM learns a continuous log-odds field $l(\bfx)$ using kernel regression in Hilbert space. The log-odds $l(\bfx)$ at position $\bfx$ is the ratio between the probabilities of being occupied and being free. BHM uses linear regression to learn the log-odds field, $l(\bfx)=\bfw^\top\bfphi(\bfx)$, with weights $\bfw$ and features $\bfphi(\bfx)$. To obtain the features, a query $\bfx$ is mapped into Hilbert space using a radial basis function (RBF) kernel $k(\cdot, \cdot)$ evaluated with respect to a set of hinged points $\{\tilde{\bfx}_i\}_{i=1}^{M}$ organized in a grid structure:
\begin{equation}
    \bfphi(\bfx) = [k(\bfx,\tilde{\bfx}_1), \cdots, k(\bfx, \tilde{\bfx}_M), 1]^\top.
\end{equation}
Intuitively, the hinged points are fixed references tiling the region, and $\bfphi(\bfx)$ collects the similarity of the query $\bfx$ to each of them.
The key step is to estimate the weights $\mathbf{w} \sim \calN(\bfmu, \bfSigma)$.

Given a point cloud $\calP$, we generate a dataset $\calD_\text{BHM}=\{\bfx_k, y_k \}_{k=1}^K \subset \bbR^3 \times \{-1, 1\}$ by sampling free-space samples, $\bfx_k, y_k=-1$, along each segment between the sensor position $\bfp$ and the hit point $\bfx\in\calP$ and adding the hit points as occupied samples, $\bfx_k = \bfx, y_k=1$. Given $\calD_\text{BHM}$, BHM uses expectation maximization (EM) \cite{bhm2017} to update the weights mean $\bfmu$ and covariance $\bfSigma$ and predict the log-odds as $l(\bfx)=\bfmu^\top \bfphi(\bfx)$.

\begin{figure*}[t]
    \centering
    \begin{subfigure}[b]{0.355\textwidth}
        \centering
        \includegraphics[width=\linewidth]{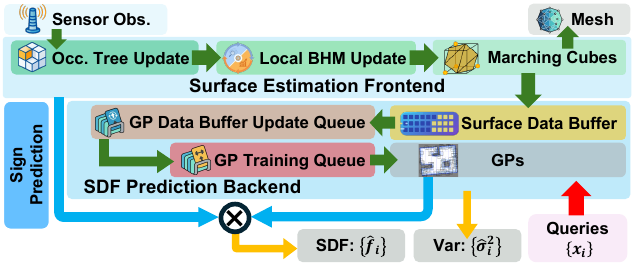}
        \caption{Overview}
        \vspace{-0.3em}
        \label{fig:method_overview}
    \end{subfigure}
    \hfill
    \begin{subfigure}[b]{0.184\textwidth}
        \centering
        \includegraphics[width=\linewidth]{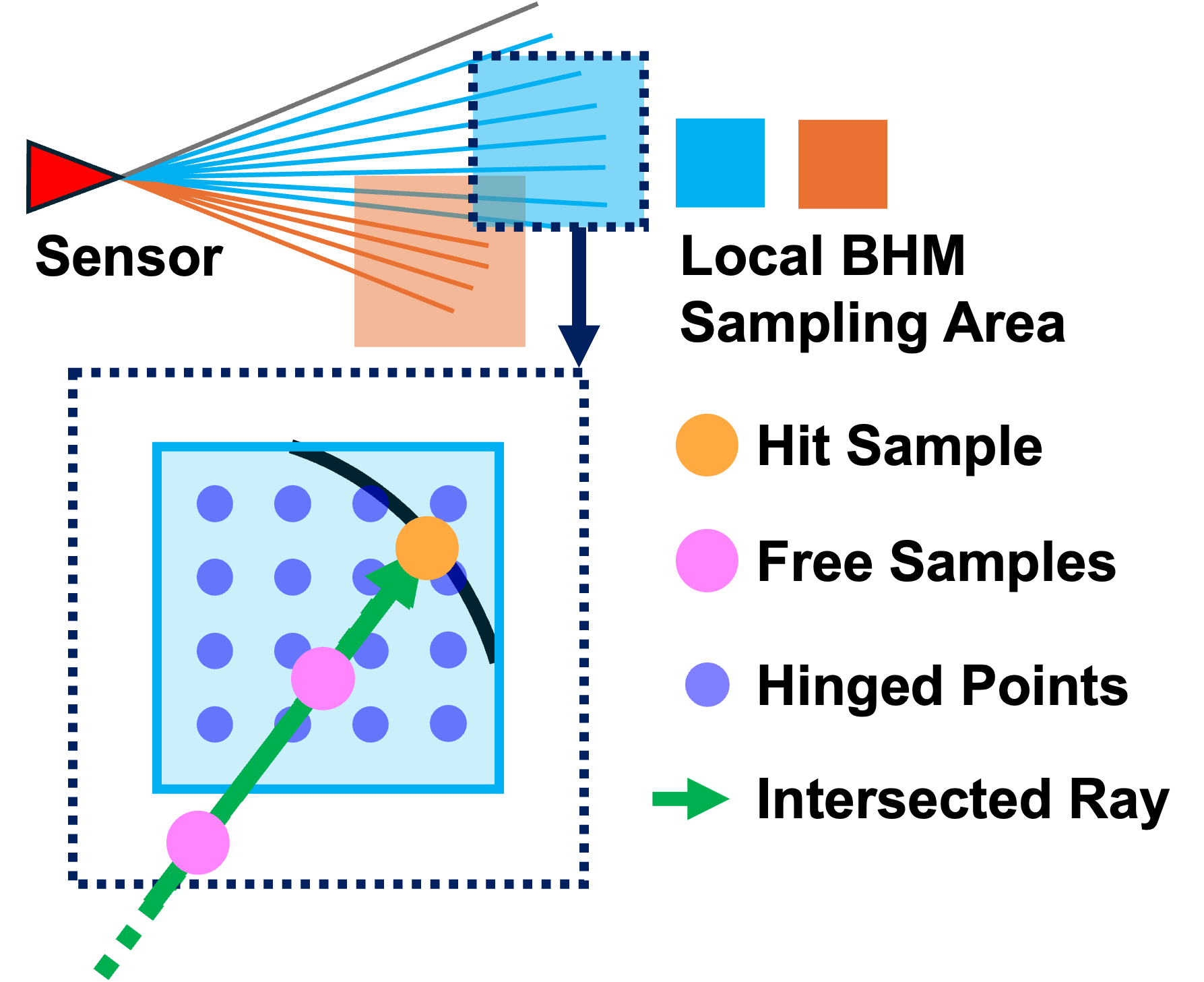}
        \caption{Sample generation}
        \vspace{-0.3em}
        \label{fig:bhm_sample_generation}
    \end{subfigure}
    \hfill
    \begin{subfigure}[b]{0.195\textwidth}
        \centering
        \includegraphics[width=\linewidth]{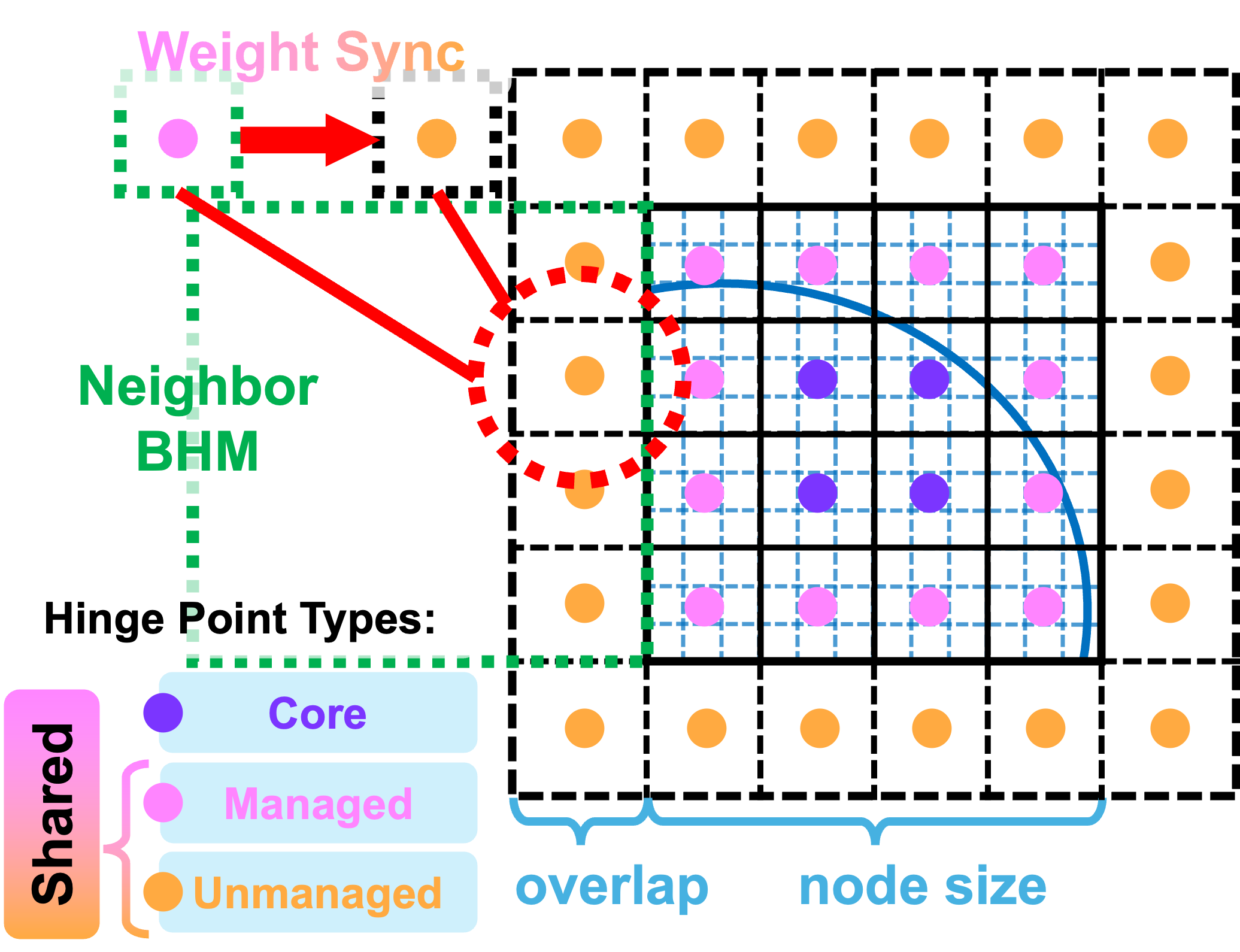}
        \caption{Weight sync}
        \vspace{-0.3em}
        \label{fig:bhm_weight_sync}
    \end{subfigure}
    \hfill
    \begin{subfigure}[b]{0.227\textwidth}
        \centering
        \includegraphics[width=\linewidth,trim={700pt 200pt 700pt 200pt},clip]{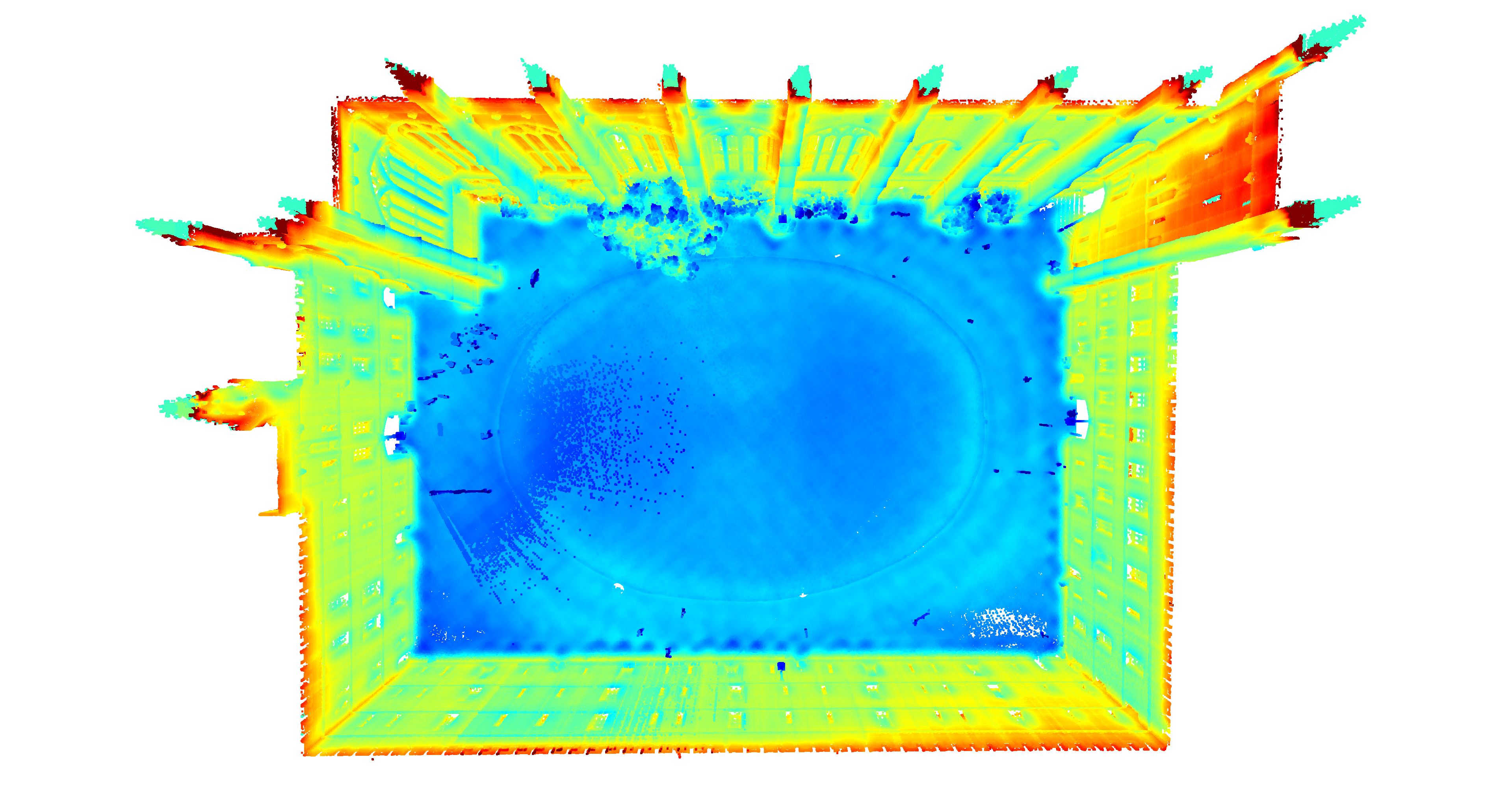}
        \caption{Surface log-odds}
        \vspace{-0.3em}
        \label{fig:gt_surf_pts_log_odds}
    \end{subfigure}
    \caption{\small (a) Overview of \methodname: the front-end estimates surface points and normals from point clouds via BHM and marching cubes; these train the back-end, which learns SDF and its gradient using multiple GPs in an octree. (b) For each ray hitting or crossing the sampling region, we generate occupied and free samples to update the BHM. (c) Weight sync between neighboring BHMs: {\color{magenta} managed} weights are synced to {\color{orange} unmanaged weights}. (d) Surface points colored by BHM log-odds; the ground has lower log-odds than the walls.}
    \label{fig:bhm_sample_and_weight_sync}
    \vspace{-0.5em}
\end{figure*}

2) {\bf Octree of BHMs}:
Since $M$ scales with scene size, BHM updates become prohibitive. We address this by partitioning the scene with an octree, maintaining a local BHM only for octants with occupied voxels, which greatly reduces the necessary hinged points.
To generate $\calD_\text{BHM}$, we prune rays outside the local BHM's viewing frustum and sample only within its bounding box (Fig.~\ref{fig:bhm_sample_generation}).
We also assume independent hinged point weights, reducing $\bfSigma$ to a diagonal matrix and enabling element-wise EM updates that drastically lower complexity.
Finally, we use a sparse representation of $\bfphi(\bfx)$, exploiting that the kernel vanishes when $\bfx$ is far from $\tilde{\bfx}$.

To maintain consistency between local BHMs, we overlap neighbors by a margin equal to the hinged point spacing and extend the sampling area when generating $\calD_\text{BHM}$ to sufficiently update boundary weights (Fig.~\ref{fig:bhm_sample_generation}).
For global consistency, we sync weights (Fig.~\ref{fig:bhm_weight_sync}): hinged point weights are categorized as core (unique to the local BHM), managed (owned locally but shared), and unmanaged (owned by a neighbor).
Periodically syncing managed weights to their unmanaged counterparts ensures consistent occupancy predictions and accelerates convergence by broadcasting local geometry to neighbors.

3) \text{\bf Sign Prediction}:
The SDF sign of a near-surface position $\bfx_*$ can be predicted by the BHM as:
\begin{equation} \label{eq:sign_prediction}
    s(\bfx_*) =
    \begin{cases}
        +1, & l(\bfx_*) < \tau, \\
        -1, & l(\bfx_*) \geq \tau,
    \end{cases}
\end{equation}
where $\tau$ is a threshold (ideally 0) learned from data.
Empirical evidence (Fig. \ref{fig:gt_surf_pts_log_odds}) shows $\tau$ varies with sensor viewing direction (e.g., ground vs. walls).
Thus, we update $\tau$ online using a moving average of the log-odds from $K$ hit samples:

\noindent\begin{tabular}{@{}p{\linewidth}@{}}
\vspace{-1.5em}
    \begin{equation}
    \tau \leftarrow (1-\alpha) \tau + \frac{\alpha}{K} \sum_{k=1}^{K} l(\bfx_k), \quad \alpha \in (0, 1], \label{eq:online_surf_logodds}
\end{equation}
\end{tabular}
where $\alpha\in(0, 1]$ is the learning rate.

4) {\bf Online Mesh Extraction and Surface Point Uncertainty Estimation}:
To recover surface points $\{\bfx_i\}_i$ at the $\tau$-level set of $l(\bfx)$, we use the marching cubes algorithm \cite{lorensen_marching_1987} on a grid of BHM log-odds values. We process only grid cells containing point cloud points and their neighbors. The uncertainty for each point $\bfx_i$ is computed as:
\begin{equation}
    \sigma_{\bfx_i}^2 = \beta (\tau - l(\bfx_i))^2 / \| \nabla l(\bfx_i) \|_2^2,
    \label{eq:surface_point_uncertainty}
\end{equation}
where $\beta>0$ is a hyperparameter and $\tau$ is updated via \eqref{eq:online_surf_logodds}.

We derive \eqref{eq:surface_point_uncertainty} via a first-order Taylor expansion of $l(\bfx_*)$ at the extracted point $\bfx_i$ relative to the true surface point $\bfx_*$:
\begin{equation}
    \underbrace{l(\bfx_*)}_{\tau} \approx l(\bfx_i) + {\underbrace{\nabla l(\bfx_i)}_{\nabla l}}^\top (\underbrace{\bfx_* - \bfx_i}_{\delta \bfx}),
\end{equation}
and assuming the error $\delta\bfx$ aligns primarily with the gradient direction, i.e., $\delta\bfx \approx {\nabla l}/{\|\nabla l\|_2} \delta_x$.
Intuitively, uncertainty rises when $l(\bfx_i)$ deviates from $\tau$ (higher surface extraction error) or when the gradient is small (flat and noise-sensitive field).
Hence, \eqref{eq:surface_point_uncertainty} reflects how confident the front-end is about each point, which allows the back-end to weight each point's contribution during SDF learning and uncertainty quantification.

\vspace{-1em}
\subsection{SDF Prediction Back-End}
\label{sec:backend}

In this section, we review the log-GP method for unsigned distance function (UDF) estimation, extend it to other kernel types to improve its numerical stability, and introduce a softmin SDF uncertainty quantification approach.

\subsubsection{\bf Log-GP for Unsigned Distance Estimation}
\label{sec:log_gp_udf}

Given the front-end sign prediction $s(\bfx)$ in \eqref{eq:sign_prediction}, we decompose the SDF as $d(\bfx) = s(\bfx) \cdot u(\bfx)$, where $u(\bfx) \geq 0$ is a UDF.
Thus, the back-end only needs to learn the UDF $u(\bfx)$.
Varadhan's formula \cite{varadhan_behavior_1967} shows that the UDF is related to the short-time heat diffusion equation:
\begin{equation} \label{eq:varadhan_equation}
    u(\bfx) = \lim_{t \to 0} \{-\sqrt{t} \log v(\bfx, t)\},
\end{equation}
where $v(\bfx,t)$ is the solution to the heat equation ${\partial v/\partial t = \Delta v}$ with boundary condition $v(\bfx,t)=1$ for $\bfx \in \partial\calO$.
Log-GPIS \cite{lan2021loggpis} shows that GP regression of $f(\bfx) = \exp(-\lambda u(\bfx))$ with Mat\'ern 3/2 kernel can approximate \eqref{eq:varadhan_equation} well when the kernel scale $\sqrt{3}/\lambda$ is small.
The UDF can be recovered as $u(\bfx) \approx - \log \bbE[f(\bfx)] / \lambda$.
The GP $f(\bfx)$ is trained on the dataset of surface points $\calD_\text{surf}=\{\bfx_i, \sigma_i^2\}_{i=1}^N$ with labels $f(\bfx_i)=\exp(-\lambda \cdot 0)=1$. The posterior mean and variance of $f(\bfx_*)$ at a query point $\bfx_*$ are given by:
\begin{equation}
	\begin{aligned}
		\begin{bmatrix} \hat{f}(\bfx_*) \\ \nabla \hat{f}(\bfx_*)
		\end{bmatrix} & =
		\begin{bmatrix} \bfk_*^\top \\ \nabla_{\bfx_*}^\top \bfk_*
		\end{bmatrix} (\bfK +\bfSigma)^{-1} \mathbf{1}, \\
		\bbV[\hat{f}(\bfx_*)] & = k(\bfx_*, \bfx_*) - \bfk_*^\top (\bfK +\bfSigma)^{-1}\bfk_*,
	\end{aligned}
	\label{eq:log_gp_posterior_mean}
\end{equation}
where $\bfk_* \in \bbR^N$ and $\bfK \in \bbR^{N\times N}$ are calculated by the kernel:
\begin{align}
	k_{*,i} & = k(\bfx_i, \bfx_*) \quad 1 \le i \le N,                \\
	K_{ij}  & = k(\bfx_i, \bfx_j) \quad 1 \le i \le N, 1 \le j \le N,
\end{align}
and $\Sigma_{ii} = \sigma_i^2$. The UDF gradient can also be obtained as $\nabla u(\bfx) = - \nabla \hat{f}(\bfx) / \| \nabla \hat{f}(\bfx) \|_2$.

Each octant with a BHM also maintains a log-GP for SDF estimation. For inter-GP consistency, we collect surface points within boxes larger than the BHM sampling box and fuse the $K$ nearest GPs via $\hat{u}(\bfx_*) = \min_k u_k\left(\bfx_*\right)$.
Because the collection boxes overlap, neighboring GPs share samples near boundaries and yield almost identical predictions to a single global GP\footnote{The difference between the min-fusion distance and gradient and a global GP was around $\sim\!10^{-8}\,$m and $<\!10^{-6}\,$rad for a 2D benchmark.}. Therefore, the min-fusion of multiple local GPs yields continuous results across octant boundaries.

\begin{table}[t]
	\centering
	\caption{Kernel functions and their nonlinear transforms.}
    \label{tab:kernel_nonlinear_transform}
	\begin{tabular}{c c c}
		\hline \rule{0pt}{2.3ex}
		Kernel       & $k(r)$  & $r(\hat{f})$ \\
		\hline \rule{0pt}{3.5ex}
		RBF          & $\exp\left(-{r^2}/{2l^2}\right)$ & $\sqrt{-2 l^2 \log \hat{f}}$       \\
		Mat\'ern 3/2 & $\left(1+{\sqrt{3}r}/{l}\right)\exp\left(-{\sqrt{3}r}/{l}\right)$ & $-{l \log \hat{f}}/{\sqrt{3}}$ \\ [1ex]
		\hline
	\end{tabular}
    \vspace{-0.5em}
\end{table}

\subsubsection{\bf Extension to Other Kernels}
\label{sec:kernel_extension}

While the Mat\'ern 3/2 kernel is standard for log-GP, we show that more efficient kernels such as the RBF can be used by identifying the posterior mean $\hat{f}$ with a softmin-like function, i.e., the softmin approximation of the UDF over $\calD_\text{surf}$: $u(\bfx_*) \approx h(\bfx_*, \{\bfx_i\}_{i=1}^N) = \bfs^\top \bfz$,
where $z_i = \left\|\bfx_i - \bfx_*\right\|_2$ and $s_i = {e^{-\alpha z_i}}/{\sum_i^N e^{-\alpha z_i}}, \alpha>0$.
Since every label equals $1$, the posterior mean \eqref{eq:log_gp_posterior_mean} reduces to $\hat{f}_* = \bfk_*^\top(\bfK+\bfSigma)^{-1}\mathbf{1}$ with $\bfSigma=\sigma^2\bfI$.
In the \emph{weakly-correlated} regime (small kernel scale relative to the sample spacing, $\bfK\!\to\!\bfI$), $\hat{f}_*\approx\tfrac{1}{1+\sigma^2}\sum_i k_{*i}=\tfrac{S}{1+\sigma^2}\max_i k_{*i}$ with $S=\sum_i k_{*i}/\max_i k_{*i}\ge1$, so
\begin{equation} \label{eq:weak_corr_limit}
    \log\hat{f}_* \approx \log\max_i k_{*i} + \log S - \log(1+\sigma^2),
\end{equation}
recovering the softmin as $S\!\to\!1$ (a single dominant kernel bump).
In the \emph{strongly-correlated} regime (dense sampling, $\bfK\!\to\!\mathbf{1}\mathbf{1}^\top$), for a query at distance $z$ from a tight surface cluster the Sherman--Morrison identity gives $\hat{f}_*\approx\tfrac{S}{\sigma^2+N}\max_i k_{*i}\approx\tfrac{N}{\sigma^2+N}k(z)$ with $S\!\to\!N$, so
\begin{equation} \label{eq:strong_corr_limit}
    \log\hat{f}_* \approx \log k(z) - \log\tfrac{\sigma^2+N}{N},
\end{equation}
where the bumps merge into one wide bump (uniform-weight, zero-temperature softmin).
In both limits, the UDF is recovered by the kernel-specific transform $r(\hat{f})$ in Table~\ref{tab:kernel_nonlinear_transform}.
In the weakly-correlated regime, the softmin inflation $\log S$ biases the recovered UDF slightly below the true distance, a deterministic under-estimate that shrinks with the kernel scale. In the strongly-correlated regime, it is absorbed by the $\sigma^2\!+\!N$ normalization. The noise $\sigma^2$ down-weights uncertain samples and offsets the under-estimate.
We pick the RBF kernel for its efficiency and better accuracy, shown in the ablation of Sec.~\ref{sec:kernel_ablation}.
While larger $\lambda$ (smaller kernel scale) improves the approximation of \eqref{eq:varadhan_equation}, it triggers numerical underflow of $k(\bfx_*,\bfx_i)$. We address this by subtracting a constant $d$ from the kernel exponent and adding it back in the inverse transforms: $r(\hat{f'})= \sqrt{-2l^2\log\hat{f'}+d^2}$ for RBF and $r(\hat{f'})= -\frac{l}{\sqrt{3}} \log \hat{f'} + d$ for Mat\'ern 3/2.

\subsubsection{\bf Consistent Uncertainty Quantification via Softmin-based Fusion}
\label{sec:consistent_uncertainty_quantification}

Standard first-order variance propagation, $\bbV[\hat{d}] \!\approx\!\left(r'\right)^2\!\bbV[\hat{f}]$, fails in log-GP: as $\bfx_*$ moves away from $\calD_\text{surf}$, $\bbV[\hat{f}]\tigtto 1$ and $|r'|\tigtto\infty$, so $\bbV[\hat{d}]\tigtto\infty$.
Instead, we leverage the softmin property of the log-GP posterior to propagate surface point uncertainties $\sigma_i^2$ directly through the softmin.

The variances of $u(\bfx_*)$ and $\nabla u(\bfx_*)$ then follow from a first-order approximation of the softmin:
\begin{align}
	\bbV[u(\bfx_*)] & \approx \sum_{i=1}^N \left\| \nabla_{\bfx_i} h \right\|_2^2 \sigma_i^2, \\
	\bbV[\nabla_k u(\bfx_*)] & \approx \sum_{i=1}^N \left\| \nabla_{\bfx_i} g_k(\bfx_*,\{\bfx_i\}_{i=1}^N) \right\|_2^2 \sigma_i^2, \label{eq:sdf_gradient_variance_approximation}
\end{align}
where $\nabla_{\bfx_i} h = -s_i(\alpha\bfs^\top\bfz - \alpha z_i + 1) ({\bfx_* - \bfx_i})/{z_i}$, $\bfg(\bfx_*,$ $\{\bfx_i\}_{i=1}^N)$ $= \nabla_{\bfx_*}h(\bfx_*,\{\bfx_i\}_{i=1}^N)$.
This directly propagates the surface point location uncertainty to the SDF and gradient predictions.
The deterministic-sign treatment is justified empirically: the occupancy field transitions sharply at the surface (on Replica \texttt{room0}, $97\%$ of g.t. surface points have $q<0.1$ or $q>0.9$, with a median transition band $\approx 3.6\,$mm, far below the map resolution), so $q$ is effectively binary off the surface. The sign mixture collapses, giving $\bbV[\hat{d}(\bfx)]=\bbV[\hat{u}(\bfx)]$.
\edit{For near-surface points without a sharp transition, $q \in (0.1,0.9)$, the deterministic-sign treatment theoretically underestimates the predictive variance.}

\vspace{-1em}
\subsection{Priority Queue Based Update}
\label{sec:implementation_details}

Beyond the core BHM and GP optimizations, we use priority-queue-based updates for efficiency.
Running surface extraction, GP buffer updates, and GP training after every BHM update is wasteful given frequent sensor measurements, so we delay these processes until necessary using a three-queue priority system (Fig.~\ref{fig:method_overview}).
To stay responsive, frequently queried GPs get higher priority for buffer updates and retraining, while the oldest BHM marching requests are prioritized to process likely stable regions.
This strategy achieves a balance between responsiveness and map correctness.

\section{Experiments}
\label{sec:experiments}

\begin{table*}[htbp]
\centering
\caption{\small Mesh reconstruction metrics on Replica, Cow\&Lady (C.L. for the scene and Cow for the cow only), and Newer College (N.C.). When computing the ratio, $\delta=5$ cm (20cm for Newer College). The best and second-best results are bold and underlined, respectively.}
\label{table:mesh_metrics}
\scriptsize
\def\fonecell{\multirow{5}{*}{F1 Score [$<\delta$]\% $\uparrow$}}
\def\recallcell{\multirow{5}{*}{Recall [$<\delta$]\% $\uparrow$}}
\def\precisioncell {\multirow{5}{*}{Precision [$<\delta$]\% $\uparrow$}}
\def\compratiocell{\multirow{5}{*}{\begin{tabular}{@{}c@{}}Completion Ratio\\ $[<\delta]\% \uparrow$\end{tabular}}}
\def\completioncell{\multirow{5}{*}{Completion [cm] $\downarrow$}}
\def\accuracycell{\multirow{5}{*}{Accuracy [cm] $\downarrow$}}
\def\chamfercell{\multirow{5}{*}{Chamfer-L1 [cm] $\downarrow$}}
\begin{tabular}{l|l|ccccccccccc}
    \hline
    Metric           & Method   & room 0            & room 1            & room 2            & office 0          & office 1          & office 2          & office 3          & office 4          & C.L.              & Cow               & N.C.              \\
    \hline
    \fonecell         &  Ours      &  \textbf{95.86}  &  \textbf{97.32}  &  \textbf{97.54}  &  \textbf{98.23}  &  \textbf{96.57}  &  \textbf{95.11}  &  \textbf{93.58}  &  \textbf{94.53}  &  \textbf{83.47}  &  \textbf{92.74}  &  \textbf{75.80} \\
                      &  FIESTA    &  76.71              &  76.49              &  78.15              &  77.64              &  82.43              &  75.45              &  79.73              &  78.19              &  19.37              &  18.03              &  54.73 \\
                      &  Voxblox   &  \underline{92.00}  &  \underline{94.77}  &  \underline{92.64}  &  \underline{93.72}  &  \underline{94.23}  &  \underline{89.50}  &  86.58              &  89.94              &  70.30              &  79.38              &  54.14 \\
                      &  iSDF      &  87.85              &  85.11              &  90.64              &  89.80              &  91.37              &  87.53              &  \underline{87.52}  &  \underline{91.21}  &  78.33              &  \underline{91.84}  &  \underline{70.15} \\
                      &  VDB-GPDF  &  60.06              &  57.20              &  58.45              &  60.99              &  56.47              &  58.00              &  54.78              &  56.92              &  \underline{78.37}  &  91.08              &  61.94 \\
    \hline
    \recallcell       &  Ours      &  \textbf{92.94}  &  \textbf{95.51}  &  \textbf{95.50}  &  \textbf{97.53}     &  \textbf{95.54}     &  \textbf{92.10}  &  \textbf{90.15}  &  \underline{90.06}  &  \textbf{97.42}  &  \textbf{98.98}  &  \underline{75.10} \\
                      &  FIESTA    &  77.56              &  78.33              &  81.04              &  81.33              &  82.00              &  77.17              &  81.15              &  78.09              &  19.89              &  17.27              &  52.60 \\
                      &  Voxblox   &  \underline{88.17}  &  \underline{91.81}  &  \underline{90.79}  &  \underline{92.18}  &  \underline{90.62}  &  \underline{85.67}  &  82.21              &  84.94              &  72.71              &  78.03              &  56.64 \\
                      &  iSDF      &  82.88              &  77.54              &  88.64              &  85.51              &  86.30              &  83.30              &  \underline{85.46}  &  \textbf{90.12}     &  \underline{82.70}  &  \underline{93.47}  &  \textbf{86.24} \\
                      &  VDB-GPDF  &  58.03              &  55.50              &  57.29              &  60.30              &  55.36              &  56.55              &  51.47              &  55.12              &  80.73              &  91.61              &  62.44 \\
    \hline
    \precisioncell    &  Ours      &  \textbf{98.96}  &  \textbf{99.21}     &  \textbf{99.66}  &  \textbf{98.95}  &  \underline{97.62}  &  \textbf{98.32}  &  \textbf{97.28}  &  \textbf{99.46}  &  73.01  &  87.23  &  \textbf{76.50} \\
                      &  FIESTA    &  75.87              &  74.73              &  75.47              &  74.28              &  82.85              &  73.81              &  78.36              &  78.28              &  18.87              &  18.86              &  57.03 \\
                      &  Voxblox   &  \underline{96.18}  &  \underline{97.94}  &  \underline{94.56}  &  \underline{95.31}  &  \textbf{98.13}     &  \underline{93.69}  &  \underline{91.44}  &  \underline{95.55}  &  68.05              &  80.77              &  51.84 \\
                      &  iSDF      &  93.45              &  94.33              &  92.74              &  94.56              &  97.08              &  92.21              &  89.68              &  92.33              &  \underline{74.40}  &  \underline{90.27}  &  59.12 \\
                      &  VDB-GPDF  &  62.24              &  59.01              &  59.66              &  61.69              &  57.63              &  59.52              &  58.53              &  58.85              &  \textbf{76.14}     &  \textbf{90.56}     &  \underline{61.45} \\
    \hline
    \compratiocell    &  Ours      &  \textbf{92.41}  &  \textbf{95.33}  &  \textbf{95.31}  &  \textbf{97.49}  &  \textbf{95.44}  &  \textbf{91.55}  &  \textbf{89.35}  &  \underline{88.99}  &  \textbf{93.37}  &  \textbf{97.33}  &  \underline{74.76} \\
                      &  FIESTA    &  78.05              &  79.32              &  82.34              &  82.94              &  81.82              &  78.16              &  81.80              &  78.03              &  23.99              &  9.67               &  48.61 \\
                      &  Voxblox   &  \underline{87.09}  &  \underline{91.26}  &  \underline{90.41}  &  \underline{91.92}  &  \underline{89.85}  &  \underline{84.33}  &  80.21              &  83.06              &  74.46              &  77.26              &  60.31 \\
                      &  iSDF      &  80.69              &  72.68              &  88.12              &  83.98              &  84.59              &  81.51              &  \underline{84.74}  &  \textbf{89.88}     &  \underline{84.43}  &  \underline{93.69}  &  \textbf{90.57} \\
                      &  VDB-GPDF  &  54.98              &  52.69              &  55.53              &  59.39              &  53.53              &  54.26              &  44.82              &  52.08              &  81.83              &  91.70              &  63.03 \\
    \hline
    \completioncell   &  Ours      &  \textbf{2.85}  &  \underline{2.37}  &  \underline{2.33}   &  \underline{2.07}   &  \underline{2.25}  &  \textbf{3.02}      &  \textbf{3.01}  &  \underline{3.50}   &  \textbf{2.60}  &  \underline{2.57}   &  \underline{14.22} \\
                      &  FIESTA    &  3.98               &  3.16               &  2.88               &  2.73               &  4.71               &  4.40               &  3.69               &  3.98               &  18.68              &  25.77              &  71.87 \\
                      &  Voxblox   &  \underline{3.69}   &  \textbf{2.16}      &  \textbf{2.31}      &  \textbf{1.71}      &  \textbf{2.07}      &  \underline{3.52}   &  4.37               &  4.38               &  4.44               &  5.77               &  21.30 \\
                      &  iSDF      &  5.63               &  8.72               &  2.62               &  4.14               &  3.98               &  5.62               &  \underline{3.03}   &  \textbf{2.23}      &  \underline{3.12}   &  \textbf{2.24}      &  \textbf{9.75} \\
                      &  VDB-GPDF  &  5.52               &  4.82               &  3.87               &  3.52               &  4.12               &  4.80               &  13.78              &  5.07               &  3.68               &  3.05               &  40.57 \\
    \hline
    \accuracycell     &  Ours      &  \underline{1.93}  &  1.82  &  1.94  &  1.90  &  1.86               &  1.98  &  \textbf{1.96}      &  2.00               &  \underline{4.69}   &  \underline{3.25}  &  \textbf{14.03} \\
                      &  FIESTA    &  3.57               &  3.19               &  3.18               &  3.43               &  3.11               &  4.23               &  3.51               &  3.22               &  26.67              &  16.58              &  21.02 \\
                      &  Voxblox   &  \textbf{1.74}      &  \textbf{1.14}      &  \textbf{1.50}      &  \textbf{1.22}      &  \textbf{0.81}      &  \textbf{1.68}      &  \underline{2.16}   &  \textbf{1.67}      &  4.98               &  3.64               &  25.44 \\
                      &  iSDF      &  2.04               &  \underline{1.60}   &  \underline{1.52}   &  \underline{1.64}   &  \underline{1.11}   &  \underline{1.89}   &  2.30               &  \underline{1.83}   &  \textbf{4.66}      &  \textbf{2.75}      &  31.66 \\
                      &  VDB-GPDF  &  3.42               &  3.34               &  3.17               &  3.30               &  3.55               &  3.32               &  3.55               &  3.37               &  4.94               &  3.28               &  \underline{18.56} \\
    \hline
    \chamfercell      &  Ours      &  \textbf{2.39}  &  \underline{2.09}  &  2.13               &  \underline{1.99}  &  \underline{2.06}   &  \textbf{2.50}  &  \textbf{2.48}      &  \underline{2.75}   &  \textbf{3.65}  &  \underline{2.91}   &  \textbf{14.13} \\
                      &  FIESTA    &  3.77               &  3.17               &  3.03               &  3.08               &  3.91               &  4.31               &  3.60               &  3.60               &  22.68              &  21.17              &  46.45 \\
                      &  Voxblox   &  \underline{2.72}   &  \textbf{1.65}      &  \textbf{1.90}      &  \textbf{1.47}      &  \textbf{1.44}      &  \underline{2.60}   &  3.26               &  3.02               &  4.71               &  4.70               &  23.37 \\
                      &  iSDF      &  3.83               &  5.16               &  \underline{2.07}   &  2.89               &  2.55               &  3.75               &  \underline{2.67}   &  \textbf{2.03}      &  \underline{3.89}   &  \textbf{2.49}      &  \underline{20.71} \\
                      &  VDB-GPDF  &  4.47               &  4.08               &  3.52               &  3.41               &  3.83               &  4.06               &  8.67               &  4.22               &  4.31               &  3.17               &  29.57 \\
    \hline
\end{tabular}
\vspace{-1.5em}
\end{table*}

\newcommand{\mccell}[1]{%
	\begin{minipage}[c]{0.152\linewidth}%
		\centering\includegraphics[width=\linewidth]{#1}%
	\end{minipage}}
\newcommand{\mccellcow}[1]{%
	\begin{minipage}[c]{0.152\linewidth}%
		\centering\includegraphics[width=\linewidth,trim={0pt 100pt 0pt 10pt},clip]{#1}%
	\end{minipage}}
\newcommand{\mchead}[1]{%
	\begin{minipage}[c]{0.152\linewidth}%
		\centering{\small #1}%
	\end{minipage}}
\newcommand{\mcrowlabel}[1]{%
	\begin{minipage}[c]{0.04\linewidth}%
		\centering\rotatebox{90}{\small #1}%
	\end{minipage}}

\begin{figure*}[t]
	\centering
	\mcrowlabel{}\hfill
	\mchead{Ground Truth}\hfill
	\mchead{Ours}\hfill
	\mchead{FIESTA \cite{han_fiesta_2019}}\hfill
	\mchead{Voxblox \cite{oleynikova_voxblox_2017}}\hfill
	\mchead{iSDF \cite{ortiz_isdf_2022}}\hfill
	\mchead{VDB-GPDF \cite{wu_vdb-gpdf_2025}}\par
	\mcrowlabel{\shortstack{Replica\\(Room 1)}}\hfill
	\mccell{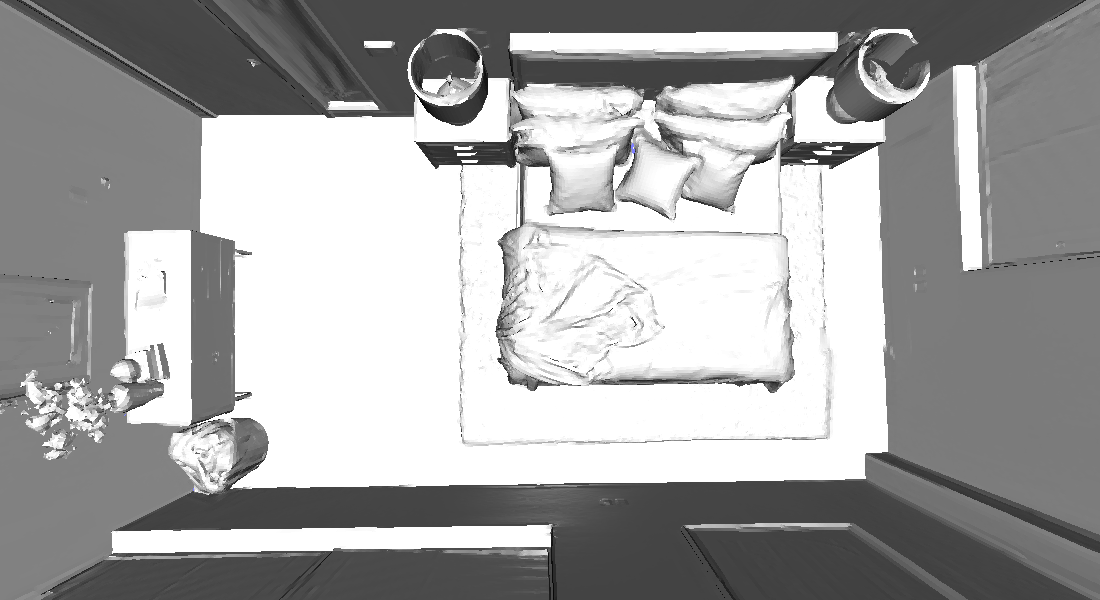}\hfill
	\mccell{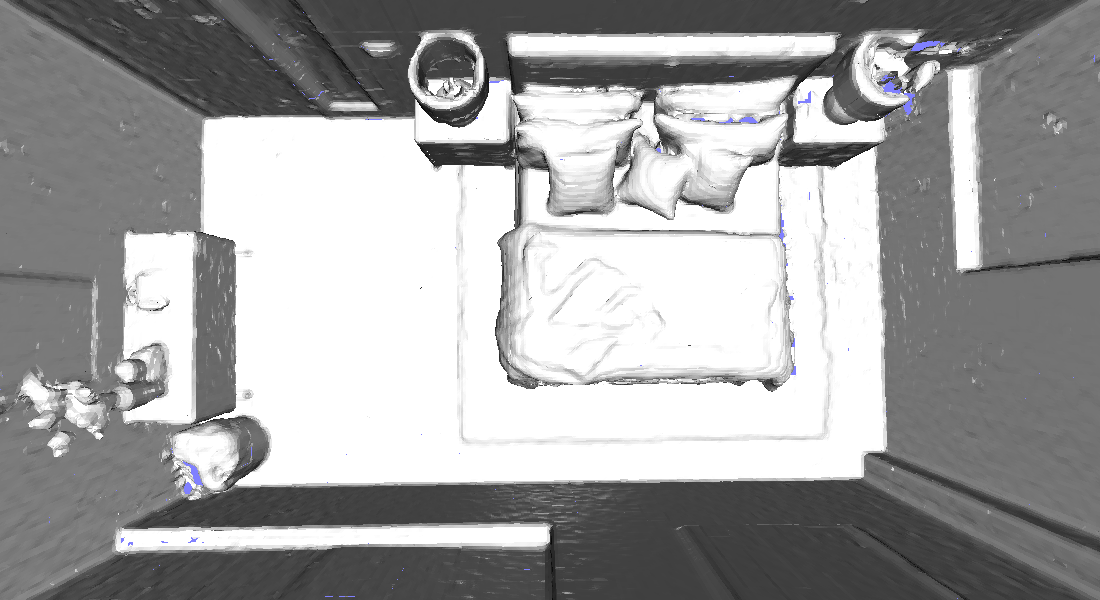}\hfill
	\mccell{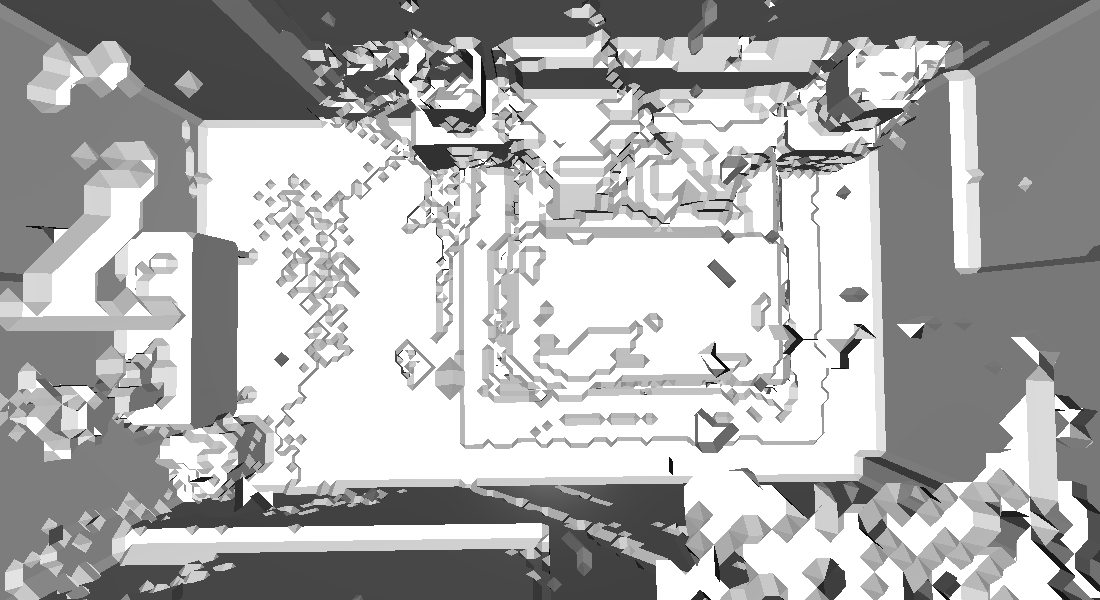}\hfill
	\mccell{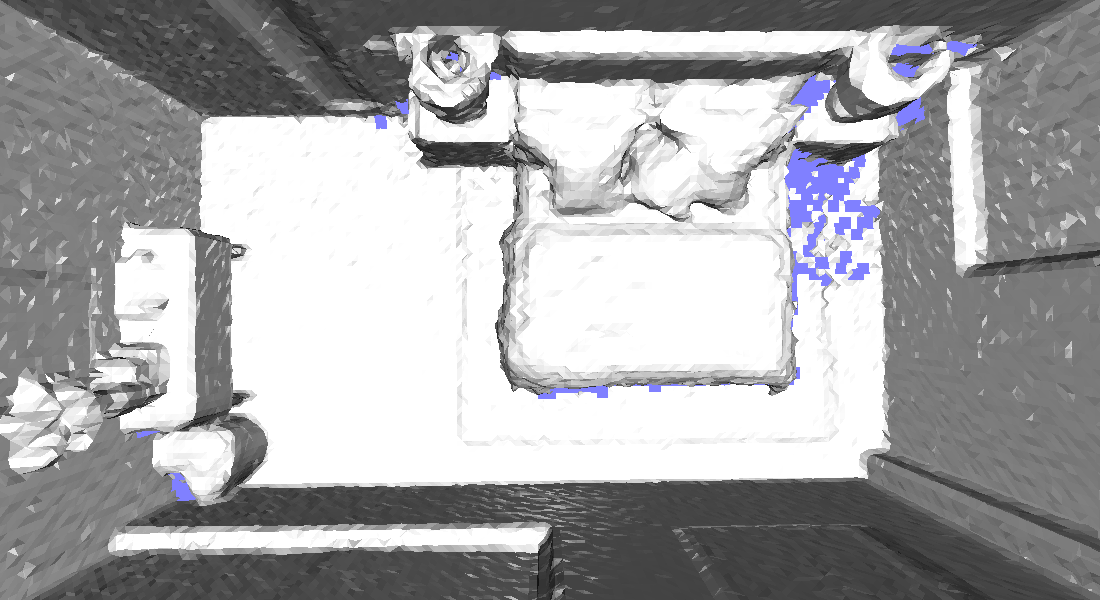}\hfill
	\mccell{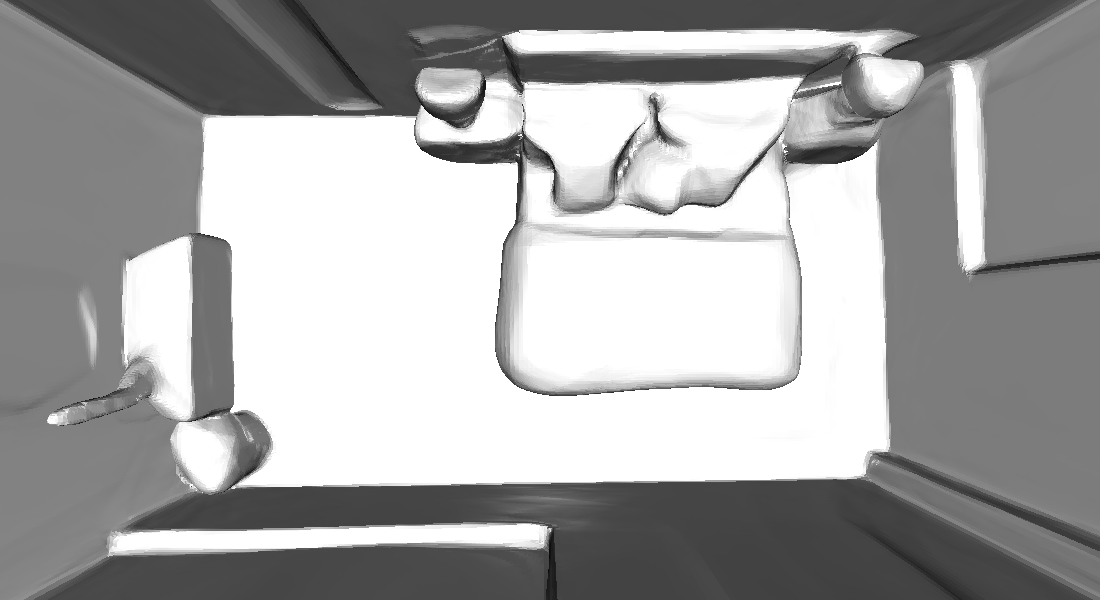}\hfill
	\mccell{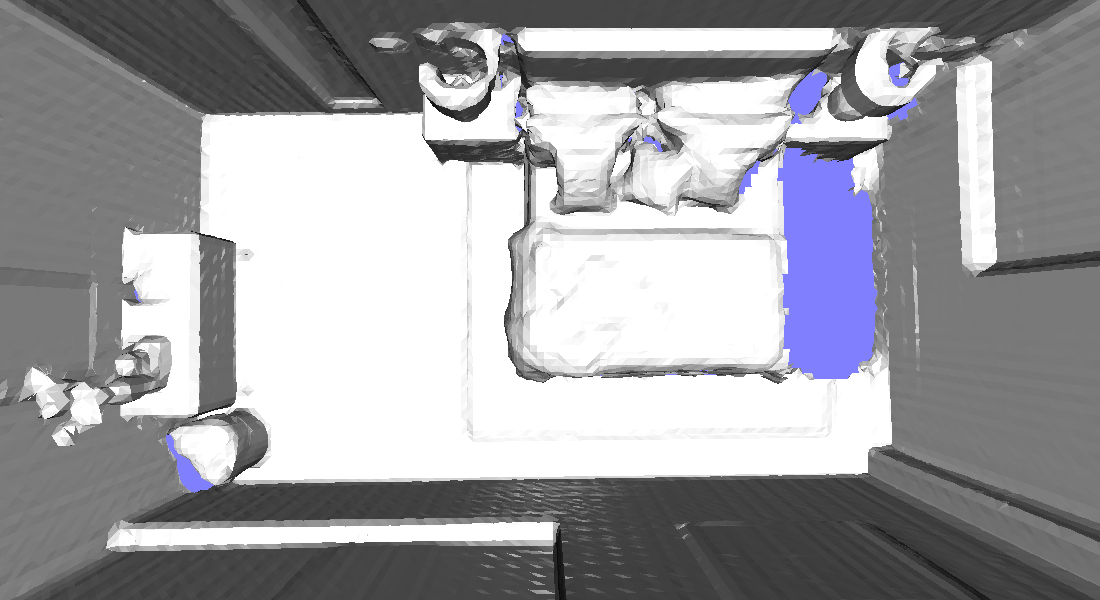}\par
	\vspace{1pt}
	\mcrowlabel{\shortstack{Newer\\College}}\hfill
	\mccell{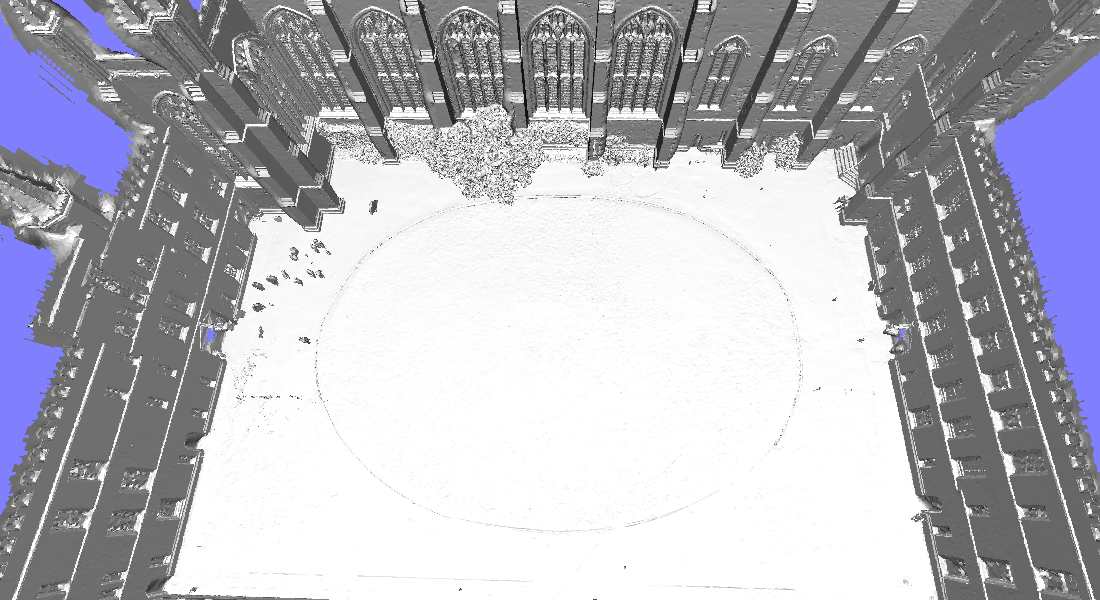}\hfill
	\mccell{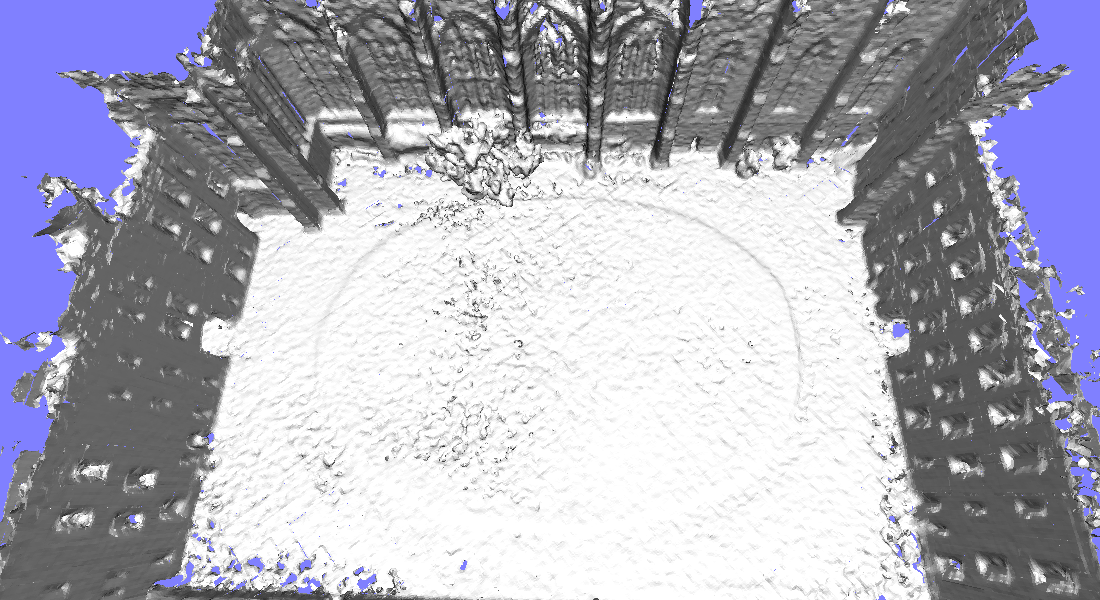}\hfill
	\mccell{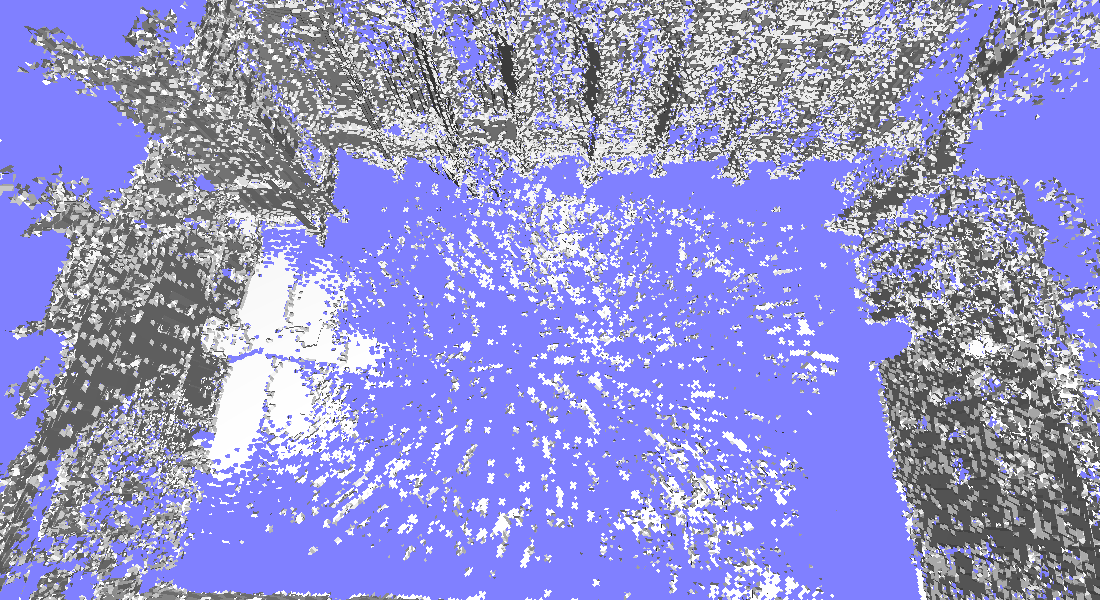}\hfill
	\mccell{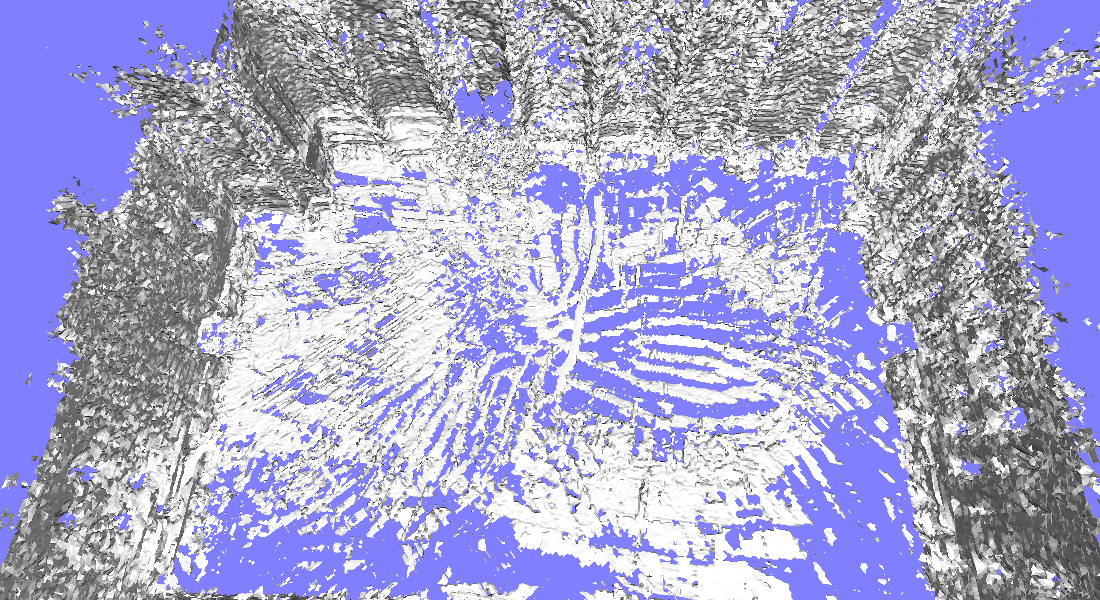}\hfill
	\mccell{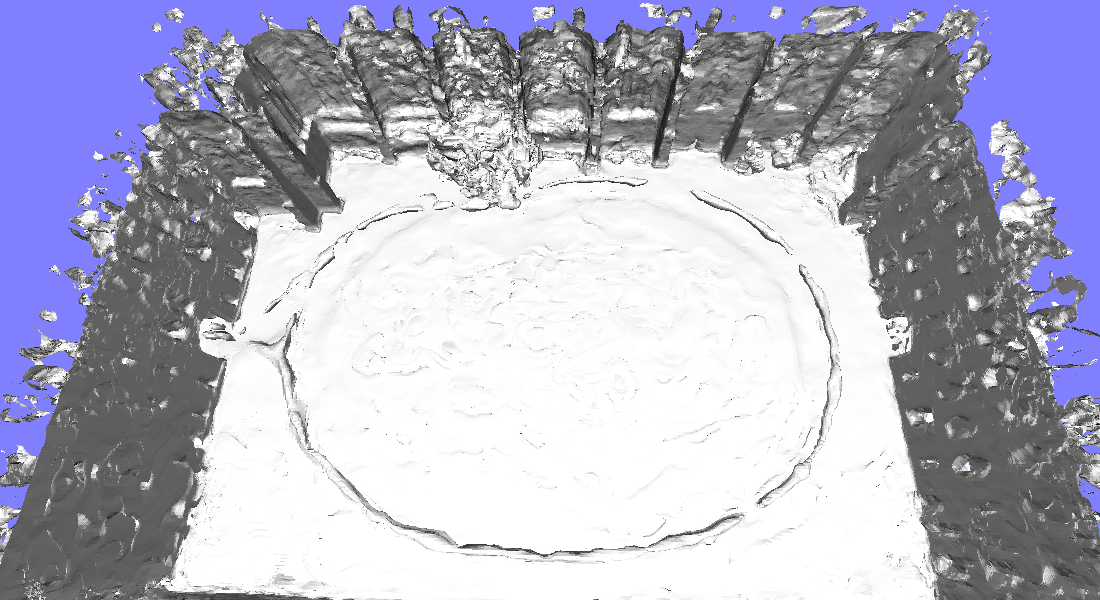}\hfill
	\mccell{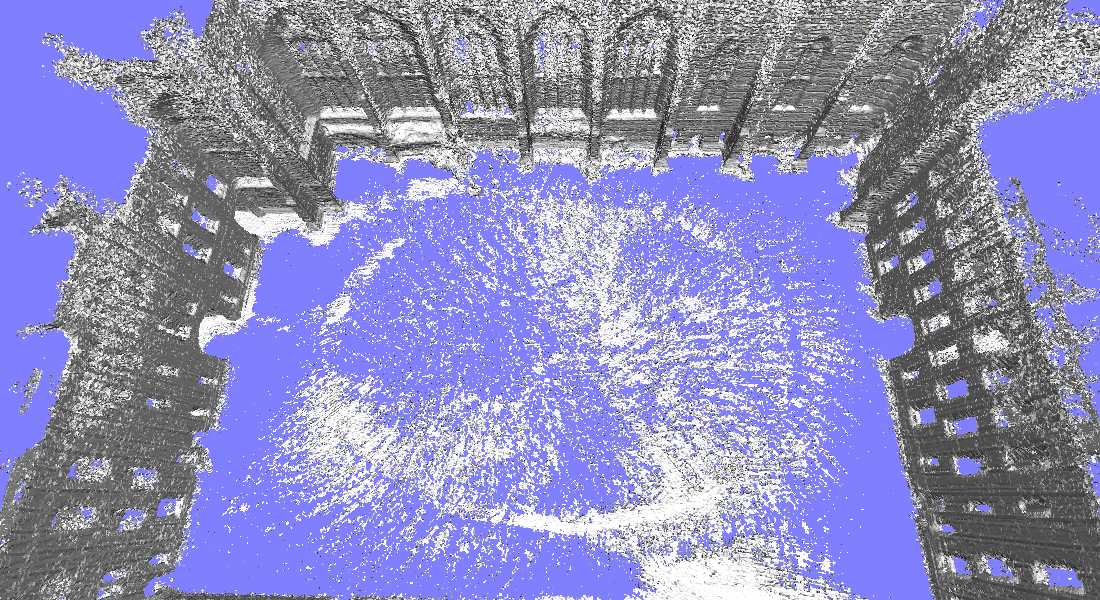}\par
	\vspace{1pt}
	\caption{
		\small Qualitative comparison of mesh reconstruction on Replica \cite{replica19arxiv} (top) and Newer College \cite{newercollege2021} (bottom). \methodname achieves the best visual quality, accuracy, and completeness; e.g., the Replica bed-sheet textures are preserved where the baselines fail.
		}
	\label{fig:mesh_comp}
	\vspace{-0.25em}
\end{figure*}

We compare \methodname to four baselines: Voxblox \cite{oleynikova_voxblox_2017}, FIESTA \cite{han_fiesta_2019}, iSDF \cite{ortiz_isdf_2022}, and VDB-GPDF \cite{wu_vdb-gpdf_2025} on the Replica \cite{replica19arxiv}, Cow\&Lady \cite{oleynikova_voxblox_2017} and Newer College \cite{newercollege2021}.
For Replica, we simulate depth noise using an axial model with $\sigma=0.0025\!z^2$ \cite{d435_sensor_noise_2019}.
All baselines use the same input point clouds, poses, and frame rates as \methodname. Each grid-based method's voxel resolution matches the evaluation resolution ($0.05\,$m for Replica and Cow\&Lady, $0.2\,$m for Newer College). Other parameters follow the authors' defaults.
We analyze reconstruction quality (Sec.\ref{sec:reconstruction_accuracy}), SDF accuracy (Sec. \ref{sec:sdf_accuracy}), efficiency (Sec.\ref{sec:computation_efficiency}), scalability (Sec. \ref{sec:scalability}), kernel choice (Sec.\ref{sec:kernel_ablation}) and individual components (Sec.\ref{sec:component_analysis}), then verify the consistency between estimated SDF variance and actual error (Sec.\ref{sec:variance_consistency}), and study the impact of sensor noise (Sec.\ref{sec:ablation_study_sensor_noise}). Finally, we show a real-world robot navigation demonstration (Sec.\ref{sec:path_planning_demo}).

\vspace{-1em}
\subsection{Reconstruction Accuracy}
\label{sec:reconstruction_accuracy}

\begin{figure}
	\centering
    \begin{subfigure}[t]{0.19\linewidth}
        \centering
        \includegraphics[width=\linewidth,trim={0pt 40pt 0pt 0pt},clip]{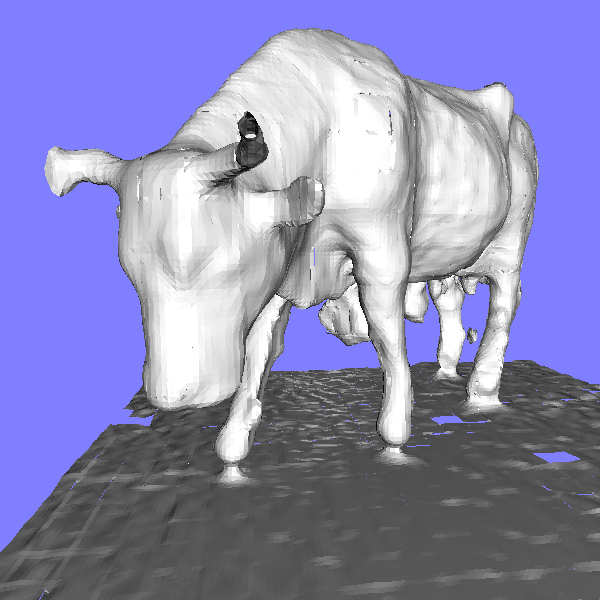}
        \caption*{Ours}
        \label{fig:mesh_comp_cow_ours}
    \end{subfigure}
    \begin{subfigure}[t]{0.19\linewidth}
        \centering
        \includegraphics[width=\linewidth,trim={0pt 40pt 0pt 0pt},clip]{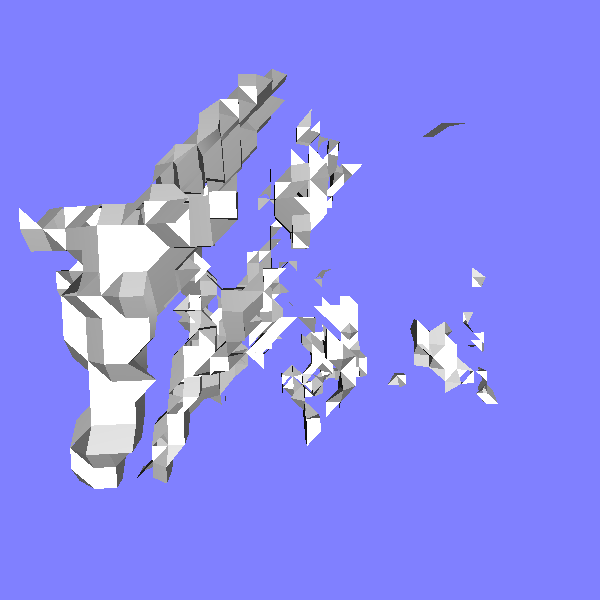}
        \caption*{FIESTA}
        \label{fig:mesh_comp_cow_fiesta}
    \end{subfigure}
    \begin{subfigure}[t]{0.19\linewidth}
        \centering
        \includegraphics[width=\linewidth,trim={0pt 40pt 0pt 0pt},clip]{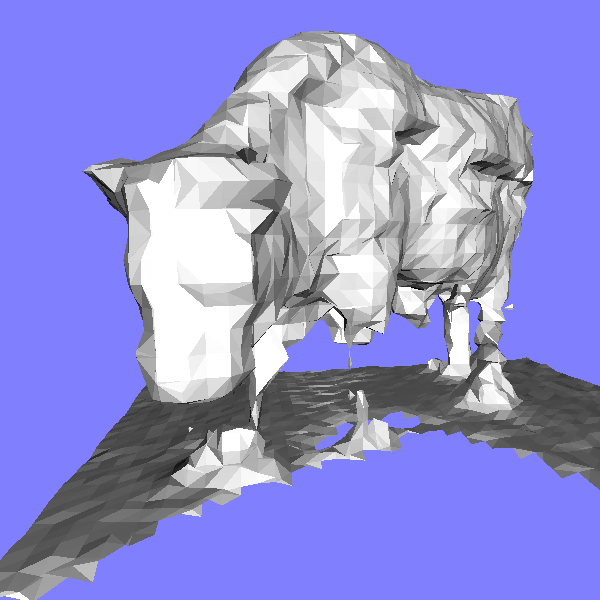}
        \caption*{Voxblox}
        \label{fig:mesh_comp_cow_voxblox}
    \end{subfigure}
    \begin{subfigure}[t]{0.19\linewidth}
        \centering
        \includegraphics[width=\linewidth,trim={0pt 40pt 0pt 0pt},clip]{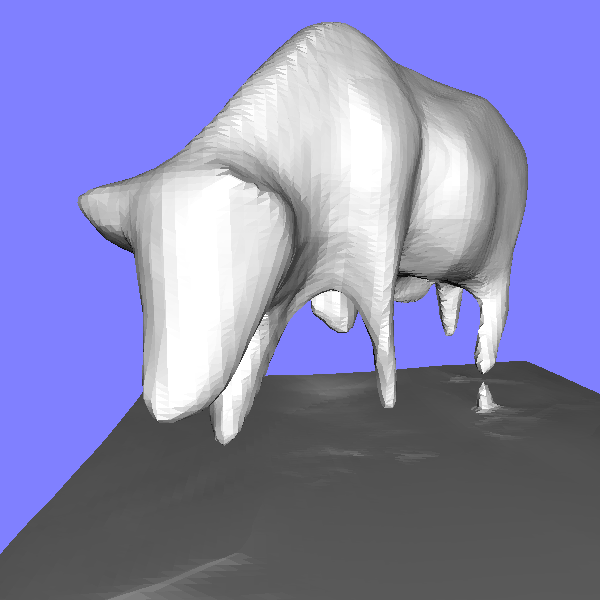}
        \caption*{iSDF}
        \label{fig:mesh_comp_cow_isdf}
    \end{subfigure}
    \begin{subfigure}[t]{0.19\linewidth}
        \centering
        \includegraphics[width=\linewidth,trim={0pt 40pt 0pt 0pt},clip]{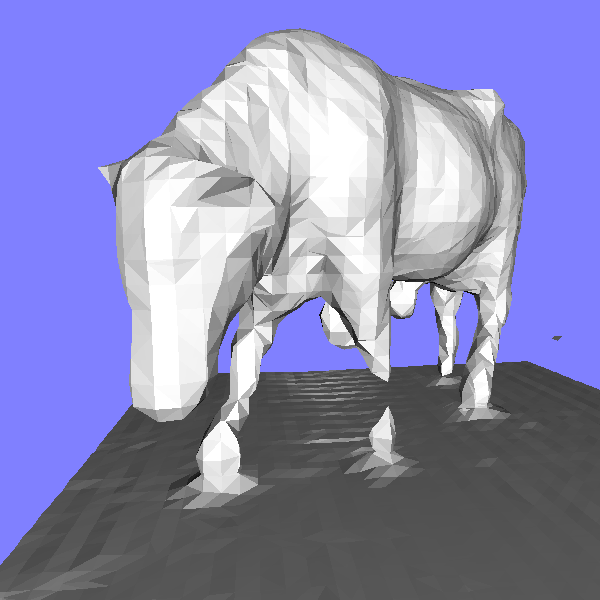}
        \caption*{VDB-GPDF}
        \label{fig:mesh_comp_cow_vdb-gpdf}
    \end{subfigure}
	\caption{\small Mesh reconstruction of the cow in Cow\&Lady. Our method successfully recovers the detailed geometry (e.g., the horns, ears, and legs), which are missing in the baseline reconstructions.}
    \label{fig:mesh_comp_cow}
\end{figure}

As shown in Figs. \ref{fig:mesh_comp} and \ref{fig:mesh_comp_cow}, \methodname produces high-quality meshes closely resembling the ground truth.
On the low-noise Replica dataset, Voxblox, FIESTA and VDB-GPDF exhibit voxel-induced block artifacts, while iSDF yields over-smoothed surfaces.
These baselines perform worse on the real-world datasets (Cow\&Lady, Newer College), where sensor noise yields incomplete or distorted meshes: the discrete grids of FIESTA, Voxblox and VDB-GPDF struggle with fine details and completeness, while iSDF lacks geometric details.

Quantitative results in Table \ref{table:mesh_metrics} confirm that \methodname achieves state-of-the-art or competitive performance across nearly all metrics and datasets, consistently outperforming the baselines by mitigating sensor noise while preserving fine geometric details.
Fine near-surface details come from extracting the mesh via marching cubes on the continuous BHM occupancy field, which recovers the surface at sub-voxel resolution and avoids the voxel-induced artifacts of the grid-based baselines.
On the low-noise Replica scenes, Voxblox attains lower Accuracy because its TSDF mesh is extracted at the exact zero-crossing, whereas our marching-cubes surface sits at an online-estimated, view-dependent log-odds level $\tau$ (Sec.~\ref{sec:frontend}); the gap averages only $0.43\,$cm and reverses on the larger, noisier Newer College scene, where \methodname is the most accurate (Table~\ref{table:mesh_metrics}).

\vspace{-1em}
\subsection{SDF Accuracy}
\label{sec:sdf_accuracy}

We evaluate SDF and gradient accuracy using Mean Absolute Error (MAE) across three regions: \textit{All} points, \textit{Near} ($\le 0.2$m from surface), and \textit{Far} ($>0.2$m).
To isolate SDF accuracy from occupancy misclassification (e.g., by FIESTA), we compare the absolute values of predicted and g.t. SDFs.
As summarized in Table \ref{table:sdf_metrics}, \methodname consistently achieves the lowest MAE for both SDF values and gradients across nearly all datasets and regions.
FIESTA is competitive on Replica but degrades significantly on Cow\&Lady and Newer College due to sensor noise.
Voxblox and iSDF exhibit higher errors due to voxelization and over-smoothing, respectively.
Notably, while VDB-GPDF outperforms Voxblox in SDF accuracy, its gradient MAE is the highest among all methods.

\methodname's superior performance in the \textit{Near} region supports manipulation and navigation, while its accurate long-range predictions and gradients facilitate path planning and optimization-based control.
In addition to SDF values, we evaluate the predicted sign (i.e. occupancy) against the ground truth on Replica and summarize the key findings here.
Far from the surface, the sign is clear and every method exceeds $99\%$, so only the near-surface region is discriminating. Taking free space as the positive class, precision is important because it corresponds to low false-free (collision) rate. \methodname showed the highest near-surface precision, ranking first in 5 of 8 scenes and staying above $99\%$ everywhere, whereas VDB-GPDF showed $69$--$82\%$ precision, labeling nearly all near-surface points as free, yielding the highest false-free rate among all methods. The BHM sign prediction is therefore accurate and, importantly, conservative: it rarely mislabels occupied space as free, supporting collision avoidance and justifying the deterministic-sign treatment in the back-end.

\begin{table*}[htbp]
\centering
\caption{\small SDF reconstruction metrics on Replica, Cow\&Lady (C.L. for the scene and Cow for cow only), and Newer College (N.C.).}
\label{table:sdf_metrics}
\scriptsize
\def\sdfmaecell{\multirow{15}{*}{\begin{tabular}{@{}c@{}}SDF MAE\\ $[\text{cm}] \downarrow$\end{tabular}}   }
\def\gradmaecell{\multirow{15}{*}{\begin{tabular}{@{}c@{}}Grad. MAE\\ $[\text{rad}] \downarrow$\end{tabular}} }
\def\allcell{\multirow{5}{*}{All}}
\def\nearcell{\multirow{5}{*}{Near}}
\def\farcell{\multirow{5}{*}{Far}}
\begin{tabular}{ll|c|ccccccccccc}
    \hline
    Metric       & Region    & Method   & room 0            & room 1            & room 2            & office 0          & office 1          & office 2          & office 3          & office 4          & C.L.              & Cow               & N.C.               \\
    \hline
    \sdfmaecell   &  \allcell   &  Ours      &  \textbf{1.691}  &  \textbf{1.548}  &  \textbf{1.638}  &  \textbf{1.576}  &  \textbf{1.562}  &  \textbf{1.712}  &  \textbf{1.595}  &  \textbf{1.699}  &  \textbf{4.628}  &  \textbf{2.367}  &  \underline{22.989} \\
                  &             &  FIESTA    &  \underline{2.175}  &  \underline{2.364}  &  \underline{2.418}  &  \underline{2.662}  &  \underline{2.461}  &  \underline{2.234}  &  \underline{2.344}  &  \underline{2.378}  &  15.344             &  15.092             &  37.244 \\
                  &             &  Voxblox   &  3.128              &  2.539              &  2.733              &  3.017              &  2.725              &  3.470              &  3.740              &  3.077              &  6.297              &  4.000              &  32.030 \\
                  &             &  iSDF      &  3.745              &  4.230              &  4.259              &  3.890              &  4.226              &  5.094              &  4.245              &  4.147              &  7.327              &  4.258              &  64.304 \\
                  &             &  VDB-GPDF  &  2.925              &  2.900              &  2.778              &  2.902              &  2.968              &  2.861              &  3.176              &  2.929              &  \underline{4.683}  &  \underline{3.056}  &  \textbf{21.783} \\
    \hhline{|~|-|-|-|-|-|-|-|-|-|-|-|-|-|}
                  &  \nearcell  &  Ours      &  \textbf{1.741}  &  \textbf{1.706}  &  \textbf{1.777}  &  \textbf{1.665}  &  \textbf{1.624}  &  \textbf{1.755}  &  \textbf{1.697}  &  \textbf{1.809}  &  \textbf{3.572}  &  \textbf{2.442}  &  \underline{8.813} \\
                  &             &  FIESTA    &  \underline{2.121}  &  \underline{2.046}  &  \underline{2.040}  &  2.233              &  \underline{1.926}  &  \underline{1.848}  &  \underline{1.974}  &  \underline{2.077}  &  10.591             &  13.582             &  32.209 \\
                  &             &  Voxblox   &  2.780              &  2.153              &  2.233              &  \underline{2.196}  &  2.172              &  2.852              &  3.409              &  2.959              &  4.287              &  3.404              &  20.186 \\
                  &             &  iSDF      &  3.649              &  3.632              &  3.295              &  3.206              &  3.646              &  3.821              &  3.397              &  3.301              &  4.258              &  3.595              &  \textbf{6.540} \\
                  &             &  VDB-GPDF  &  2.847              &  2.924              &  2.802              &  2.827              &  2.936              &  2.770              &  3.136              &  2.870              &  \underline{3.938}  &  \underline{3.025}  &  20.693 \\
    \hhline{|~|-|-|-|-|-|-|-|-|-|-|-|-|-|}
                  &  \farcell   &  Ours      &  \textbf{1.658}  &  \textbf{1.431}  &  \textbf{1.543}  &  \textbf{1.505}  &  \textbf{1.504}  &  \textbf{1.683}  &  \textbf{1.520}  &  \textbf{1.630}  &  \textbf{5.157}  &  \textbf{2.306}  &  \underline{23.899} \\
                  &             &  FIESTA    &  \underline{2.205}  &  \underline{2.574}  &  2.656              &  \underline{2.961}  &  \underline{2.898}  &  \underline{2.461}  &  \underline{2.563}  &  \underline{2.545}  &  17.753             &  16.331             &  37.744 \\
                  &             &  Voxblox   &  3.368              &  2.845              &  3.081              &  3.711              &  3.270              &  3.918              &  4.002              &  3.158              &  7.338              &  4.471              &  33.267 \\
                  &             &  iSDF      &  3.811              &  4.705              &  4.932              &  4.466              &  4.797              &  6.016              &  4.918              &  4.718              &  8.864              &  4.802              &  69.652 \\
                  &             &  VDB-GPDF  &  3.486              &  2.620              &  \underline{2.606}  &  3.778              &  3.585              &  3.472              &  3.511              &  3.320              &  \underline{7.066}  &  \underline{3.272}  &  \textbf{22.739} \\
    \hline
    \gradmaecell  &  \allcell   &  Ours      &  \textbf{0.151}  &  \textbf{0.134}  &  \textbf{0.157}  &  \textbf{0.146}  &  \textbf{0.140}     &  \textbf{0.165}  &  \textbf{0.159}  &  \textbf{0.149}  &  \underline{0.498}  &  \textbf{0.297}  &  \textbf{0.299} \\
                  &             &  FIESTA    &  \underline{0.220}  &  0.207              &  \underline{0.184}  &  \underline{0.216}  &  0.250              &  \underline{0.198}  &  \underline{0.224}  &  \underline{0.180}  &  1.141              &  1.093              &  0.498 \\
                  &             &  Voxblox   &  0.230              &  \underline{0.195}  &  0.222              &  0.251              &  \underline{0.190}  &  0.271              &  0.279              &  0.232              &  \textbf{0.477}     &  \underline{0.338}  &  \underline{0.428} \\
                  &             &  iSDF      &  0.358              &  0.200              &  0.339              &  0.276              &  0.279              &  0.286              &  0.292              &  0.346              &  0.569              &  0.400              &  0.667 \\
                  &             &  VDB-GPDF  &  0.998              &  1.051              &  1.022              &  1.050              &  1.165              &  0.998              &  1.032              &  0.980              &  1.066              &  0.982              &  1.176 \\
    \hhline{|~|-|-|-|-|-|-|-|-|-|-|-|-|-|}
                  &  \nearcell  &  Ours      &  \textbf{0.183}  &  \underline{0.177}  &  \textbf{0.174}     &  \textbf{0.201}     &  \textbf{0.179}  &  \textbf{0.195}  &  \textbf{0.199}  &  \textbf{0.178}  &  \underline{0.680}  &  \underline{0.405}  &  \textbf{0.948} \\
                  &             &  FIESTA    &  0.304              &  0.276              &  0.230              &  0.313              &  0.323              &  0.253              &  0.300              &  \underline{0.232}  &  1.329              &  1.224              &  1.398 \\
                  &             &  Voxblox   &  \underline{0.282}  &  0.183              &  \underline{0.204}  &  0.259              &  \textbf{0.179}     &  0.301              &  0.336              &  0.299              &  \textbf{0.679}     &  0.466              &  1.342 \\
                  &             &  iSDF      &  0.290              &  \textbf{0.176}     &  0.247              &  \underline{0.229}  &  \underline{0.206}  &  \underline{0.222}  &  \underline{0.257}  &  0.277              &  \textbf{0.679}     &  \textbf{0.395}     &  \underline{1.309} \\
                  &             &  VDB-GPDF  &  1.070              &  1.101              &  1.088              &  1.099              &  1.193              &  1.075              &  1.093              &  1.057              &  1.116              &  1.030              &  1.413 \\
    \hhline{|~|-|-|-|-|-|-|-|-|-|-|-|-|-|}
                  &  \farcell   &  Ours      &  \textbf{0.138}  &  \textbf{0.111}  &  \textbf{0.148}  &  \textbf{0.117}  &  \textbf{0.116}  &  \textbf{0.151}  &  \textbf{0.140}  &  \textbf{0.138}  &  \underline{0.433}  &  \textbf{0.229}  &  \textbf{0.275} \\
                  &             &  FIESTA    &  \underline{0.182}  &  \underline{0.171}  &  \underline{0.162}  &  \underline{0.166}  &  0.203              &  \underline{0.171}  &  \underline{0.189}  &  \underline{0.158}  &  1.050              &  0.985              &  0.412 \\
                  &             &  Voxblox   &  0.208              &  0.201              &  0.230              &  0.246              &  \underline{0.198}  &  0.257              &  0.256              &  0.205              &  \textbf{0.406}     &  \underline{0.258}  &  \underline{0.351} \\
                  &             &  iSDF      &  0.389              &  0.214              &  0.389              &  0.303              &  0.330              &  0.319              &  0.309              &  0.377              &  0.528              &  0.403              &  0.632 \\
                  &             &  VDB-GPDF  &  0.613              &  0.622              &  0.627              &  0.643              &  0.762              &  0.603              &  0.684              &  0.590              &  0.907              &  0.644              &  0.968 \\
    \hline
\end{tabular}
\vspace{-0.75em}
\end{table*}

\vspace{-1em}
\subsection{Computation Efficiency}
\label{sec:computation_efficiency}

We evaluate efficiency by measuring the average per-frame SDF update time and the prediction latency for 1k query positions. As shown in Table \ref{table:time_metrics}, \methodname consistently achieves real-time performance with an average update time of approximately 150 ms across scene scales.
While Voxblox builds the TSDF rapidly, its runtime is dominated by BFS-based SDF integration, slowed by the high volume of voxel updates from the large cutoff required for accurate SDF estimation.
FIESTA achieves the fastest integration but, as shown in Sec. \ref{sec:reconstruction_accuracy} and \ref{sec:sdf_accuracy}, is inaccurate in noisy environments.
iSDF remains slow due to its multi-iteration convergence and analytical gradient backpropagation overhead.
Although VDB-GPDF also uses log-GP, its reliance on VDB structures for spatial partitioning yields significantly higher update latencies.
Overall, \methodname balances computational speed and mapping precision, providing a robust solution for real-time robotic applications.

\begin{table}[htbp]
\centering
\caption{\small Timing metrics: average per-frame processing time (FPT) and 1k-point query time (QT-1k) on Replica \texttt{room0} \cite{replica19arxiv}, Cow\&Lady \cite{oleynikova_voxblox_2017}, and Newer College \cite{newercollege2021}. Best and second best are bold and underlined. Measured on an Intel i9-14900K and NVIDIA RTX 3090.}
\label{table:time_metrics}
\def\replicacell{\multicolumn{2}{c|}{Replica}}
\def\cowandladycell{\multicolumn{2}{c|}{Cow \& Lady}}
\def\newercollegecell{\multicolumn{2}{c}{Newer College}}
\resizebox{\linewidth}{!}{\begin{tabular}{l|cc|cc|cc}
\hline
\multirow{2}{*}{Method} & \replicacell     & \cowandladycell  & \newercollegecell                                                          \\
						& FPT (s)          & QT-1k (ms)       & FPT (s)           & QT-1k (ms)       & FPT (s)          & QT-1k (ms)       \\
\hline
\methodname             & \underline{0.19} & 2.80             & 0.11              & 4.43             & \textbf{0.17}    & 3.04             \\
VDB-GPDF                & 0.68             & 10.66            & \underline{0.10}  & 8.29             & \underline{0.26} & 10.16            \\
Voxblox                 & 21.23            & 103.81           & 4.48              & 99.42            & 94.01            & 85.20            \\
iSDF                    & 2.49             & \underline{0.29} & 0.84              & \underline{0.27} & 1.38             & \underline{0.28} \\
FIESTA                  & \textbf{0.05}    & \textbf{0.05}    & \textbf{0.02}     & \textbf{0.03}    & 0.28             & \textbf{0.04}    \\
\hline
\end{tabular}
}
\vspace{-0.75em}
\end{table}

\vspace{-1em}
\subsection{Scalability}
\label{sec:scalability}

\begin{figure}[t]
    \centering
    \begin{subfigure}[t]{0.48\linewidth}
        \centering
        \includegraphics[width=\linewidth]{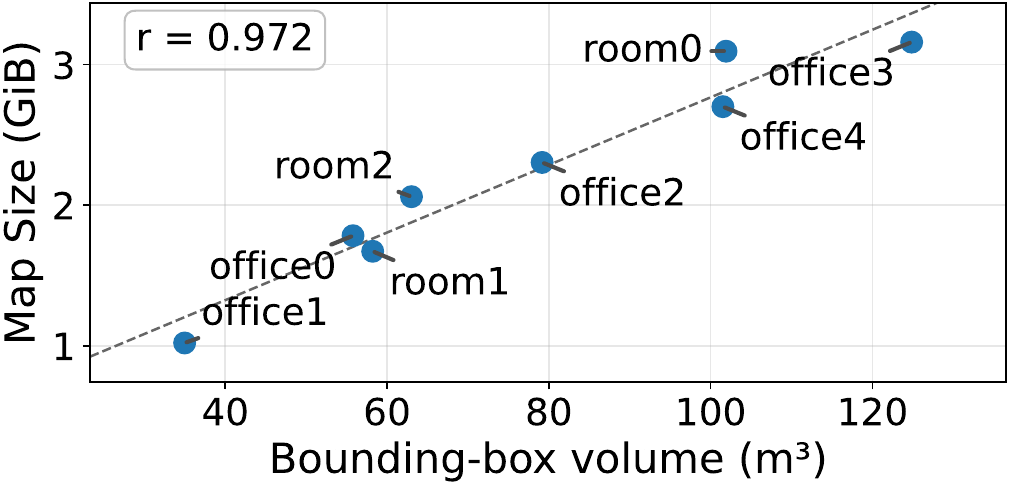}
        \caption{Map size vs. scene volume}
        \label{fig:scalability:mapsize}
    \end{subfigure}%
    \hfill%
    \begin{subfigure}[t]{0.48\linewidth}
        \centering
        \includegraphics[width=\linewidth]{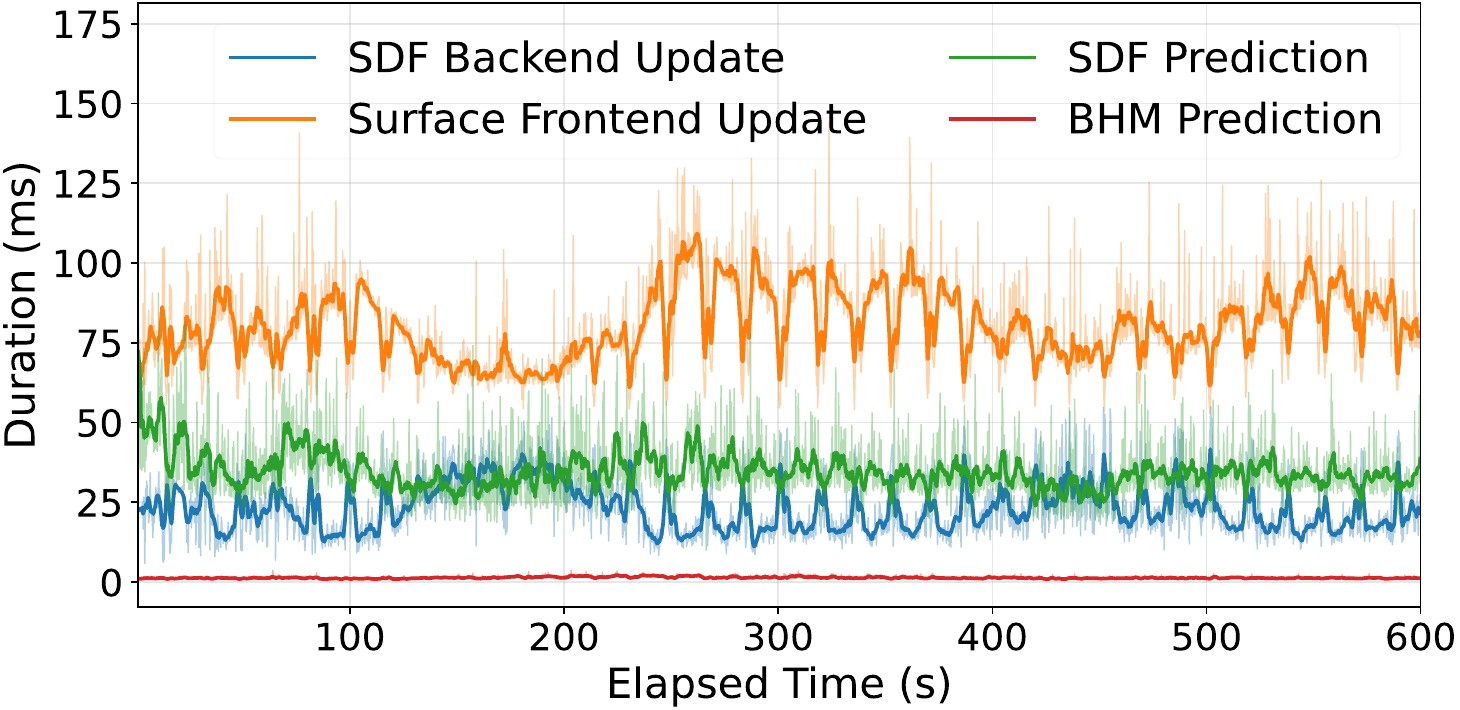}
        \caption{Runtime: Replica room0}
        \label{fig:scalability:timing_room0}
    \end{subfigure}
    \hfill%
    \begin{subfigure}[t]{0.48\linewidth}
        \centering
        \includegraphics[width=\linewidth]{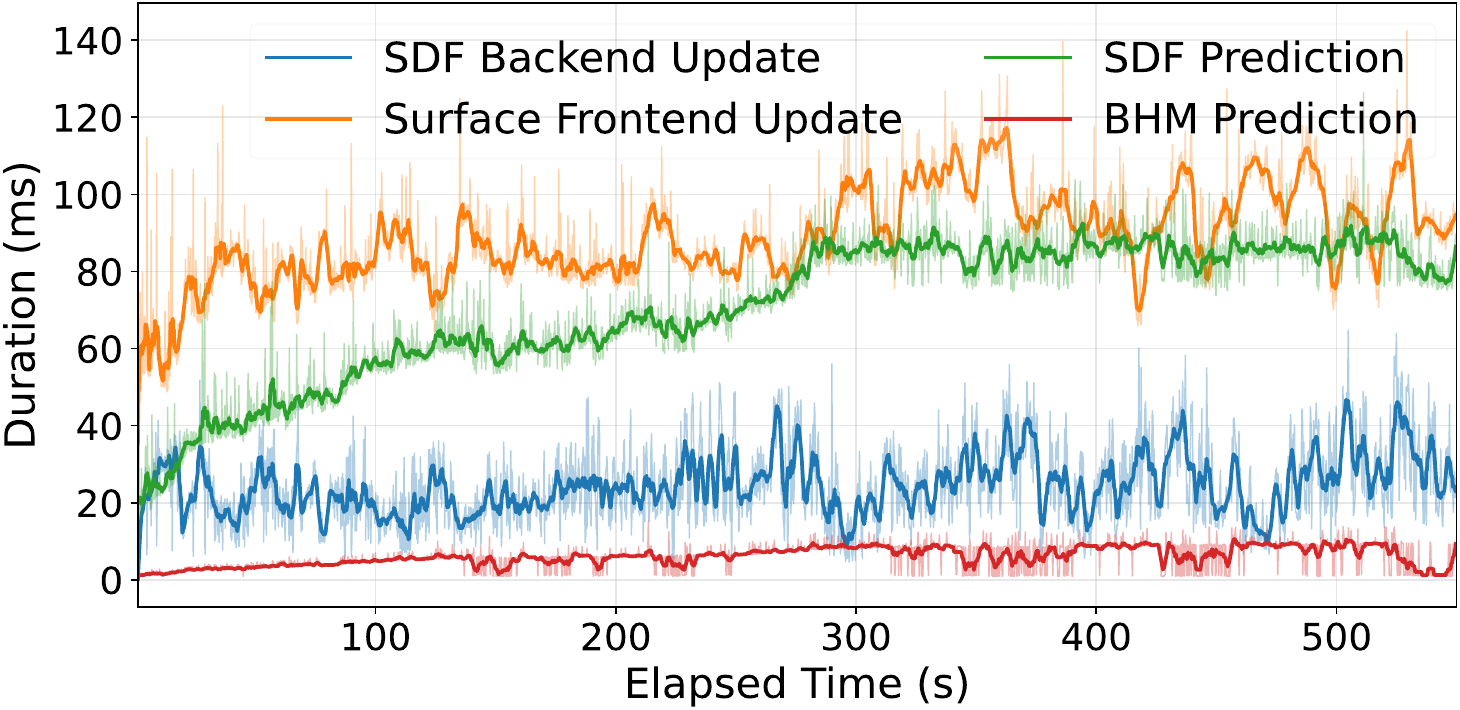}
        \caption{Runtime: Cow\&Lady}
        \label{fig:scalability:timing_cow_and_lady}
    \end{subfigure}%
    \hfill%
    \begin{subfigure}[t]{0.48\linewidth}
        \centering
        \includegraphics[width=\linewidth]{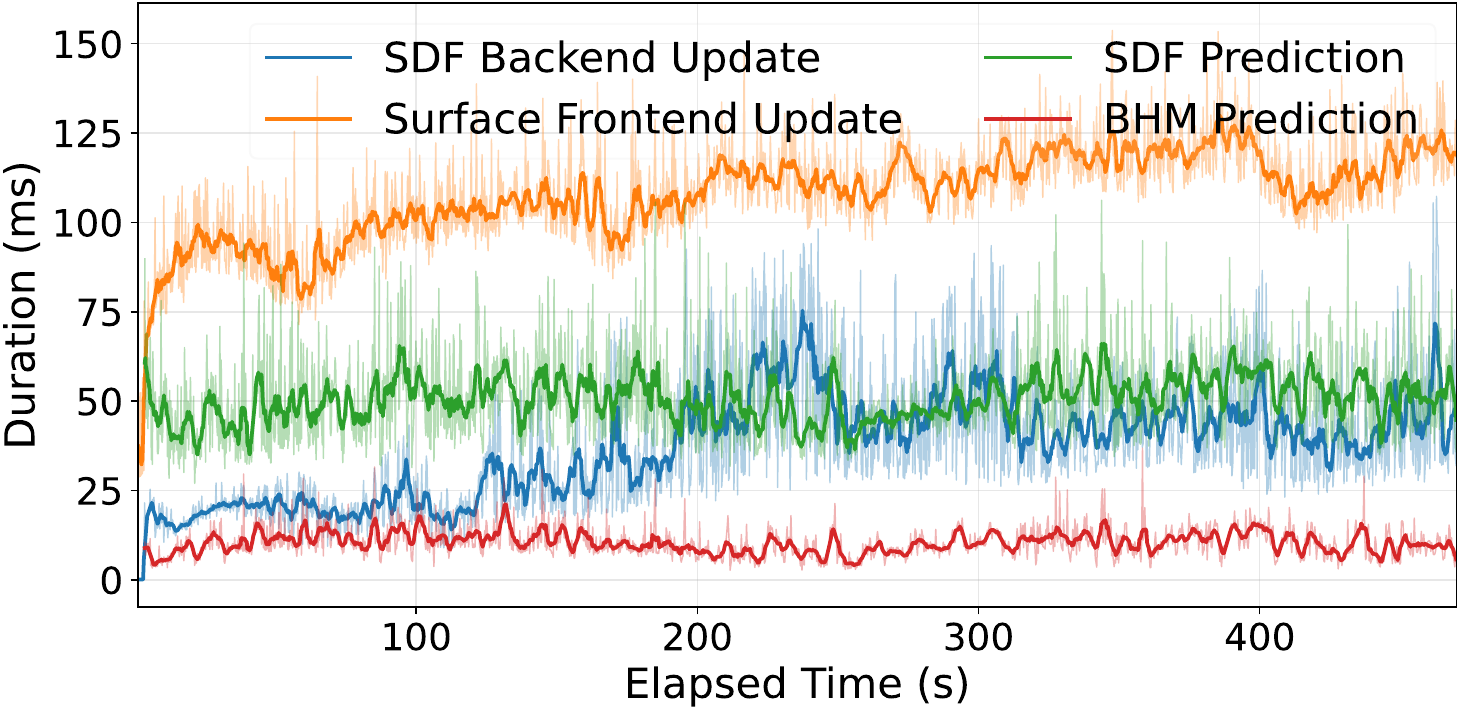}
        \caption{Runtime: Newer College}
        \label{fig:scalability:timing_newer_college}
    \end{subfigure}
    \caption{\small \edit{Scalability of \methodname. (a) The serialized map size scales linearly with the environment volume across the eight Replica scenes ($r=0.972$). (b)--(d) Per-component processing time over the full session (BHM front-end, SDF back-end, SDF prediction, and BHM prediction); all components stay stable with increasing map size and are independent of the global environment size.}}
    \label{fig:scalability}
    \vspace{-1.5em}
\end{figure}

The persistent map size scales linearly with the scene's spatial extent (Fig.~\ref{fig:scalability:mapsize}). Across the eight Replica scenes, the serialized map grows from about 1\,GiB for the smallest ($\approx 35\,$m$^3$) to about 3\,GiB for the largest ($\approx 125$m$^3$), strongly correlated with the bounding-box volume ($r=0.972$). This footprint is set by the number of stored GP surface samples (observed surface area times sampling density), so it grows with the observed extent rather than without bound.
\edit{On Newer College, the serialized map reaches about 6\,GiB. For memory-constrained onboard deployment in large environments, the memory use can be reduced by using a coarser map resolution or reducing the density of surface samples estimated by the front-end.}
The per-component update and prediction times (Fig.~\ref{fig:scalability:timing_room0}--\ref{fig:scalability:timing_newer_college}) do not grow with the accumulated map size, since the local octree-organized BHM/GP design ties each incremental update to the newly observed region rather than the global map. This bounded per-update cost lets \methodname scale to large environments in real time.

\vspace{-1em}
\subsection{Kernel Ablation}
\label{sec:kernel_ablation}

\begin{figure}[t]
    \centering
    \begin{subfigure}[t]{0.48\linewidth}
        \centering
        \includegraphics[width=\linewidth]{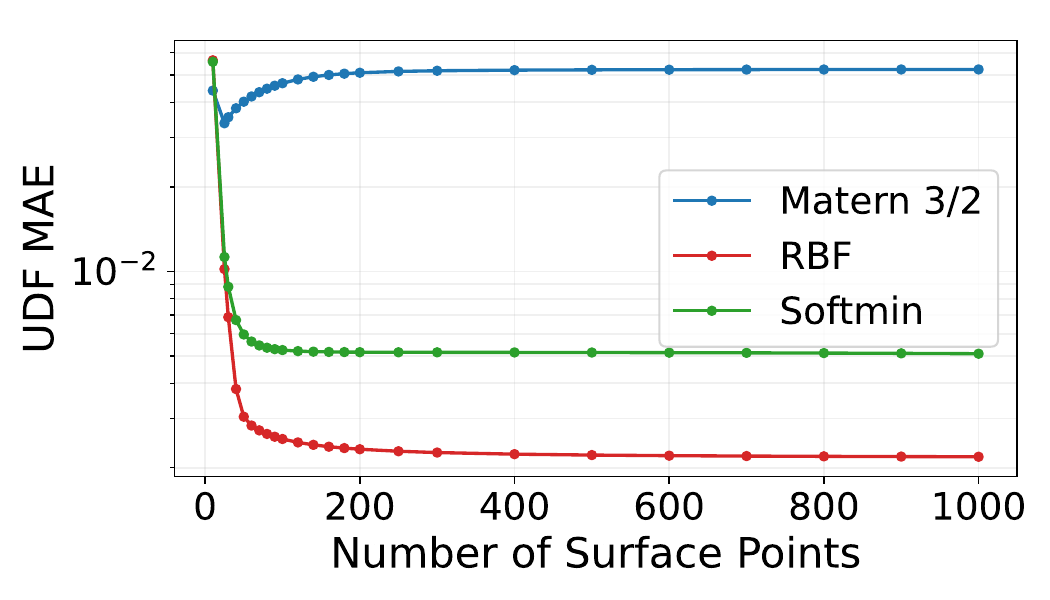}
    \end{subfigure}%
    \hfill%
    \begin{subfigure}[t]{0.48\linewidth}
        \centering
        \includegraphics[width=\linewidth]{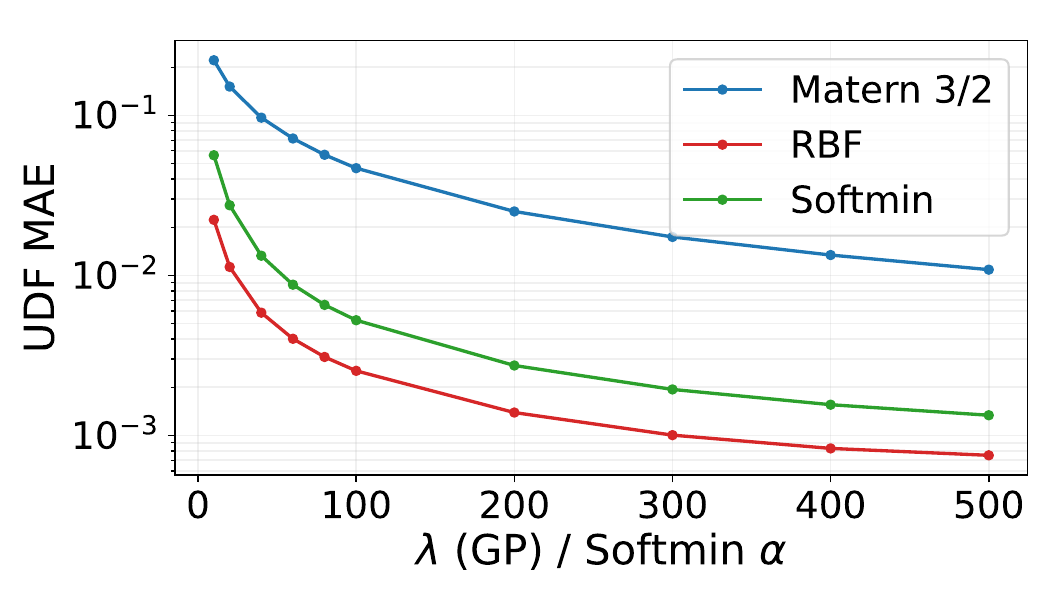}
    \end{subfigure}
    \caption{\small \edit{Kernel ablation on a 2D unit-circle dataset, empirically validating
    the softmin view of the log-GP back-end. UDF MAE vs. number of surface samples (left)
    $N$ and kernel scale $\lambda$ (equivalently softmin $\alpha$) (right). Except for the
    swept variable, defaults are $\lambda=100$, $\alpha=100$, and $N=100$. The RBF kernel
    attains the lowest error and improves with larger scale and denser sampling.}}
    \label{fig:kernel_ablation}
    \vspace{-0.5em}
\end{figure}

We validate the softmin view (Sec.~\ref{sec:kernel_extension}) of the log-GP back-end by sweeping the kernel scale $\lambda$ (equivalently softmin $\alpha$) and the number of surface samples $N$ on a 2D unit-circle dataset (Fig.~\ref{fig:kernel_ablation}).
The UDF error shrinks as $\lambda$ and $N$ grow, matching the limit analysis in \eqref{eq:weak_corr_limit} and \eqref{eq:strong_corr_limit}.
The RBF kernel attains the lowest error and is therefore our default.

\vspace{-1em}
\subsection{Component Analysis}
\label{sec:component_analysis}

\begin{table}[t]
    \centering
    \caption{\edit{\small Back-end ablation: ours vs.\ nearest-neighbor (NN) and softmin fusion, sharing the same BHM front-end and surface samples. SDF MAE for \textit{All} and \textit{Near} query regions, and gradient angular MAE for the \textit{Near} region (Replica is averaged over its 8 scenes).}}
    \label{table:component_ablation:a}
    \scriptsize
    \renewcommand{\arraystretch}{0.85}
    \adjustbox{max width=\linewidth}{
    \begin{tabular}{ll|l|ccc}
        \hline
        Metric & Region & Method & Replica & Cow \& Lady & Newer College \\
        \hline
        \multirow{6}{*}{\makecell{SDF MAE\\ $[\text{cm}]\downarrow$}} & \multirow{3}{*}{All} & Ours (GP) & 1.628 & 4.628 & 22.989 \\
         &  & NN & \textbf{1.622} & \textbf{4.605} & \textbf{22.811} \\
         &  & Softmin & 1.988 & 4.648 & 33.391 \\
        \hhline{|~|-|-|-|-|-|}
         & \multirow{3}{*}{\makecell{Near\\($<0.2$m)}} & Ours (GP) & 1.722 & 3.572 & \textbf{8.813} \\
         &  & NN & \textbf{1.683} & \textbf{3.524} & 8.838 \\
         &  & Softmin & 3.420 & 3.612 & 59.086 \\
        \hline
        \multirow{3}{*}{\makecell{Grad. MAE\\ $[\text{rad}]\downarrow$}} & \multirow{3}{*}{Near} & Ours (GP) & \textbf{0.1858} & \textbf{0.6800} & \textbf{0.9480} \\
         &  & NN & 0.2659 & 0.6982 & 0.9988 \\
         &  & Softmin & 0.4221 & 0.7342 & 1.1575 \\
        \hline
    \end{tabular}
    }
    \vspace{-0.25em}
\end{table}

\begin{table}[t]
    \centering
    \caption{\edit{\small Ablation on the BHM surface-uncertainty weighting (Replica room0, all query points). Ours uses per-sample, occupancy-derived variance $\sigma_i^2$; the alternatives replace it with a single constant variance, either $\sigma^2{=}0$ (pure softmin) or $\sigma^2{=}0.1$.}}
    \label{table:component_ablation:b}
    \begin{tabular}{l|cc}
        \hline
        Weighting & \makecell{SDF MAE $[\text{cm}]\downarrow$} & \makecell{Grad. MAE $[\text{rad}]\downarrow$} \\
        \hline
        Ours (w/ BHM var.)              & \textbf{1.691} & \textbf{0.1514} \\
        w/o BHM var. ($\sigma^2{=}0$)   & 1.731 & 0.1722 \\
        w/o BHM var. ($\sigma^2{=}0.1$) & 1.734 & 0.1749 \\
        \hline
    \end{tabular}
    \vspace{-1em}
\end{table}

\edit{To isolate the back-end's contribution, we replace the GP back-end with nearest-neighbor distance and with a pure softmin, keeping the same BHM front-end and surface samples (Table~\ref{table:component_ablation:a}).
The three back-ends are essentially equivalent in the \textit{All} region: GP and nearest-neighbor SDF MAE differ by less than $0.2\,$cm on every dataset.
Near the surface, however, softmin is far less robust. Its SDF MAE on Newer College is over $6\times$ the GP's ($59.1$ vs.\ $8.8\,$cm) and its Replica gradient error more than doubles ($0.422$ vs.\ $0.186\,$rad).
The GP's advantage is therefore not distance accuracy but differentiability with closed-form gradients, calibrated variance (Sec.~\ref{sec:variance_consistency}), and graceful degradation under noise.

We also evaluate the BHM-derived uncertainty \eqref{eq:surface_point_uncertainty} by replacing the per-sample variance $\sigma_i^2$ with a constant, either $0$ (pure softmin) or $0.1$ (Table~\ref{table:component_ablation:b}).
The BHM-derived weighting improves both the SDF and gradient accuracy, most notably the gradient angle error, which drops from $0.172$ to $0.151\,$rad ($12\%$). This confirms the regularization effect analyzed in Sec.~\ref{sec:backend} that down-weighting samples with high occupancy uncertainty sharpens both the SDF and its gradient.}

\vspace{-1em}
\subsection{Uncertainty Calibration}
\label{sec:variance_consistency}

\begin{figure}[t]
	\centering
	\begin{subfigure}[t]{0.33\linewidth}
		\centering
		\includegraphics[width=\linewidth,trim={10pt 0pt 30pt 0pt},clip]{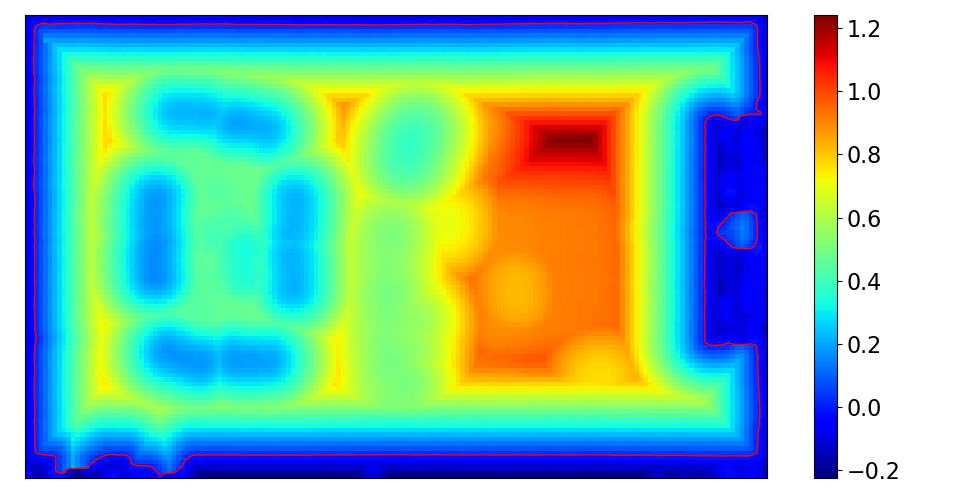}
		\caption{Predicted SDF}
		\label{fig:pred_sdf_slice}
	\end{subfigure}%
	\begin{subfigure}[t]{0.33\linewidth}
		\centering
		\includegraphics[width=\linewidth,trim={10pt 0pt 30pt 0pt},clip]{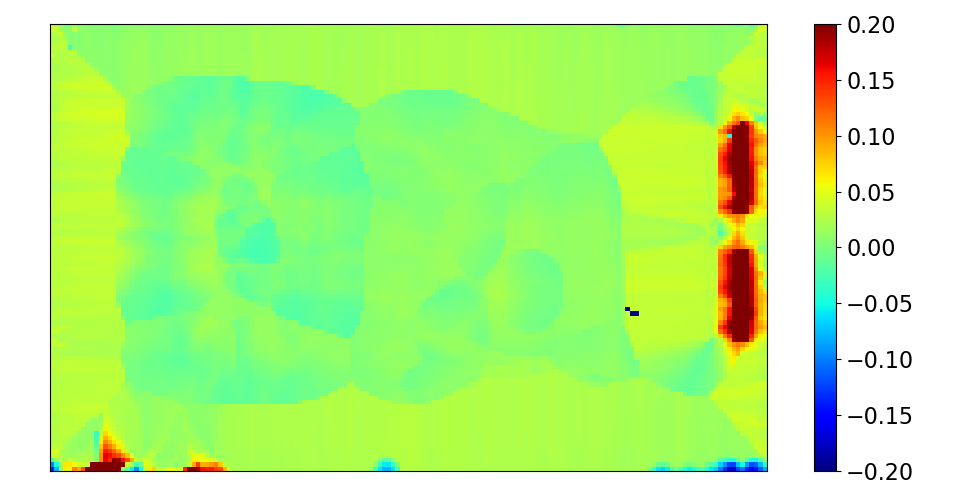}
		\caption{SDF Error}
		\label{fig:sdf_err_slice}
	\end{subfigure}%
	\begin{subfigure}[t]{0.33\linewidth}
		\centering
		\includegraphics[width=\linewidth,trim={10pt 0pt 30pt 0pt},clip]{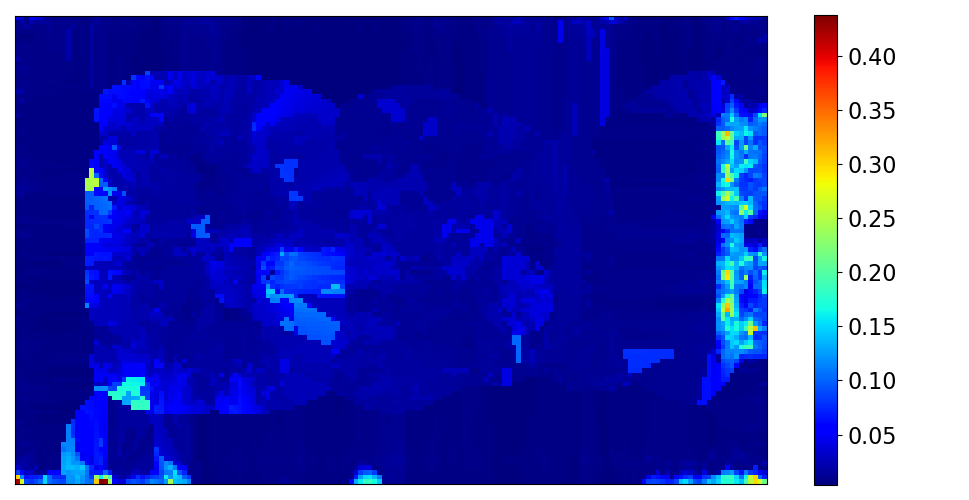}
		\caption{SDF Variance}
		\label{fig:sdf_var_slice}
	\end{subfigure}
	\caption{\small Consistency between SDF error and estimated SDF variance on a slice of Replica \texttt{Office3} \cite{replica19arxiv}: (a) predicted SDF, (b) absolute SDF error against ground truth, and (c) estimated SDF variance. Regions of higher error coincide with higher estimated variance, showing that \methodname effectively captures uncertainty.}
	\label{fig:sdf_var_consistency}

	\begin{subfigure}[t]{0.33\linewidth}
		\centering
		\includegraphics[width=\linewidth]{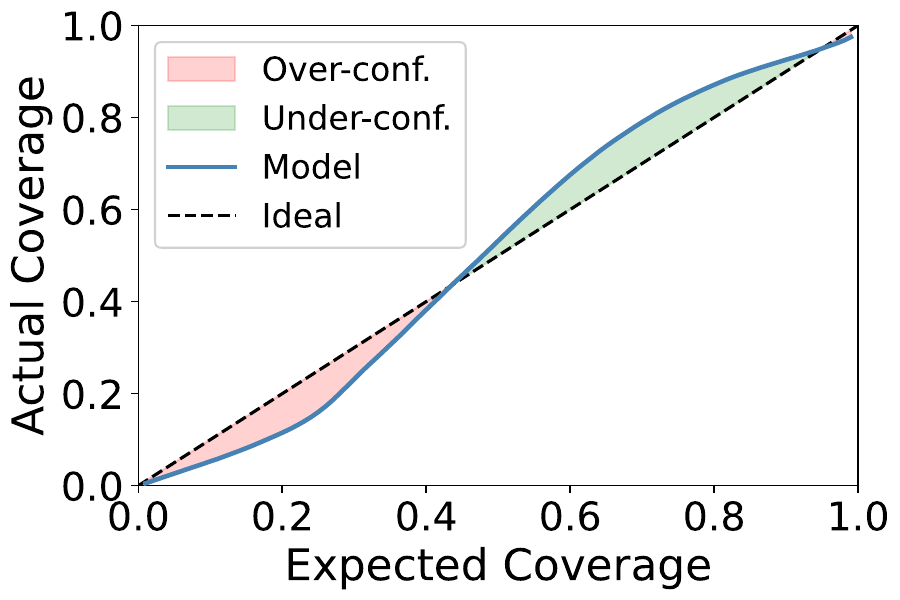}
		\caption{Replica room0}
	\end{subfigure}%
	\begin{subfigure}[t]{0.33\linewidth}
		\centering
		\includegraphics[width=\linewidth]{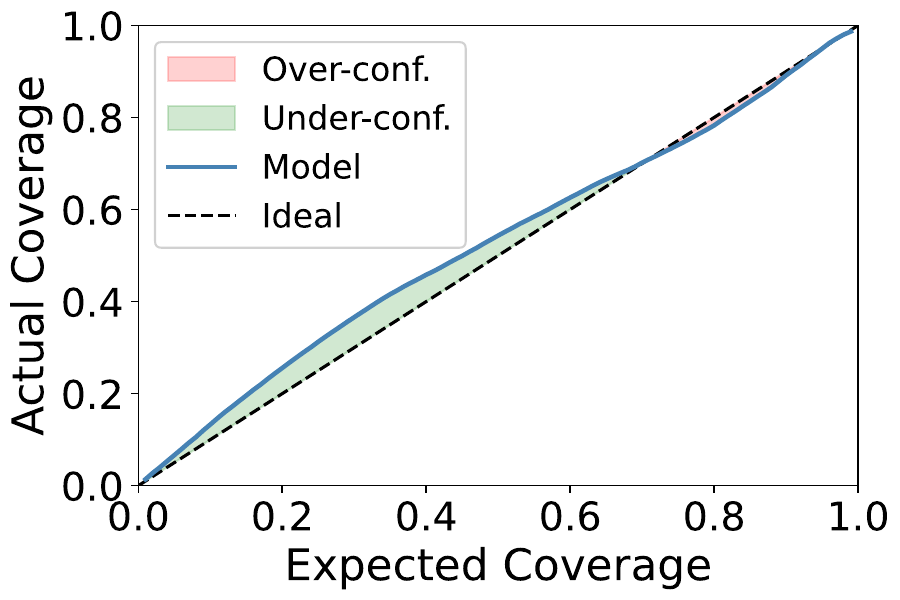}
		\caption{Cow\&Lady}
	\end{subfigure}%
	\begin{subfigure}[t]{0.33\linewidth}
		\centering
		\includegraphics[width=\linewidth]{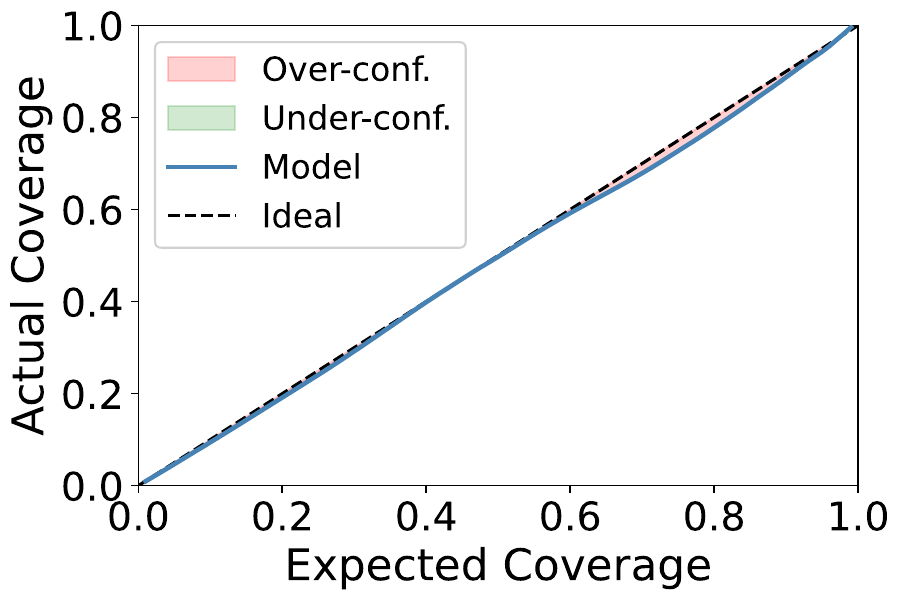}
		\caption{Newer College}
	\end{subfigure}
	\caption{\small Reliability curves of the predicted SDF uncertainty: for each target confidence level (x-axis), the fraction of query points whose error falls within it (y-axis). Empirical coverage closely tracks the ideal diagonal (ECE $0.052$, $0.031$, $0.011$), confirming calibration.}
	\label{fig:variance_calibration}
\end{figure}

A key advantage of \methodname is its calibrated SDF uncertainty. Qualitatively, Fig.~\ref{fig:sdf_var_consistency} shows that on a slice of Replica Office3 \cite{replica19arxiv} regions of high SDF error coincide with regions of high estimated variance.
Quantitatively, we examine the empirical first and second moments of normalized error $z=(u-\hat{u})/\hat{\sigma}$. Good calibration corresponds to $\bbE[z]=0$ and $\bbE[z^2]=1$.
Across Replica room0, Cow\&Lady, and Newer College, $\bbE[z^2]=1.05,1.03,1.04$ (within $5.5\%$ of $1$), and the empirical coverage closely tracks the target with an expected calibration error (ECE) of $0.052,0.031,0.011$ (Fig.~\ref{fig:variance_calibration}), \edit{showing the predicted variance closely matches the true error variance}.
\edit{The first moment is positive on all three scenes ($\bbE[z]=0.49,0.55,0.02$), which is consistent with the softmin perspective (Sec.~\ref{sec:kernel_extension}) that the predicted distance is biased to be conservative. Such a bias is related to the scale $\beta$ in \eqref{eq:surface_point_uncertainty}, the kernel scale $\lambda$ and the surface sample density, which can be tuned to reduce it if desired.}

\vspace{-1em}
\subsection{Ablation Study on Sensor Noise}
\label{sec:ablation_study_sensor_noise}

\begin{figure}
    \centering
    \begin{subfigure}[t]{0.48\linewidth}
        \centering
        \includegraphics[width=\linewidth]{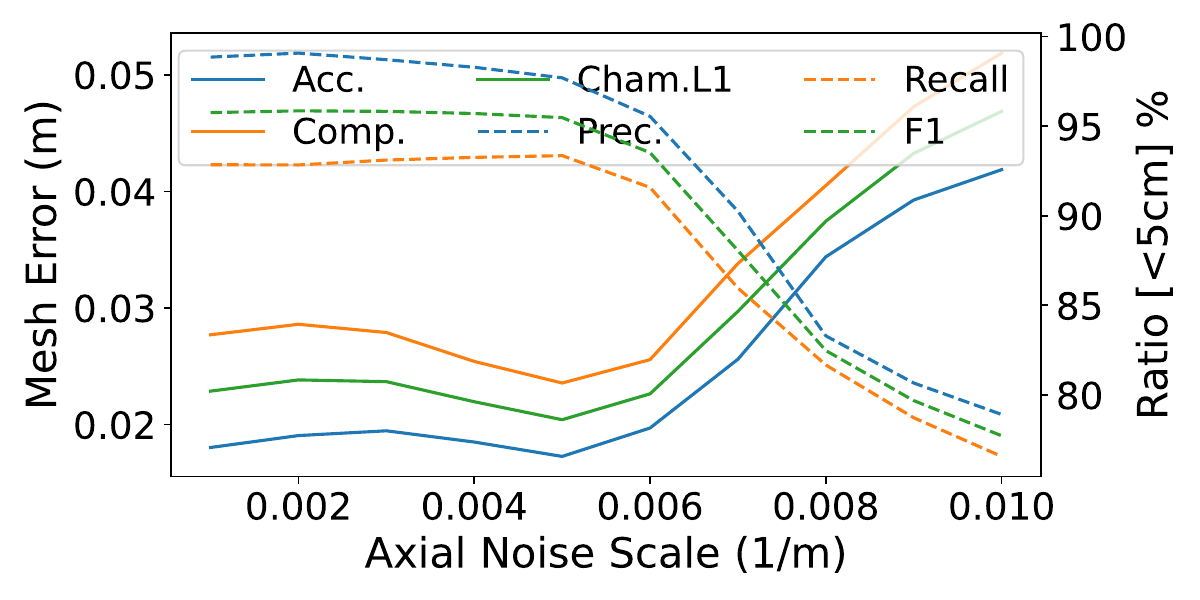}
    \end{subfigure}%
    \hfill%
    \begin{subfigure}[t]{0.48\linewidth}
        \centering
        \includegraphics[width=\linewidth]{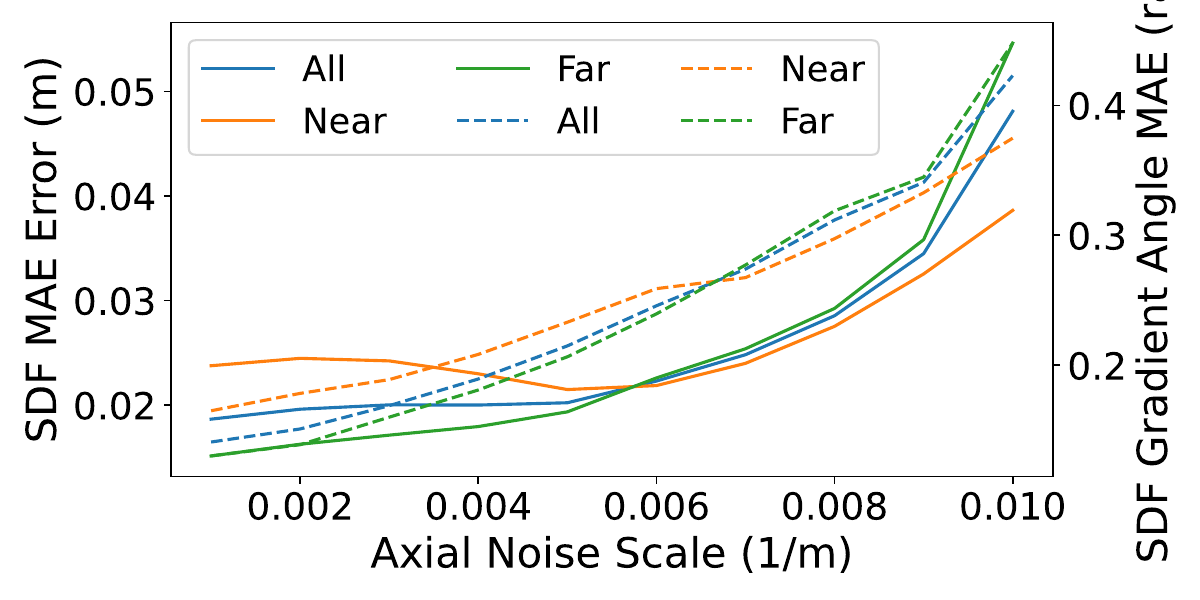}
    \end{subfigure}
    \caption{\small Effect of sensor noise on (left) mesh reconstruction and (right) SDF metrics on Replica \texttt{room0}~\cite{replica19arxiv}. \methodname stays robust as noise increases, with only gradual degradation.}
    \label{fig:ablation_sensor_noise}
    \vspace{-1em}
\end{figure}

To assess noise robustness, we evaluated \methodname on Replica \texttt{room0} by varying the axial noise parameter $k$ in $\sigma=kz^2$ with other parameters fixed.
Fig. \ref{fig:ablation_sensor_noise} shows the resulting mesh and SDF metrics. \methodname degrades only gradually as noise increases, confirming that the BHM front-end mitigates sensor noise to maintain reliable SDF mapping in adverse environments.

\vspace{-1em}
\subsection{Autonomous Robot Navigation Demonstration}
\label{sec:path_planning_demo}

\begin{figure}
    \centering
    \begin{subfigure}[t]{0.48\linewidth}
        \centering
        \includegraphics[width=\linewidth]{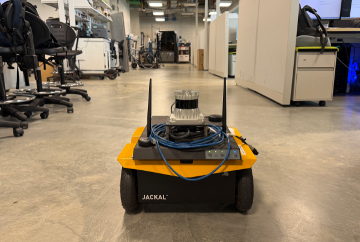}
        \caption{Jackal robot}
        \label{fig:robot_in_env}
    \end{subfigure}%
    \hfill%
    \begin{subfigure}[t]{0.48\linewidth}
        \centering
        \includegraphics[width=\linewidth]{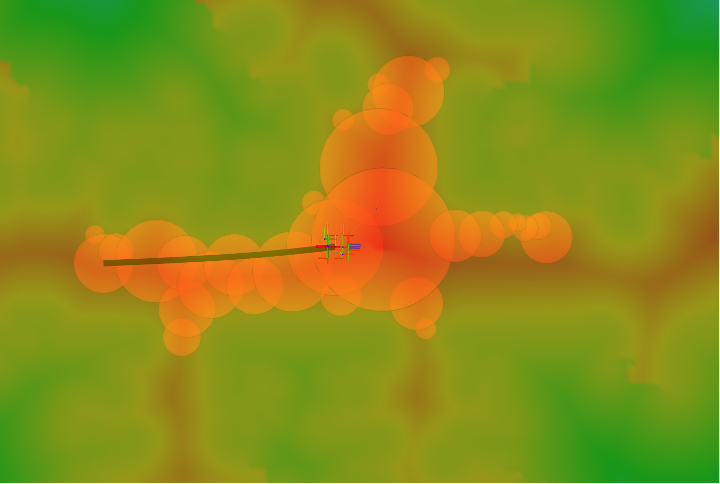}
        \caption{SDF and safe bubble cover}
        \label{fig:real_robot_sdf_demo}
    \end{subfigure}
	\caption{\small \methodname enables safe bubble cover construction \cite{lee_safe_2024} and safe robot navigation. (a) A Clearpath Jackal robot with LiDAR sensor navigating in a lab environment. (b) Real-time \methodname estimate shown as a heatmap with safe bubble cover (union of orange circles) and planned trajectory (black line).}
	\label{fig:bubble_experiment}
    \vspace{-0.5em}
\end{figure}

Real-time, uncertainty-quantified SDF mapping is essential for safe robot navigation. We demonstrate this by planning a safe corridor in the SDF map from a chain of overlapping bubbles, using the sampling-based approach of \cite{lee_safe_2024}.
The planner builds a rapidly exploring random tree from free-space samples with associated safe bubbles, whose radii are the SDF values minus a safety margin, optionally scaled by the estimated SDF uncertainty.
As shown in Fig. \ref{fig:bubble_experiment}, \methodname allows the planner to generate a connected chain of bubbles spanning the start and goal, keeping the path strictly within the mapped free space. This demonstrates \methodname's utility in providing accurate distance and uncertainty information for risk-aware motion planning.

\section{Conclusion}
\label{sec:conclusion}

We presented \methodname, a real-time kernel regression framework for differentiable SDF learning with uncertainty quantification. By coupling BHMs for robust surface estimation with GPs for distance prediction, our method achieves superior reconstruction accuracy and efficiency. The hierarchical tree structure ensures scalability to large environments, while our softmin-based uncertainty quantification enables risk-aware decision-making for downstream robotic tasks.
Extensive evaluations on synthetic and real-world datasets show that \methodname outperforms state-of-the-art baselines in both quality and speed, while its BHM front-end mitigates sensor noise and adapts to dynamic environments, providing a robust, practical solution for real-time robotic applications.

\balance
{\small\putbib}
\end{bibunit}

\setcounter{section}{0}
\setcounter{equation}{0}
\setcounter{figure}{0}
\setcounter{table}{0}
\renewcommand{\thesection}{\Alph{section}}                         
\renewcommand{\thesubsection}{\thesection.\arabic{subsection}}     
\renewcommand{\thesubsectiondis}{\thesection.\arabic{subsection}\hskip 0.5em} 
\renewcommand{\thesubsubsection}{\thesubsection.\arabic{subsubsection}}
\renewcommand{\thesubsubsectiondis}{\thesubsection.\arabic{subsubsection}\hskip 0.5em}
\renewcommand{\theequation}{\thesection.\arabic{equation}}
\renewcommand{\thefigure}{\thesection.\arabic{figure}}
\renewcommand{\thetable}{\thesection.\arabic{table}}
\makeatletter
\@addtoreset{equation}{section}
\@addtoreset{figure}{section}
\@addtoreset{table}{section}
\makeatother

\begin{bibunit}
\nocite{IEEEexample:BSTcontrol}
\twocolumn[%
\begin{center}
    {\LARGE Appendix of \methodname: An Open-Source Library for Real-Time Signed Distance Function Estimation using Kernel Regression\par}
    \vspace{1em}
    {\large Zhirui Dai \qquad Tianxing Fan \qquad Mani Amani \qquad Jaemin Seo\\
    Ki Myung Brian Lee \qquad Hyondong Oh \qquad Nikolay Atanasov\par}
    \vspace{2em}
\end{center}
]%
\let\ksdfsavenewpage\newpage
\def\newpage{\let\newpage\ksdfsavenewpage}

This appendix is provided as a separate document with our code release.
To obtain the latest information, please refer to our GitHub repository: \url{https://github.com/ExistentialRobotics/kernel_sdf}.

\label{sec:appendix}

\section{EM Algorithm for Updating BHM}
\label{sec:app:bhm_em_algorithm}

Readers can refer to \cite{bayesianlr1997,bhm2017} for more details about the Bayesian Hilbert Map (BHM). Here, we provide a brief derivation of the EM algorithm for updating the BHM model parameters, $\bfmu$ and $\bfSigma$.

BHM represents a continuous occupancy field $P(y \mid\bfx)$ by using a set of $M$ hinge points $\tilde{\bfX}$ with weights $\bfw$ to do kernel regression in the Hilbert space:
\begin{equation}
    P(y \mid \bfx, \bfw) = \sigma\big((2y-1) \bfw^\top \bfphi(\bfx) \big),
    \label{eq:bhm_logistic_regression_model}
\end{equation}
where $\bfphi(\bfx) = [k(\bfx,\tilde{\bfx}_1), \cdots, k(\bfx, \tilde{\bfx}_M), 1]^\top$ is a vector of kernel values, $k(\cdot, \cdot)$ is RBF kernel, and $\sigma$ is the sigmoid function.

Given a dataset $\calD = \{(\bfx_k,y_k)\}_{k=1}^{K}$ of $K$ generated samples at time $t$, our goal is to update the weights $\bfw$ of the model such that we get the posterior distribution:
\begin{equation}
    P(\bfw|\calD) = \frac{P(\calD|\bfw)P(\bfw)}{P(\calD)},
\end{equation}
where $P(\bfw)$ is the prior distribution of the weights, and the likelihood is given by:
\begin{equation}
    \begin{aligned}
        P(\calD|\bfw) & = \prod_{k=1}^{K} P(y_k|\bfx_k,\bfw)                                             \\
                      & = \prod_{k=1}^{K} \sigma(-y_k\bfw^\top \bfphi(\bfx_k)), \quad y_k \in \{-1, 1\}.
    \end{aligned}
\end{equation}

The marginal likelihood $P(\calD)$ is intractable, so we resort to approximate inference methods.
Our goal to maximize the log marginal likelihood:
\begin{equation}
    \begin{aligned}
        \log P(\calD) & = \log \int P(\calD|\bfw)P(\bfw) d\bfw                             \\
                      & = \log \int Q(\bfw) \frac{P(\calD|\bfw)P(\bfw)}{Q(\bfw)} d\bfw     \\
                      & \geq \int Q(\bfw) \log \frac{P(\calD|\bfw)P(\bfw)}{Q(\bfw)} d\bfw, \\
                      & \uparrow \text{Jensen's inequality}
    \end{aligned}
\end{equation}
where $Q(\bfw)$ is an approximate posterior distribution of the weights.

Since $P(\calD|\bfw)$ is product of sigmoids that contains the weights $\bfw$ inside,
we use the variational bound \cite{bayesianlr1997} to get a looser lower bound:
$$\sigma(r) \geq \sigma(\xi) \exp\left(\frac{r - \xi}{2} +\lambda\left(\xi\right) \left(r^2-\xi^2\right) \right),$$
where $\xi$ is a variational parameter and $\lambda(\xi) = \frac{1}{2\xi}\left(\frac{1}{2}-\sigma(\xi)\right)$.

Using the idea of Sequential BHM EM update \cite{bhm2017}, we assume a Gaussian prior $P(\bfw) = \calN(\bfw|\bfmu_{t-1},\bfSigma_{t-1})$ and a Gaussian approximate posterior $Q(\bfw) = \calN(\bfw|\bfmu_t,\bfSigma_t)$ for the weights.
Hence, we can write the variational lower bound of the log marginal likelihood as:
$$
\begin{aligned}
& \calL = \bbE_{Q} \left[\log P(\calD|\bfw)\right] + \bbE_{Q} \left[\log P(\bfw)\right] - \bbE_{Q} \left[\log Q(\bfw)\right] \\
& \geq \sum_{k=1}^{K} \bbE_{Q} \left[\log \sigma(-y_k\bfw^\top \bfphi(\bfx_k))\right] \\
& \quad + \bbE_{Q} \left[\log P(\bfw)\right] - \bbE_{Q} \left[\log Q(\bfw)\right]
\end{aligned}
$$
$$
\begin{aligned}
& \geq \sum_{k=1}^{K} \left( \log \sigma(\xi_k) - \frac{\xi_k}{2} + \frac{\bbE_{Q}\left[r_k\right]}{2} + \lambda(\xi_k) \left( \bbE_{Q} \left[ r_k^2 \right] - \xi_k^2 \right) \right) \\
& \quad + \bbE_{Q} \left[\log P(\bfw)\right] - \bbE_{Q} \left[\log Q(\bfw)\right] \\
& = \sum_{k=1}^{K} \left( \log \sigma(\xi_k) - \frac{\xi_k}{2} - \xi_k^2\lambda(\xi_k) + (y_k - \frac{1}{2}) \bfmu_t^\top \bfphi(\bfx_k) \right) \\
& \quad +\sum_{k=1}^{K} \left( \lambda(\xi_k) \bfphi^\top (\bfx_k) \left( \bfSigma_t + \bfmu_t\bfmu_t^\top\right) \bfphi(\bfx_k) \right) \\
& \quad + \frac{1}{2} \operatorname{tr}\left(\left(\bfSigma_{t}^{-1}-\bfSigma_{t-1}^{-1}\right)\bfSigma_t\right) + \frac{1}{2}\log \frac{|\bfSigma_t|}{|\bfSigma_{t-1}|} \\
& \quad - \frac{1}{2} \left(\bfmu_t - \bfmu_{t-1}\right)^\top \bfSigma_{t-1}^{-1} \left(\bfmu_t - \bfmu_{t-1}\right),
\end{aligned}
$$
where $r_k = y_k\bfw^\top \bfphi(\bfx_k), y_k \in \{0, 1\}.$

To maximize the lower bound $\calL$, we take derivatives with respect to the variational parameters $\{\xi_k\}_{k=1}^{K}$ and $\bfmu_t$, and set them to zero.
This results in the following update equations:
\paragraph{E-Step} $\partial \calL / \partial \bfmu_t = 0$ leads to
\begin{align}
        \bfmu_t         & = \bfSigma_t \left( \bfSigma_{t-1}^{-1} \bfmu_{t-1} + \sum_{k=1}^{K} \left(y_k - \frac{1}{2}\right) \bfphi(\bfx_k) \right), \label{eq:e_step_mu_multi_points} \\
        \bfSigma_t^{-1} & = \bfSigma_{t-1}^{-1} + \sum_{k=1}^{K} \Bigg|\frac{1/2 - \sigma(\xi_k)}{\xi_k}\Bigg| \bfphi(\bfx_k) \bfphi^\top (\bfx_k). \label{eq:e_step_sigma_multi_points}
\end{align}

\paragraph{M-Step} $\partial \calL / \partial \xi_k = 0$ leads to
\begin{equation}
    \xi_k = \sqrt{\bfphi^\top (\bfx_k) \left( \bfSigma_t + \bfmu_t\bfmu_t^\top \right) \bfphi(\bfx_k)}, \quad \forall k = 1, \ldots, K.
    \label{eq:m_step_multi_points}
\end{equation}
Note that $\xi_{k,t}=\xi_k$, where $t$ is omitted for simplicity. We suggest initialize $\xi_k$ to 0.0 or 1.0 for all $k$ (Empirically, both work well).

\subsection{\bf Practical Optimization for the Implementation}
For each generated dataset $\calD$, usually 1 to 2 EM iterations are sufficient for convergence.
To make the algorithm more efficient, we set $\phi_m(\bfx_k) = 0$ if its value is smaller than a threshold (e.g., $10^{-3}$) so that we can speed up the computation by the sparsity of $\bfphi(\bfx_k)$.
Besides, the update of $\bfmu$ and $\bfSigma$ needs matrix inversion of $\bfSigma_t^{-1}$, which is handled by Cholesky decomposition for efficiency and numerical stability.

\section{Approximation of BHM Prediction}
\label{sec:app:bhm_prediction}

\subsection{\bf Mean Occupied Probability}
\label{sec:app:bhm_occupied_probability}

With enough iterations, we get the converged $\bfmu$ and $\bfSigma$, and predict the occupancy probability at $\bfx_*$ by \eqref{eq:bhm_logistic_regression_model}:
\begin{equation}
    g(\bfx_*,\bfw) = P(y=1 \mid \bfx_*, \bfw) = \sigma(\bfw^\top \bfphi_*), \label{eq:func_g}
\end{equation}
where $\bfw \sim \calN(\bfmu,\bfSigma)$, $\bfphi_*=\bfphi(\bfx_*)$. Let $g_*(\bfw) = g(\bfx_*,\bfw)$. Then, the mean of $g_*(\bfw)$ is
\begin{equation}
    \bbE[g_*(\bfw)] = P(y=1 \mid \bfx_*) = \int \sigma\left(\bfw^\top \bfphi_*\right) P(\bfw) d\bfw.
\end{equation}
However, we cannot compute the above integral analytically. One way of approximation is to collect $N$ samples of $\bfw$ from $\calN(\bfmu,\bfSigma)$, so that
\begin{equation}
    \bbE[g_*(\bfw)] = P(y=1\mid \bfx_*) \approx \frac{1}{N} \sum_{i=1}^N \sigma(\bfw_i^\top \bfphi_*).
\end{equation}

A faster and more accurate approximation is
\begin{equation}
    \bbE[g_*(\bfw)] \approx \sigma\left( h \right), \quad h = \frac{\bfmu^\top \bfphi_*}{\sqrt{1 + \frac{\pi}{8}\bfphi_*^\top\bfSigma\bfphi_*}}. \label{eq:bhm_mean_prob}
\end{equation}

\begin{proof}
    We approximate the sigmoid function $\sigma(x)$ with the cumulative distribution function of the normal distribution, $\Phi(\lambda x)$, where $\lambda=\sqrt{\pi/8}$. Therefore,
    \begin{align*}
         & P(y=1\mid \bfx_*)                                                                                                                                                   \\
         & \approx \int \Phi(\lambda \bfw^\top \bfphi_*) P(\bfw) d\bfw = \int \Phi(\lambda u)P(u)du                                                                            \\
         & = \int \Phi\left( \frac{v + \bfmu^\top\bfphi_*/\sqrt{\bfphi_*^\top\bfSigma\bfphi_*}}{\left(\lambda\sqrt{\bfphi_*^\top\bfSigma\bfphi_*}\right)^{-1}} \right) P(v) dv \\
         & = \Phi\left( \frac{\bfmu^\top\bfphi_*/\sqrt{\bfphi_*^\top\bfSigma\bfphi_*}}{\sqrt{1 + \left(\lambda\sqrt{\bfphi_*^\top\bfSigma\bfphi_*}\right)^{-2}}} \right)       \\
         & = \Phi\left( \frac{\bfmu^\top\bfphi_*}{\sqrt{\lambda^{-2} + \bfphi_*^\top\bfSigma\bfphi_*}} \right)
        \approx \sigma\left( \frac{\bfmu^\top \bfphi_*}{\sqrt{1 + \lambda^{2}\bfphi_*^\top\bfSigma\bfphi_*}} \right)                                                           \\
         & = \sigma\left( \frac{\bfmu^\top \bfphi_*}{\sqrt{1 + \frac{\pi}{8}\bfphi_*^\top\bfSigma\bfphi_*}} \right),
    \end{align*}
    where $u = \bfw^\top\bfphi_* \sim \calN\left(\bfmu^\top\bfphi_*,\bfphi_*^\top\bfSigma\bfphi_*\right)$, $v=\frac{u-\bfmu^\top\bfphi_*}{\sqrt{\bfphi_*^\top\bfSigma\bfphi_*}} \sim \calN(0, 1)$. And the following integral is used
    \begin{equation}
        \int \Phi\left( \frac{w-a}{b} \right) P(w)dw = \Phi\left( \frac{-a}{\sqrt{1 + b^2}} \right). \label{eq:gaussian_cdf_integral}
    \end{equation}
    To prove the above equation, let $X \sim \calN(a, b^2)$ and $Y \sim \calN(0, 1)$ be independent random variables. Then,
    \begin{equation}
        P(X\le Y \mid Y = w) = P(X \le w) = \Phi\left( \frac{w-a}{b} \right),
    \end{equation}
    \begin{equation}
        \begin{aligned}
            P(X \le Y) & = \int P(X\le Y\mid Y=w) P(w)dw                 \\
                       & = \int \Phi\left( \frac{w-a}{b} \right) P(w)dw.
        \end{aligned}
    \end{equation}
    Since $P(X\le Y) = P(X-Y\le 0)$ and $X-Y \sim \calN(a, b^2 + 1)$, $P(X\le Y) = \Phi\left( \frac{-a}{\sqrt{1 + b^2}} \right)$. Therefore, \eqref{eq:gaussian_cdf_integral} holds.
\end{proof}
When it converges, $\det(\bfSigma) \approx 0$, we have
\begin{equation}
    \bbE[g_*(\bfw)] \approx \sigma\left( \bfmu^\top \bfphi_* \right),
\end{equation}
where $\bfmu^\top \bfphi_*$ is the log-odds of occupancy at $\bfx_*$.

\subsection{\bf Gradient and Surface Normal Prediction}
\label{sec:app:bhm_gradient_surface_normal}

We can also derive the gradient of \eqref{eq:bhm_mean_prob} as
\begin{equation}
    \begin{aligned}
         & \nabla \bbE[g_*(\bfw)] = \sigma(h)\sigma(-h) \nabla_{\bfx} \bfphi_* \nabla_{\bfphi_*}h,                                                                                                                               \\
         & \nabla_{\bfphi_*} h = \frac{1}{\sqrt{1 + \frac{\pi}{8}\bfphi_*^\top\bfSigma\bfphi_*}}\left(\bfmu - \frac{\frac{\pi}{8} \bfmu^\top \bfphi_* \bfSigma \bfphi_*}{1 + \frac{\pi}{8}\bfphi_*^\top\bfSigma\bfphi_*}\right),
    \end{aligned}
\end{equation}
where $\nabla_{\bfx}\bfphi_* = -2\gamma \bfX \operatorname{diag}(\bfphi_*)$, $\bfX=[\bfx_*-\tilde{\bfx}_1, \cdots, \bfx_*-\tilde{\bfx}_M,\mathbf{0}]$ for the RBF kernel and $M$ hinge points $\{\tilde{\bfx}_i\}_{i=1}^{M}$.


When $\det(\bfSigma) \approx 0$, the gradient of the occupancy probability at $\bfx_*$ is approximated as
\begin{equation}
    \bfv_* \approx \sigma(\bfmu^\top \bfphi_*)\sigma(-\bfmu^\top \bfphi_*)\nabla_{\bfx}\bfphi_* \bfmu.
\end{equation}


The surface normal at $\bfx_*$ is the opposite direction of $\bfv_*$. Since we do not care about the magnitude of $\bfv_*$ when calculate the surface normal, we have
\begin{equation}
    \bfn_* = -\frac{\bfz(\bfmu)}{\| \bfz(\bfmu) \|_2}, \quad \bfz(\bfmu) = \nabla_\bfx \bfphi_* \bfmu,
\end{equation}
and the variance is
\begin{equation}
    \bbV[\bfn_*] = \bfG^\top \bfSigma \bfG, \quad \bfG = \left( \nabla_{\bfx} \bfphi_* \right)^\top \frac{\bfI - \bfn_*\bfn_*^\top}{\| \bfz(\bfmu) \|_2}.
\end{equation}

\section{SDF Gradient Variance Calculation}
\label{sec:app:sdf_gradient_variance}

Based on the softmin approximation, we can obtain the approximation of the UDF gradient:
\begin{equation}
    \begin{aligned}
        \nabla d(\bfx_*) & \approx \nabla_{\bfx_*}h(\bfx_*,\{\bfx_i\}_{i=1}^N) = g(\bfx_*, \{\bfx_i\}_{i=1}^N) \\
                         & = \bfV \left(\alpha\bfs\bfs^\top\bfz-\alpha\diag(\bfs)\bfz + \bfs\right),           \\
        \bfV             & =
        \begin{bmatrix}\bfv_1 \cdots \bfv_N
        \end{bmatrix}, \bfv_i = \frac{\bfx_* - \bfx_i}{z_i}.
    \end{aligned}
\end{equation}

Therefore, assuming isotropic variance $\bbV_i$ for each dimension of $\bfx_i$, we can calculate the variance of $\nabla d(\bfx_*)$ approximately by Taylor expansion and get \maineqref{eq:sdf_gradient_variance_approximation}:
\begin{equation*}
    \bbV[\nabla_k u(\bfx_*)] \approx \sum_{i=1}^N \left\| \nabla_{\bfx_i} g_k(\bfx_*,\{\bfx_i\}_{i=1}^N) \right\|_2^2 \sigma_i^2, 1 \le k \le n,
\end{equation*}
where
\begin{equation}
    \begin{aligned}
         & \nabla_{\bfx_i} g(\bfx_*, \{\bfx_i\}_{i=1}^N) =                                     \\
         & \alpha\bfv_i \left(l_i\left(\bfv_i-\bfV\bfs\right) + s_i(\bfv_i - \bfg)\right)^\top
        + \frac{l_i}{z_i} \left(\bfv_i \bfv_i^\top - \bfI\right),                              \\
         & l_i = s_i(\alpha\bfs^\top\bfz - \alpha z_i + 1).
    \end{aligned}
\end{equation}

\section{Adaptive Scaling for GP Stability}
\label{sec:app:adaptive_scaling}

While larger $\lambda$ (smaller kernel scale) improves the approximation of \maineqref{eq:varadhan_equation}, it triggers numerical underflow of $k(\bfx_*,\bfx_i)$.
To address this, we introduce a scaling factor $\gamma$ and rewrite \maineqref{eq:log_gp_posterior_mean}:
\begin{equation}
    \begin{bmatrix} \hat{f'}(\bfx_*) \\ \nabla \hat{f'}(\bfx_*)
    \end{bmatrix} =
    \begin{bmatrix} {\bfk'}_*^\top \\ \nabla_{\bfx_*}^\top {\bfk'}_*
    \end{bmatrix} (\bfK +\bfSigma_y)^{-1} \mathbf{1},
    \label{eq:scaled_log_gp_posterior_mean}
\end{equation}
where $k'_{*,i} = k'(\bfx_i, \bfx_*) = \gamma k(\|\bfx_i - \bfx_*\|_2)$.
$\gamma>0$ can be merged into the exponent. By setting $\gamma$ based on the distance $d=\|\bfx_* - \bfx_1\|_2$, we effectively normalize the kernel values and prevent underflow. We set $\gamma = e^{d^2/(2l^2)}$ for the RBF kernel or $\gamma = e^{\sqrt{3} d/l}$ for the Mat\'ern 3/2 kernel. Then, the corresponding inverse transformations $r(\hat{f'})$ is adjusted to $r(\hat{f'})= \sqrt{-2l^2\log\hat{f'}+d^2}$ for RBF and $r(\hat{f'})= -\frac{l}{\sqrt{3}} \log \hat{f'} + d$ for Mat\'ern 3/2 kernel.

\section{Limit Analysis of the Softmin Approximation}
\label{sec:app:softmin_limit_analysis}

This section provides the full derivation summarized by \maineqref{eq:weak_corr_limit} and \maineqref{eq:strong_corr_limit} in Sec.~\mainref{sec:kernel_extension}, showing that the log-GP posterior mean behaves as a softmin over the surface set in both the weakly- and strongly-correlated regimes of a noisy training set.

Let $\bfK\in\bbR^{N\times N}$ with $K_{ij}=k(\bfx_i,\bfx_j)$ and $\bfSigma=\sigma^2 \bfI$. Because every training label in $\calD_\text{surf}$ is $f(\bfx_i)=1$, the posterior mean reduces to $\hat{f}_* = \bfk_*^\top(\bfK+\bfSigma)^{-1}\mathbf{1}$.

\subsection{\bf Weakly-correlated regime}
When the kernel scale is small relative to the sample spacing, $K_{ij}\approx 0$ for $i\neq j$, so $\bfK \rightarrow \bfI$ and
\begin{align}
    (\bfK+\bfSigma)^{-1}\mathbf{1} &\approx \frac{\mathbf{1}}{1+\sigma^2}, \\
    \hat{f}_* &\approx \frac{1}{1+\sigma^2}\sum_i k_{*i} = \frac{S}{1+\sigma^2} \max_i k_{*i},
\end{align}
where $S=\sum_i k_{*i} / \max_i k_{*i} \ge 1$ is the ratio of the sum of kernel values to the maximum kernel value, which accounts for the inflation effect in the softmin approximation. Taking the logarithm gives \maineqref{eq:weak_corr_limit}:
\begin{equation}
    \log\hat{f}_* \approx \log \max_i k_{*i} + \log S - \log (1 + \sigma^2),
\end{equation}
which shows that the log-GP posterior mean is approximately a softmin function, and the UDF is recovered via the kernel-specific nonlinear transform $r(\hat{f})$ in Table~\mainref{tab:kernel_nonlinear_transform}.

Two opposite biases act on the recovered UDF. Because the softmin sums over all nearby surface samples, $\sum_i k_{*i}\ge\max_i k_{*i}$, so $\hat{f}_*$ is slightly inflated and, through the decreasing transform $r$, the recovered distance is biased \emph{below} the true distance. This is a deterministic recovery bias, present even in the noise-free case, that shrinks as the kernel scale decreases. The noise variance $\sigma^2$ acts in the opposite direction: it shrinks $\hat{f}_*$ by $1/(1+\sigma^2)$, biasing the recovered distance slightly \emph{above} the true distance, and additionally down-weights uncertain samples. In practice, the recovery bias dominates, so the net prediction is a conservative under-estimate, which is beneficial for safety.

\begin{figure*}[t]
    \centering
    \begin{subfigure}{0.32\linewidth}
        \centering
        \includegraphics[width=\textwidth]{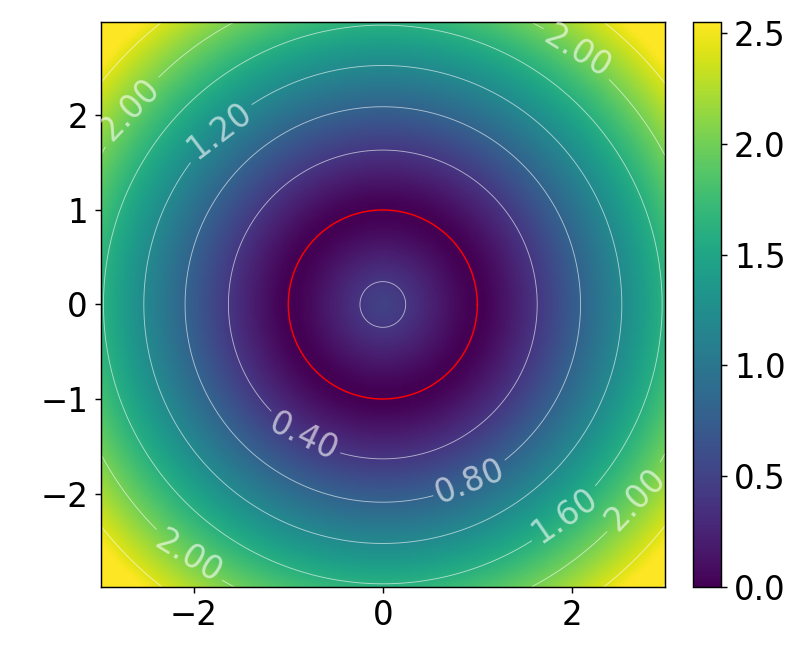}
        \caption{} \label{fig:logedf_native_error:map}
    \end{subfigure}
    \hfill
    \begin{subfigure}{0.32\linewidth}
        \centering
        \includegraphics[width=\textwidth]{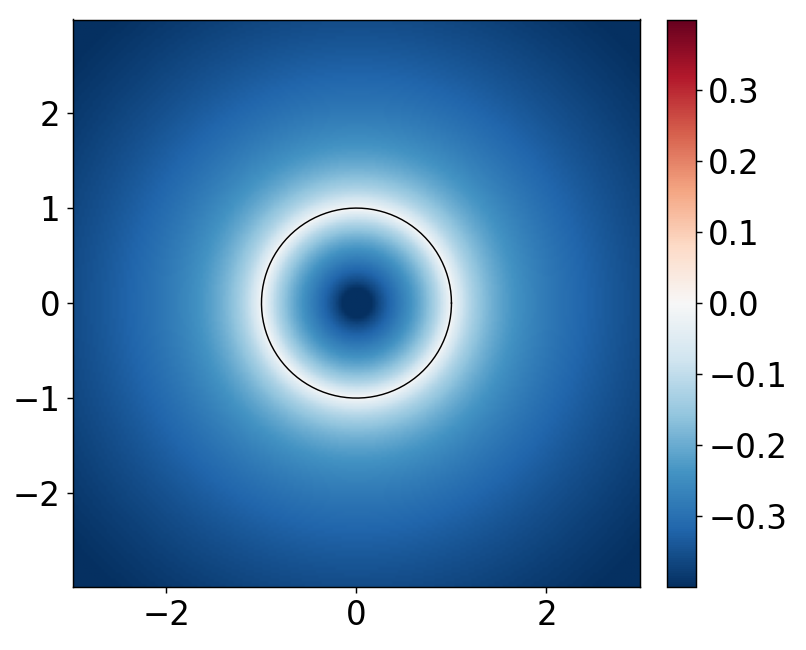}
        \caption{} \label{fig:logedf_native_error:error_plot}
    \end{subfigure}
    \hfill
    \begin{subfigure}{0.32\linewidth}
        \centering
        \includegraphics[width=\textwidth]{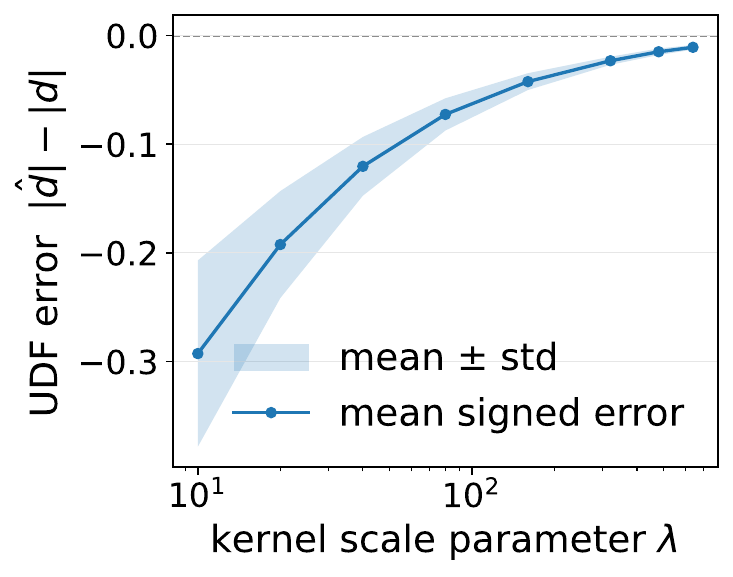}
        \caption{} \label{fig:logedf_native_error:error}
    \end{subfigure}
    \caption{\small Noise-free ($\sigma^2=0$) benchmark of the log-GP UDF on a 2D unit-circle. (a) Predicted UDF map at $\lambda=10$; the red curve is the unit circle. (b) Signed error (prediction minus ground truth): negative everywhere off the surface, a pure under-estimate. (c) Error vs.\ $\lambda$: the bias stays negative and shrinks as $\lambda$ grows, matching the recovery term $-2l^2\log S$.}
    \label{fig:logedf_native_error}
\end{figure*}

Taking the RBF kernel as an example, with reverse transform $\hat{d}=r(\hat{f}) = l\sqrt{-2\log \hat{f}}$,
\begin{equation}
    \hat{d} \approx \sqrt{\,\underbrace{-2l^2\log \max_i k_{*i}}_{d^2~(\text{true UDF})} \underbrace{-\,2l^2\log S}_{\le\,0~(\text{recovery})} +\underbrace{2l^2\log(1+\sigma^2)}_{\ge\,0~(\text{noise})}\,}.
\end{equation}
The recovery term pulls $\hat{d}$ below the true distance $d$, while the noise term pushes it above. Since $S\ge 1+\sigma^2$ in practice, the net bias is a conservative under-estimate, and both terms shrink with the kernel scale $l$. A noise-free recovery study ($\sigma^2=0$, Fig.~\ref{fig:logedf_native_error}) confirms this: the predicted UDF (Fig.~\ref{fig:logedf_native_error:map}) stays below the true distance (Fig.~\ref{fig:logedf_native_error:error_plot}), and the gap shrinks as $\lambda$ grows (Fig.~\ref{fig:logedf_native_error:error}), matching the recovery term.

\subsection{\bf Strongly-correlated regime}
In the dense-sampling case, $K_{ij} \to 1$ so $\bfK \to \mathbf{1}\mathbf{1}^\top$, and by the Sherman--Morrison formula,
\begin{align}
    (\bfK+\bfSigma)^{-1}\mathbf{1} &\approx (\sigma^2 \bfI + \mathbf{1}\mathbf{1}^\top)^{-1}\mathbf{1} = \frac{\mathbf{1}}{\sigma^2 + N}, \\
    \hat{f}_* &\approx \frac{1}{\sigma^2 + N}\sum_i k_{*i} = \frac{S}{\sigma^2 + N} \max_i k_{*i},
\end{align}
where $S=\sum_i k_{*i}/\max_i k_{*i}$ as before. Here the tightly-clustered samples have nearly equal kernel values, so $S\to N$ (maximal inflation) and $\hat{f}_*$ is approximately the arithmetic mean of the $k_{*i}$. This is the $\alpha \to 0$ degeneracy of the softmin (uniform weights): the kernel bumps merge into a single wide bump. For a query at distance $\sim z$ from a tight cluster, all $\|\bfx_*-\bfx_i\|\approx z$, so $\hat{f}_* \approx \frac{N}{\sigma^2 + N} k(z)$ and, taking the logarithm, we obtain \maineqref{eq:strong_corr_limit}:
\begin{equation}
\begin{aligned}
    \log \hat{f}_* &\approx \log k(z) + \log S - \log(\sigma^2 + N) \\
    &\approx \log k(z) - \log{\left(\frac{\sigma^2}{N} + 1\right)},
\end{aligned}
\end{equation}
where the second step uses $S\to N$. The kernel-specific transform $r(\hat{f})$ still recovers $z$. The recovery inflation $\log S\to\log N$ is now absorbed by the $N$-sample normalization $\log(\sigma^2+N)$, so the deterministic under-estimate cancels and only the noise over-estimate $\log(\frac{\sigma^2}{N} + 1)$ remains. This bias is much smaller than $\log(1+\sigma^2)$ in the weakly-correlated case, indicating that highly clustered samples are more confident; in practice $N \gg \sigma^2$ and the bias becomes negligible.

\subsection{\bf Intermediate regime}
The residual gap lives in the intermediate regime, where $(\bfK+\bfSigma)^{-1}\mathbf{1}$ produces weights that are neither uniform nor a nearest-sample indicator. Combining the two limits, the posterior mean can be written approximately as
\begin{equation}
    \hat{f}_* \approx \sum_{i=1}^N \frac{1}{\sigma_i^2 + 1} k(\|\bfx_*-\bfx_i\|),
\end{equation}
for $N$ clusters of different sizes and correlation levels, where $\sigma_i^2$ is the equivalent noise variance of the $i$-th cluster and is smaller for denser or more confident clusters (e.g., $\sigma_i^2 = \sigma^2/N_i$ when the $i$-th cluster of $N_i$ samples is strongly correlated). The prediction remains a weighted average of kernel bumps dominated by the near samples, with the per-sample noise term $\sigma_i^2$ down-weighting uncertain samples, a desirable regularization effect. In the noise-free case $\sigma_i^2=0$, the GP posterior mean reduces exactly to the softmin, leaving only the deterministic recovery under-estimate.

\section{Consistency of Min-Fused Local GPs}
\label{sec:app:gp_consistency}

This section provides the 2D demonstration supporting the footnote in Sec.~\mainref{sec:log_gp_udf}, that min-fusing overlapping local GPs matches a single global GP in both value and gradient across octant boundaries.

\begin{figure}[t]
    \centering
    \begin{subfigure}{0.32\linewidth}
        \centering
        \includegraphics[width=\linewidth]{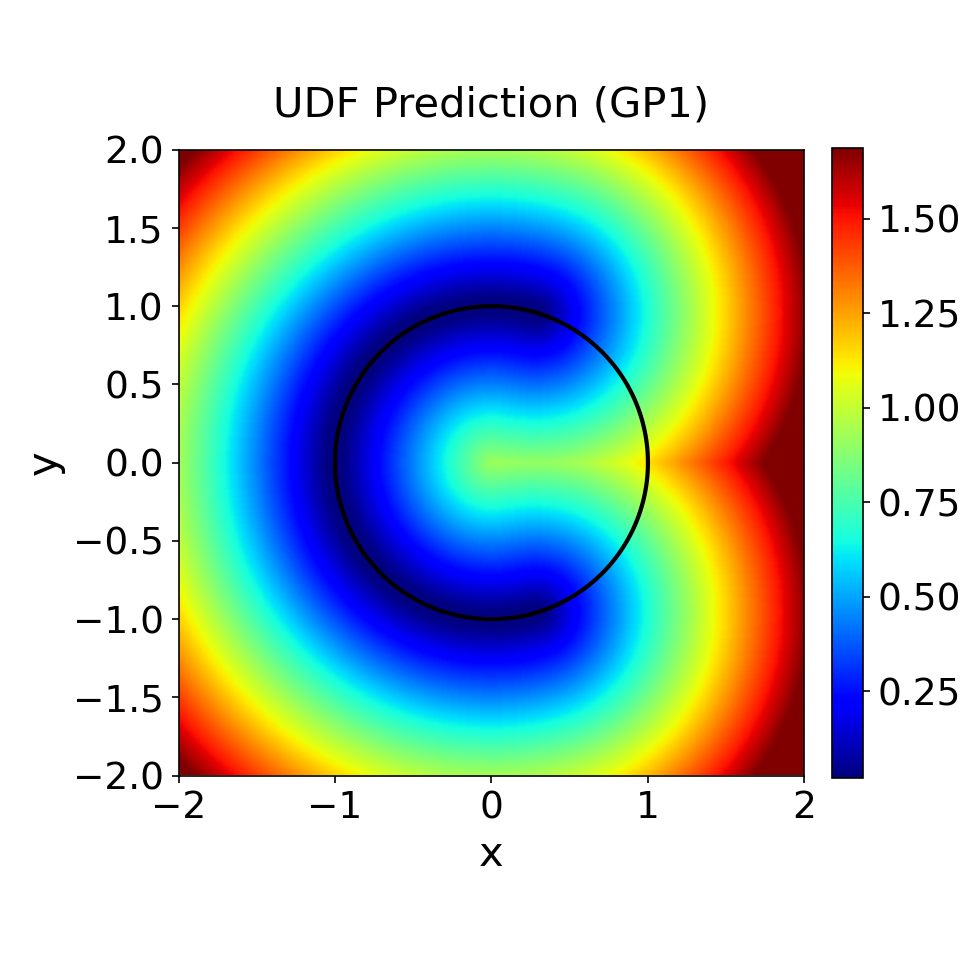}
        \caption{Local GP1 $(x\le m)$} \label{fig:gp_consistency_demo:gp1}
    \end{subfigure}
    \hfill
    \begin{subfigure}{0.32\linewidth}
        \centering
        \includegraphics[width=\linewidth]{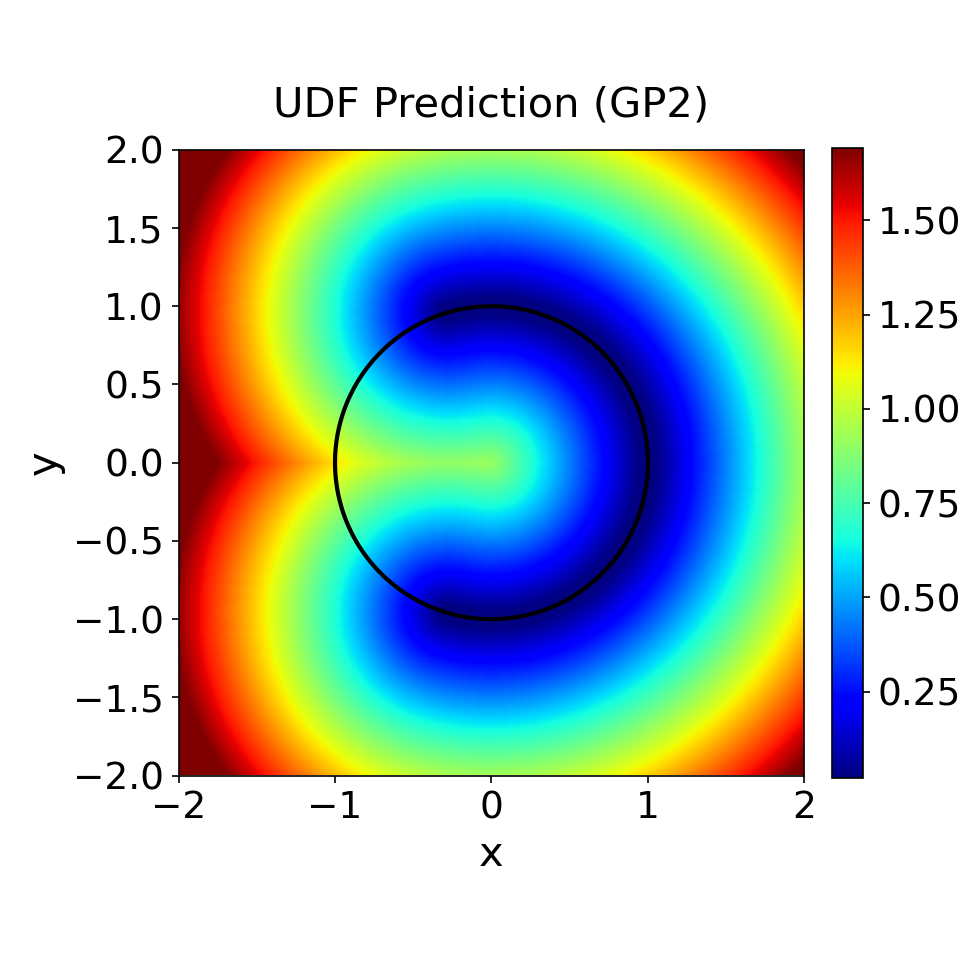}
        \caption{Local GP2 $(x\ge -m)$} \label{fig:gp_consistency_demo:gp2}
    \end{subfigure}
    \hfill
    \begin{subfigure}{0.32\linewidth}
        \centering
        \includegraphics[width=\linewidth]{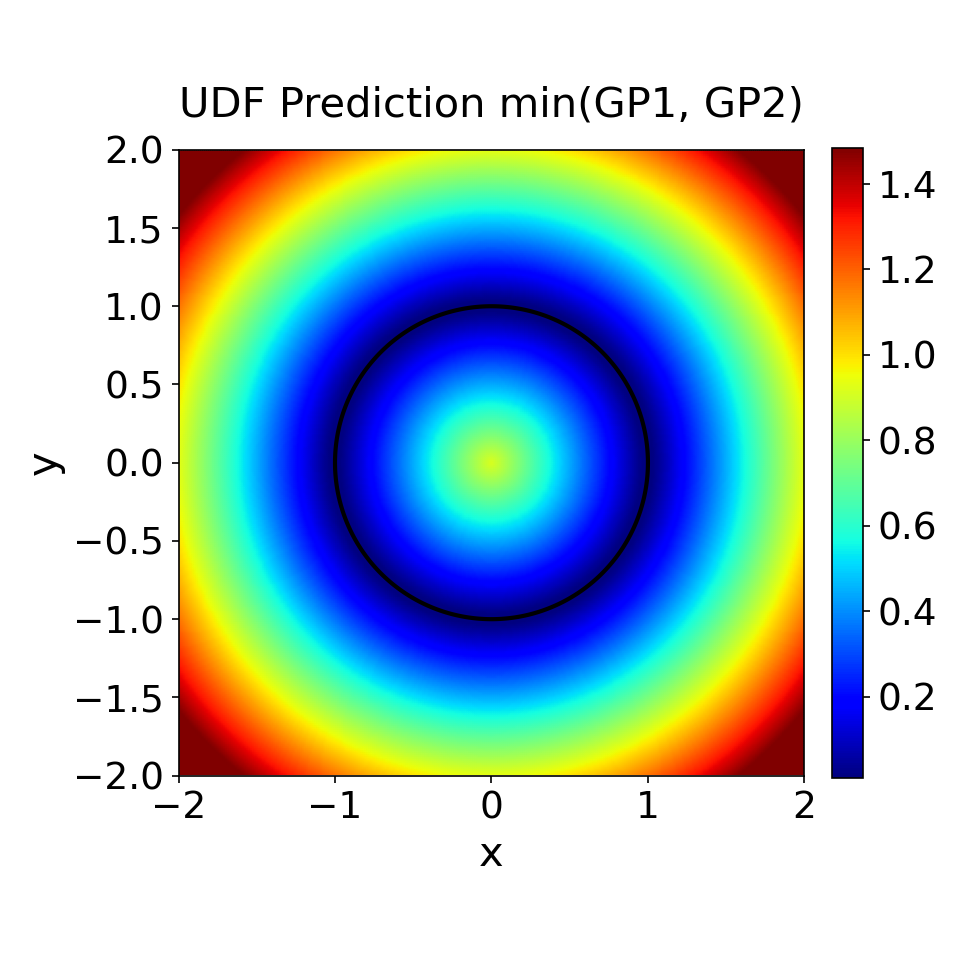}
        \caption{Min fusion} \label{fig:gp_consistency_demo:min}
    \end{subfigure}

    \begin{subfigure}{0.32\linewidth}
        \centering
        \includegraphics[width=\linewidth]{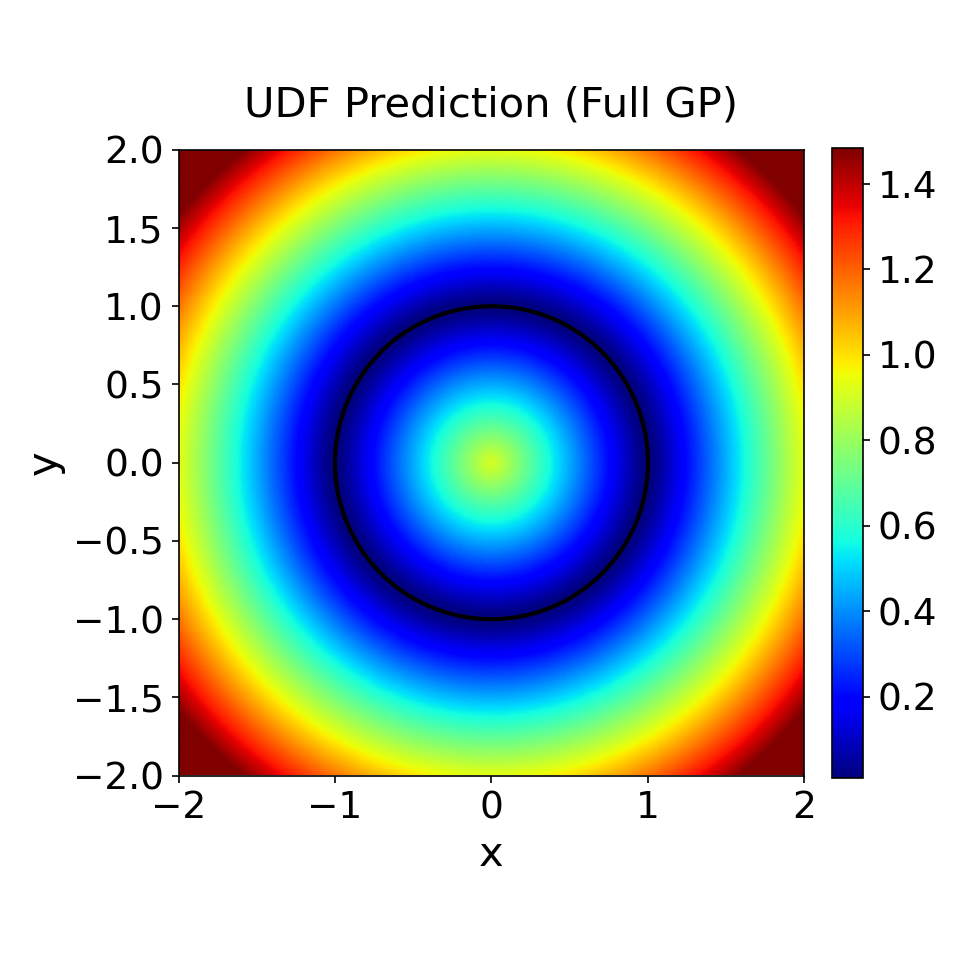}
        \caption{Global GP} \label{fig:gp_consistency_demo:full}
    \end{subfigure}
    \hfill
    \begin{subfigure}{0.32\linewidth}
        \centering
        \includegraphics[width=\linewidth]{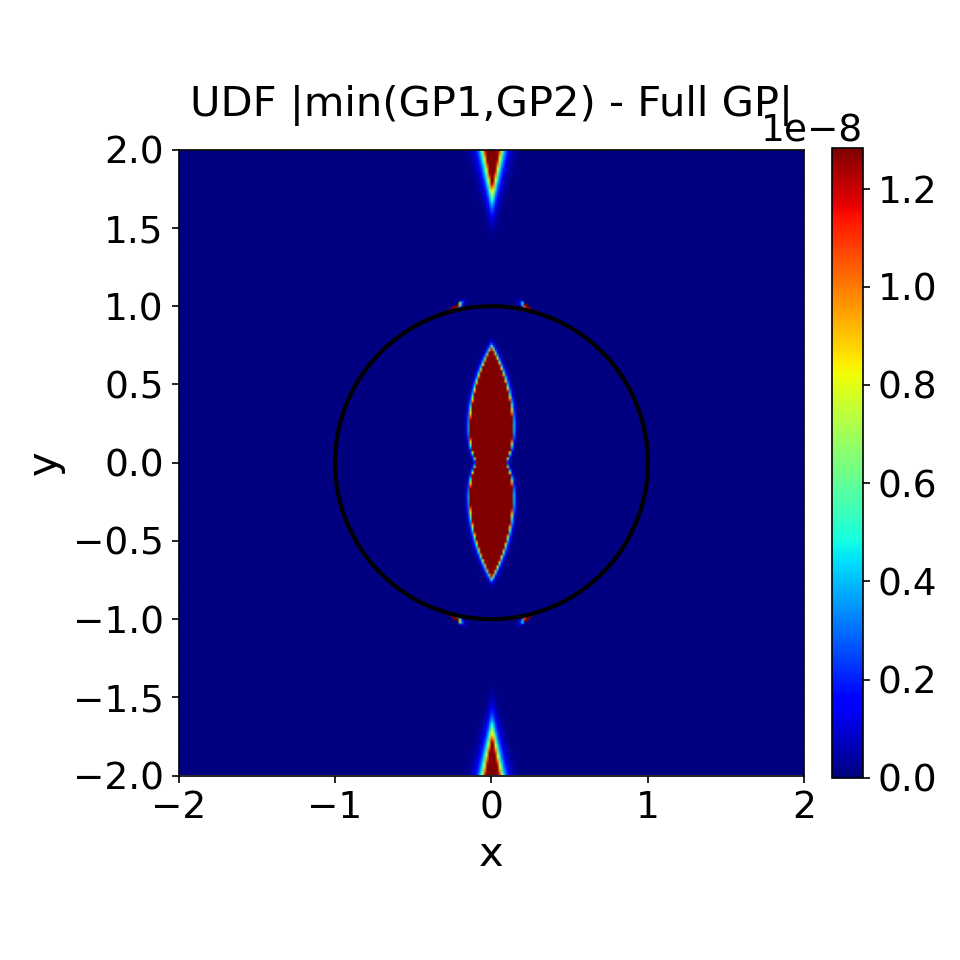}
        \caption{$|\text{Min}-\text{Global}|$} \label{fig:gp_consistency_demo:diff}
    \end{subfigure}
    \hfill
    \begin{subfigure}{0.32\linewidth}
        \centering
        \includegraphics[width=\linewidth]{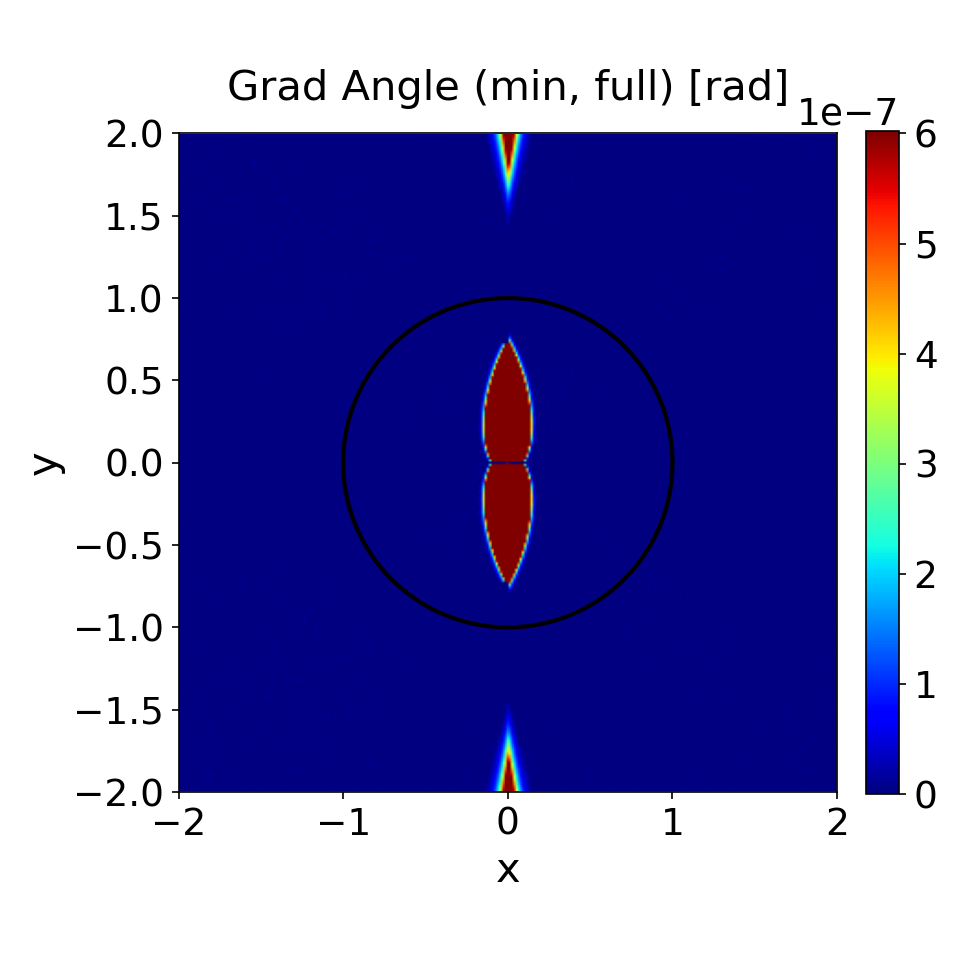}
        \caption{Gradient angle diff.} \label{fig:gp_consistency_demo:grad_diff}
    \end{subfigure}
    \caption{\small 2D consistency demo of minimum-fusion local GPs versus a single global GP. Two overlapping local GPs (a), (b) are minimum-fused (c) and compared against a global GP trained on all samples (d). The fusion is almost identical to the global GP: the absolute UDF difference (e) is at the $10^{-8}$ level and the gradient angular error (f) is below $10^{-6}\,$rad across the whole domain, including the GP-switching boundary. The black curve marks the zero-level set (the unit circle).}
    \label{fig:gp_consistency_demo}
\end{figure}

The training samples are drawn uniformly along a unit circle and partitioned by their $x$-coordinate into two overlapping subsets: local GP1 is trained on the samples with $x\le m$ and local GP2 on those with $x\ge -m$, so that the two GPs share an overlap band of width $2m$ (here $m=0.3$). We compare the UDF predicted by each local GP, their minimum fusion, and a single global GP trained on all samples (Fig.~\ref{fig:gp_consistency_demo}). The minimum fusion closely matches the global GP: across the entire domain, the UDF difference stays at the $10^{-8}$ level and the angular error between the predicted gradients remains below $10^{-6}\,$rad, confirming that the overlap preserves both value and gradient consistency, including at the GP-switching boundary.

\section{Priority-Based Update Scheme}
\label{sec:app:priority_based_update}

In \methodname, several operations take significant time, including BHM EM update, BHM marching, GP training data collection, and GP training. However, not all operations need to be performed at every time step. Therefore, we introduce a priority-based update scheme to efficiently allocate computational resources to the most needed updates.

To maintain the responsiveness of the system to the latest sensor data, BHM EM updates are performed at every time step for all BHMs that have new sensor data. Newly created BHMs also need to be marched immediately to get surface points for its corresponding GP training data buffer update. For all GP training operations, we delay them until they are required for SDF prediction.

For other operations, we use priority queues to determine the order of updates based on their necessity.
In order to make the system also responsive to the latest query trends (e.g., the down-stream planner may focus on a specific area), each GP has a query counter $c_0$ that records how many times the GP is requested for SDF prediction. The counter helps to reshape the priority of GP training and data buffer updates. When a GP is trained, its query counter is reduced by half. And the query counter is capped to a maximum value (10000) to avoid over-prioritization.

\subsection{\bf Marching Priority Queue}
Note that a single EM update of a BHM only affects a small portion of the surface points. Especially for well-updated BHMs, the surface point changes are even smaller. Hence, we first introduce a priority queue for marching BHMs, which orders the BHMs by the latest timestamp $t_{\calB}$ when a BHM is requested to update surface estimation. The BHM with the oldest timestamp (smaller $t_{\calB}$) in the queue is processed first because it is more likely to be unchanged in the near future:
\begin{equation}
    s_\text{march} = t_{\calB} \downarrow,
\end{equation}
where $\downarrow$ indicates smaller values have higher priority.

\subsection{\bf GP Data Buffer Update Queue}
After marching a BHM, its corresponding GP training data buffer needs to be updated with the new surface points. However, it also takes time to get significant changes of $\calD_\text{surf}$ for updating the GP. And $\calD_\text{surf}$ is not required until the GP needs to be trained. Therefore, we introduce a priority queue that orders the GPs by the following score:
\begin{equation}
    s_\text{buffer update} = c_1 (1 + \eta_1 c_0) \uparrow,
\end{equation}
where $c_1$ is the number of times the GP data buffer is marked to be updated since the last update, $\eta_1$ is a weight parameter, and $\uparrow$ indicates larger values have higher priority. $c_1$ is increased by 1 each time the GP is marked to update its data buffer after marching its corresponding BHM and reset to 0 after the data buffer is updated. For newly created GPs, we initialize $c_1$ to $c_1^{\max} \exp(-\gamma \|\bfc_{GP} - \bfo_t\|_2)$, where $\bfc_{GP}$ is the center of the GP training data collection bounding box, $\bfo_t$ is the current sensor origin, and $\gamma > 0$ is a decay parameter. This initialization gives higher priority to GPs closer to the sensor origin.

\subsection{\bf GP Training Queue}
Similar to the GP data buffer update, GP training is not required until SDF prediction is requested. In addition, training a GP only makes sense when its training data buffer has significant changes. Therefore, we introduce another priority queue that orders the GPs by the following score:
\begin{equation}
    s_\text{GP train} = c_2 (1 + \eta_2 c_0) \uparrow,
\end{equation}
where $c_2$ is the number of times the GP is marked to be trained since the last training, $\eta_2$ is a weight parameter, and $\uparrow$ indicates larger values have higher priority. $c_2$ is increased by $c_1$ each time the GP's buffer is updated and reset to 0 after the GP is trained. When the GP is trained, $c_0$ is also halved in case the GP is over-prioritized for future updates as the query trend may change.

\section{Efficient Tree Construction}
\label{sec:app:efficient_tree_construction}

\begin{figure}[t]
    \centering
    \begin{subfigure}[t]{0.38\linewidth}
        \centering
        \includegraphics[width=\linewidth]{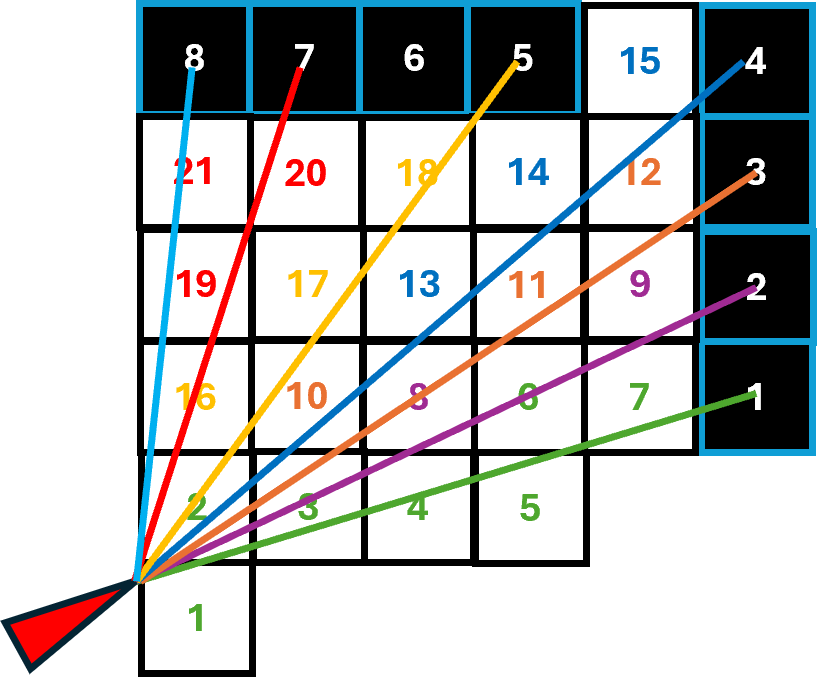}
        \caption{Voxel update order}
        \label{fig:occupancy_tree_sorted_update}
    \end{subfigure}
    \begin{subfigure}[t]{0.494\linewidth}
        \centering
        \includegraphics[width=\linewidth]{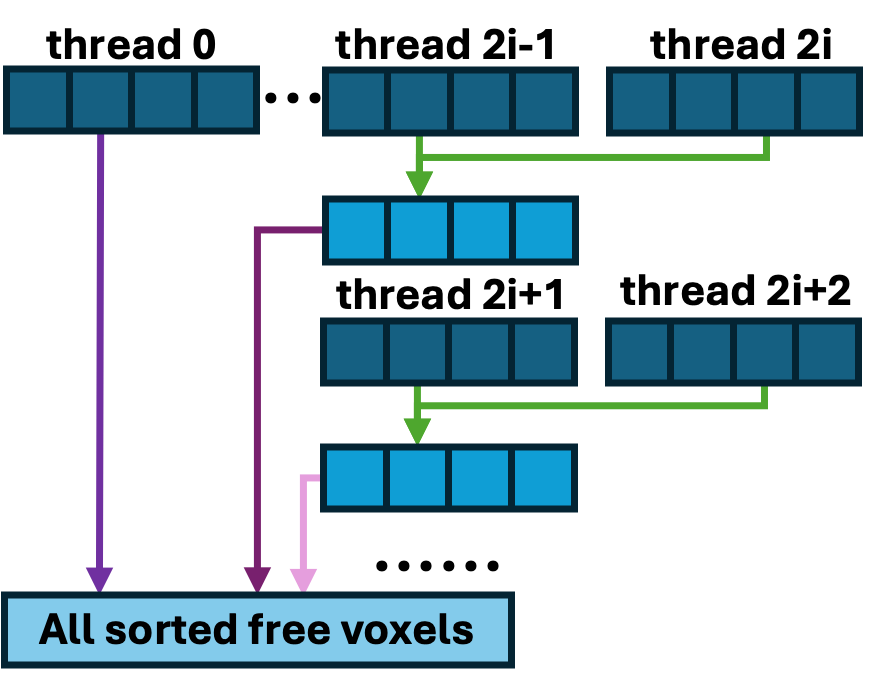}
        \caption{Free voxel parallel generation}
        \label{fig:free_cells_computation}
    \end{subfigure}
    \caption{
    To construct the tree efficiently, we implement two key optimizations: (a) voxels are updated in ray-sorted order to maximize the CPU cache hit rate; and (b) this ordering is generated via a stride-2 parallel reduction. In this scheme, threads first process ray batches in parallel, then odd-indexed threads ($2i-1$) merge results from their even-indexed counterparts ($2i$). Finally, these partially merged results are aggregated sequentially into thread 0.
    }
    \label{fig:occupancy_tree_optimized_update}
\end{figure}

To optimize occupancy mapping, we extend the Octomap \cite{hornung13octomap} to include a dedicated 2D quadtree implementation. Performance is further boosted by updating voxels in a sorted, along-ray order (Fig. \ref{fig:occupancy_tree_sorted_update}).
This approach leverages spatial locality that when a free voxel is updated, its neighboring voxels are more likely to be updated, improving the CPU cache hit rate.
While the order of occupied voxels is obvious, we implement an algorithm demonstrated in Fig. \ref{fig:free_cells_computation} to efficiently generate the sorted free-voxel sequence.

\section{Experiment Details}
\label{sec:app:experiment_details}

\subsection{Parameter Settings}

For \methodname, we set the octree resolution to 0.08m, 0.05m and 0.7m for the Replica, Cow\&Lady, and Newer College datasets, respectively. The local BHMs are placed at the depth $D_\text{tree}-1$, which means the local BHM bounding box size is twice the octree voxel size. The number of hinge points per axis is set to 7 for the Replica and Cow\&Lady datasets, and 9 for the Newer College dataset. The BHM kernel scale $l$ is set to 0.016m, 0.013m and 0.11m for the three datasets, respectively. For the GP, we use RBF kernel and set $\lambda=\frac{1}{2l^2}$ to 500, 300, and 250 for the three datasets, respectively. For other parameters, please refer to our released code.

For the baselines, we use the default parameters provided in their released code except for necessary modifications to adapt to the datasets. For example, iSDF \cite{ortiz_isdf_2022} only supports image input, so we convert the LiDAR scans of the Newer College dataset to depth images using the camera intrinsic parameters that give the same field of view as the LiDAR. We set the voxel size of Voxblox and FIESTA to be 0.05m for the Replica and Cow\&Lady datasets, and 0.2m for the Newer College dataset to get a better trade-off between accuracy and efficiency.

For our method and VDB-GPDF \cite{wu_vdb-gpdf_2025}, a mesh is available from the algorithm directly. For Voxblox \cite{oleynikova_voxblox_2017}, FIESTA \cite{han_fiesta_2019}, and iSDF \cite{ortiz_isdf_2022}, we extract the mesh using the marching cubes algorithm \cite{lorensen_marching_1987} with the voxel size as the marching cubes resolution, except for iSDF, where we use a resolution of 0.02m for the Replica and Cow\&Lady datasets and 0.1m for the Newer College dataset.

\subsection{Mesh Metrics}

For Replica dataset, we sample 2 million points from the ground truth mesh and the reconstructed mesh, respectively. For Cow\&Lady dataset, we sample 54k points from the ground truth point cloud and the reconstructed mesh, respectively. For Newer College dataset, we sample 2 million points from the ground truth point cloud and the reconstructed mesh, respectively.

We compute precision, recall, and F1 score with a threshold of 0.05m for the Replica and Cow\&Lady datasets, and 0.2m for the Newer College dataset. A point is considered to be correctly reconstructed if its distance to the other point cloud is less than the threshold.

Besides, we compute the accuracy and the completion, which are defined as:
\begin{align}
    \text{Accuracy}   & = \frac{1}{|\calP|} \sum_{\bfp \in \calP} \min_{\bfq \in \calQ} \|\bfp - \bfq\|_2,\\
    \text{Completion} & = \frac{1}{|\calQ|} \sum_{\bfq \in \calQ} \min_{\bfp \in \calP} \|\bfp - \bfq\|_2,
\end{align}
where $\calP$ is the reconstructed point cloud and $\calQ$ is the ground truth point cloud.

We also choose to use Chamfer-L1 distance in place of the standard Chamfer distance for more intuitive interpretation of the error in meters, which is the average of accuracy and completion.

\subsection{SDF Metrics}

For SDF evaluation, we query each method with a regular grid of points covering the bounding box of the ground truth mesh or point cloud. The grid resolution is set to 0.05m for the Replica and Cow\&Lady datasets, and 0.2m for the Newer College dataset. Prediction may fail for Voxblox and FIESTA at some query points because they rely on the eight voxel corners to interpolate the SDF value. In such cases, we exclude the failed points when calculating the metrics for Voxblox and FIESTA.

\subsection{Time Metrics}

While it is straightforward to measure the prediction time for all methods, measuring the update time is more complex due to the different update strategies employed by each method. For a fair comparison, we measure the average update time per scan for all methods. Specifically, we record the total time taken to process all scans in a dataset (but exclude time of data loading, logging and etc.) and divide it by the number of scans to obtain the average update time per scan. This approach provides a consistent basis for comparing the efficiency of each method's update process.

\section{Full Sign-Prediction Metrics}
\label{sec:app:sign_metrics}

This section provides the full quantitative support for the deterministic-sign treatment discussed in Sec.~\mainref{sec:consistent_uncertainty_quantification} and the sign-accuracy summary in Sec.~\mainref{sec:sdf_accuracy}, including the empirical sharp-transition evidence, the sign-mixture variance derivation, and the complete per-scene sign-metric table.

\subsection{\bf Sharp-Transition Evidence}
\begin{figure}[t]
    \centering
    \begin{subfigure}{0.49\linewidth}
        \centering
        \includegraphics[width=\linewidth]{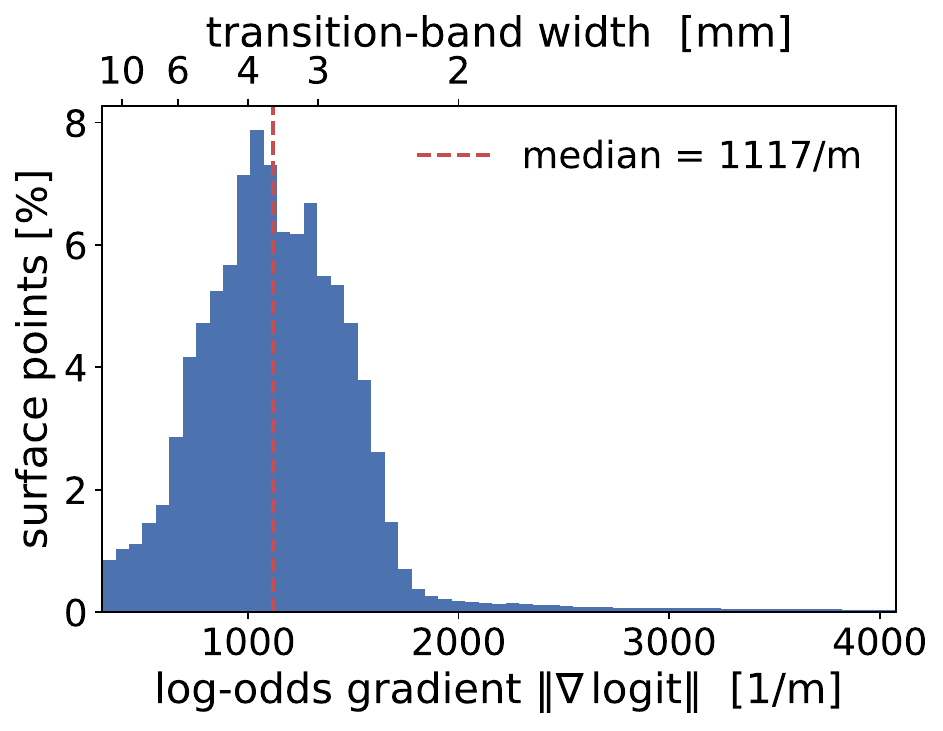}
        \caption{} \label{fig:bhm_sharp_transition:a}
    \end{subfigure}
    \hfill
    \begin{subfigure}{0.49\linewidth}
        \centering
        \includegraphics[width=\linewidth]{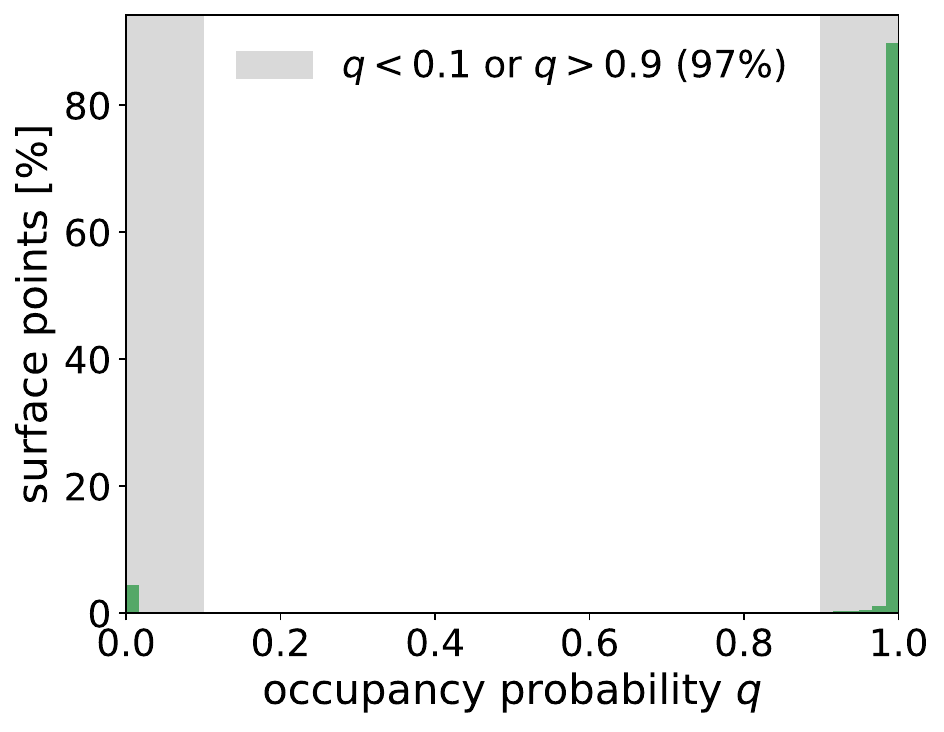}
        \caption{} \label{fig:bhm_sharp_transition:b}
    \end{subfigure}
    \caption{\small Sharp transition of the BHM occupancy field at the ground-truth surface vertices of the Replica \texttt{room0} scene. (a) Occupancy log-odds gradient norm $\|\nabla\log\frac{q}{1-q}\|$, with the equivalent transition-band width (top axis, median $\approx 3.6\,$mm). (b) Occupancy probability $q$ at the surface: $97\%$ of points have $q<0.1$ or $q>0.9$ (shaded), so the sign is effectively deterministic.}
    \label{fig:bhm_sharp_transition}
\end{figure}

BHM provides a probabilistic sign prediction $P(s(\bfx)=1) = q$ and $P(s(\bfx)=-1) = 1-q$, where $s(\bfx)$ is the sign at query $\bfx$. To substantiate the sharp-transition assumption empirically, we evaluate the learned BHM occupancy field at all ground-truth surface vertices of the Replica \texttt{room0} scene (Fig.~\ref{fig:bhm_sharp_transition}). At the surface, the occupancy log-odds gradient $\nabla\log\frac{q}{1-q}$ has a median norm of $\approx 1.1\times10^{3}\,\text{m}^{-1}$ (Fig.~\ref{fig:bhm_sharp_transition:a}). Equivalently, the occupancy probability rises from $0.12$ to $0.88$ over a transition band of median width $\approx 3.6\,\text{mm}$ ($95$th percentile $\approx 7.4\,\text{mm}$), far below the map resolution. As a direct consequence, $97\%$ of the surface points already have $q<0.1$ or $q>0.9$ (Fig.~\ref{fig:bhm_sharp_transition:b}): the occupancy probability saturates within a sub-voxel shell. So, for any query slightly off the surface, $q$ is effectively $0$ or $1$, which justifies the deterministic-sign treatment.

\subsection{\bf Sign-Mixture Variance}
Under the probabilistic sign, the SDF prediction $\hat{d}(\bfx)$ has a mixture distribution with density
\begin{equation}
    p(\hat{d}(\bfx)) = q\, p(\hat{u}(\bfx)\mid s(\bfx)=1) + (1-q)\, p(\hat{u}(\bfx)\mid s(\bfx)=-1),
\end{equation}
where $\hat{u}(\bfx)$ is the predicted unsigned distance at $\bfx$. Because BHM learns a continuous occupancy field with a sharp transition at the surface, $q$ is close to either $0$ or $1$, so the mixture collapses to $p(\hat{d}(\bfx))=p(\hat{u}(\bfx))$. Thus $\bbE[\hat{d}(\bfx)] = s(\bfx)\bbE[\hat{u}(\bfx)]$ and $\bbV[\hat{d}(\bfx)] = \bbV[\hat{u}(\bfx)]$: the uncertainty of the SDF prediction is solely determined by the UDF prediction from the GP back-end, and the deterministic-sign treatment introduces no additional uncertainty in practice. For the $\sim\!3\%$ of near-surface points without a sharp transition ($q\in(0.1,0.9)$), the exact mixture variance contains an additional term proportional to $q(1-q)\bbE[\hat{u}]^2$ that this treatment discards, so the predictive variance is theoretically underestimated in that near-surface subset.

\subsection{\bf Full Per-Scene Sign Metrics}
Table~\ref{table:sign_metrics} reports a complete evaluation of the sign prediction on the Replica dataset, where each method's predicted sign is compared against the ground-truth sign extracted from the ground-truth mesh. Ground-truth sign for Cow \& Lady and Newer College is unavailable due to the lack of ground-truth meshes. We report F1 score, precision, recall, and accuracy separately for points near the surface (Near), far from the surface (Far), and all points (All). Far from the surface the sign is unambiguous and every method exceeds $99\%$ on all metrics, so the near-surface region is the discriminating one, since this is where sign errors are both most likely and most consequential for downstream planning.

Treating \emph{free} as the positive class, precision is a safety-critical quantity because it is the fraction of points declared free that are truly free; high precision corresponds to a low false-free (collision) rate. In the near region, our BHM front-end attains the highest precision, ranking first in 5/8 scenes and second in the remaining 3, and stays above $99\%$ in every scene. Its recall and accuracy are not the best: FIESTA and iSDF rank higher because they label more points as free, but our scores remain consistent across all eight scenes and trade some recall for a substantially lower false-free rate, which is desirable for a conservative overestimation of occupied space. VDB-GPDF makes the opposite trade-off: it reaches perfect recall ($100\%$) by predicting nearly all near-surface points as free, but its near-surface precision drops to $69$--$82\%$, giving it the highest false-free rate among all methods. This confirms that recall alone is misleading and that precision (false-free rate) must be reported to expose sign mistakes.

\begin{table*}
\centering
\caption{\small Sign metrics. Results are reported as the F1 score, precision, recall and accuracy for all points (All), points near the surface (Near) and points far from the surface (Far) on the Replica dataset. The best, second best and third best results are bold, underlined and curly-underlined, respectively.}
\label{table:sign_metrics}
\scriptsize
\resizebox{\linewidth}{!}{%

\def\signfonecell{\multirow{15}{*}{\begin{tabular}{@{}c@{}}F1 Score\\ $[\%] \uparrow$\end{tabular}}}
\def\signpreccell{\multirow{15}{*}{\begin{tabular}{@{}c@{}}Precision\\ $[\%] \uparrow$\end{tabular}}}
\def\signreccell{\multirow{15}{*}{\begin{tabular}{@{}c@{}}Recall\\ $[\%] \uparrow$\end{tabular}}}
\def\signacccell{\multirow{15}{*}{\begin{tabular}{@{}c@{}}Accuracy\\ $[\%] \uparrow$\end{tabular}}}
\def\allcell{\multirow{5}{*}{All}}
\def\nearcell{\multirow{5}{*}{Near}}
\def\farcell{\multirow{5}{*}{Far}}
\begin{tabular}{ll|c|cccccccc}
    \hline
    Metric           & Region    & Method   & room 0            & room 1            & room 2            & office 0          & office 1          & office 2          & office 3          & office 4          \\
    \hline
    \signfonecell    & \allcell  & Ours     & \textbf{98.25}    & \uwave{97.69}     & \underline{97.66} & \uwave{97.00}     & 96.52             & \uwave{97.92}     & \underline{97.79} & \underline{97.91} \\
                     &           & FIESTA   & \uwave{97.74}     & \uwave{97.69}     & \underline{97.66} & 96.74             & \uwave{96.53}     & \underline{98.21} & \textbf{97.97}    & \uwave{97.83}     \\
                     &           & Voxblox  & 96.77             & \underline{97.91} & \uwave{96.82}     & \underline{97.24} & \underline{97.35} & 95.33             & 94.97             & 95.46             \\
                     &           & iSDF     & \underline{97.93} & \textbf{98.43}    & \textbf{98.26}    & \textbf{98.60}    & \textbf{98.09}    & \textbf{98.34}    & \uwave{97.63}     & \textbf{98.14}    \\
                     &           & VDB-GPDF & 86.85             & 86.07             & 91.54             & 85.26             & 86.34             & 87.90             & 84.04             & 87.94             \\
    \hhline{|~|-|-|-|-|-|-|-|-|-|-|}
                     & \nearcell & Ours     & \textbf{94.51}    & 93.43             & \underline{93.26} & \uwave{91.71}     & 91.33             & \uwave{93.83}     & \underline{93.49} & \underline{93.26} \\
                     &           & FIESTA   & \uwave{93.02}     & \uwave{93.50}     & \uwave{93.20}     & 91.05             & \uwave{91.54}     & \underline{94.65} & \textbf{93.95}    & \uwave{93.06}     \\
                     &           & Voxblox  & 90.72             & \underline{94.22} & 91.38             & \underline{93.75} & \underline{94.31} & 88.96             & 86.34             & 87.29             \\
                     &           & iSDF     & \underline{93.76} & \textbf{95.76}    & \textbf{95.24}    & \textbf{96.29}    & \textbf{95.47}    & \textbf{95.34}    & \uwave{93.29}     & \textbf{94.41}    \\
                     &           & VDB-GPDF & 84.75             & 84.71             & 90.25             & 83.80             & 85.52             & 85.86             & 81.75             & 85.87             \\
    \hhline{|~|-|-|-|-|-|-|-|-|-|-|}
                     & \farcell  & Ours     & \textbf{100.00}   & \textbf{100.00}   & \textbf{100.00}   & \textbf{100.00}   & \textbf{100.00}   & \underline{99.98} & \underline{99.95} & \textbf{100.00}   \\
                     &           & FIESTA   & \textbf{100.00}   & \textbf{100.00}   & \textbf{100.00}   & \textbf{100.00}   & \textbf{100.00}   & \textbf{100.00}   & \textbf{100.00}   & \textbf{100.00}   \\
                     &           & Voxblox  & \uwave{99.58}     & \uwave{99.93}     & \uwave{99.75}     & \underline{99.29} & \uwave{99.49}     & 98.54             & 99.18             & \uwave{99.22}     \\
                     &           & iSDF     & \underline{99.99} & \underline{99.99} & \underline{99.99} & \textbf{100.00}   & \underline{99.99} & \uwave{99.97}     & \uwave{99.93}     & \underline{99.96} \\
                     &           & VDB-GPDF & \textbf{100.00}   & \textbf{100.00}   & \textbf{100.00}   & \textbf{100.00}   & \textbf{100.00}   & \textbf{100.00}   & \textbf{100.00}   & \textbf{100.00}   \\
    \hline
    \signpreccell    & \allcell  & Ours     & \underline{99.81} & \underline{99.81} & \textbf{99.90}    & \underline{99.68} & \textbf{99.64}    & \textbf{99.86}    & \textbf{99.85}    & \textbf{99.92}    \\
                     &           & FIESTA   & 97.85             & \uwave{98.72}     & \uwave{99.15}     & 98.45             & 97.27             & \uwave{98.85}     & 98.28             & 98.58             \\
                     &           & Voxblox  & \textbf{99.90}    & \textbf{99.84}    & \underline{99.74} & \textbf{99.74}    & \underline{99.61} & \underline{99.80} & \underline{99.84} & \underline{99.87} \\
                     &           & iSDF     & \uwave{98.73}     & 98.24             & 98.80             & \uwave{99.09}     & \uwave{98.42}     & 98.72             & \uwave{98.90}     & \uwave{99.75}     \\
                     &           & VDB-GPDF & 76.76             & 75.55             & 84.39             & 74.31             & 75.96             & 78.41             & 72.47             & 78.48             \\
    \hhline{|~|-|-|-|-|-|-|-|-|-|-|}
                     & \nearcell & Ours     & \underline{99.37} & \underline{99.44} & \textbf{99.69}    & \underline{99.06} & \textbf{99.05}    & \textbf{99.56}    & \textbf{99.55}    & \textbf{99.74}    \\
                     &           & FIESTA   & 93.35             & \uwave{96.32}     & \uwave{97.47}     & 95.62             & 93.26             & \uwave{96.54}     & 94.83             & 95.39             \\
                     &           & Voxblox  & \textbf{99.67}    & \textbf{99.54}    & \underline{99.22} & \textbf{99.28}    & \underline{99.03} & \underline{99.38} & \underline{99.47} & \underline{99.53} \\
                     &           & iSDF     & \uwave{96.08}     & 95.23             & 96.66             & \uwave{97.57}     & \uwave{96.22}     & 96.34             & \uwave{96.76}     & \uwave{99.22}     \\
                     &           & VDB-GPDF & 73.54             & 73.47             & 82.24             & 72.11             & 74.70             & 75.22             & 69.14             & 75.24             \\
    \hhline{|~|-|-|-|-|-|-|-|-|-|-|}
                     & \farcell  & Ours     & \textbf{100.00}   & \textbf{100.00}   & \textbf{100.00}   & \textbf{100.00}   & \textbf{100.00}   & \textbf{100.00}   & \textbf{100.00}   & \textbf{100.00}   \\
                     &           & FIESTA   & \textbf{100.00}   & \textbf{100.00}   & \textbf{100.00}   & \textbf{100.00}   & \textbf{100.00}   & \textbf{100.00}   & \textbf{100.00}   & \textbf{100.00}   \\
                     &           & Voxblox  & \textbf{100.00}   & \textbf{100.00}   & \textbf{100.00}   & \textbf{100.00}   & \textbf{100.00}   & \textbf{100.00}   & \textbf{100.00}   & \textbf{100.00}   \\
                     &           & iSDF     & \textbf{100.00}   & \textbf{100.00}   & \textbf{100.00}   & \textbf{100.00}   & \textbf{100.00}   & \textbf{100.00}   & \textbf{100.00}   & \textbf{100.00}   \\
                     &           & VDB-GPDF & \textbf{100.00}   & \textbf{100.00}   & \textbf{100.00}   & \textbf{100.00}   & \textbf{100.00}   & \textbf{100.00}   & \textbf{100.00}   & \textbf{100.00}   \\
    \hline
    \signreccell     & \allcell  & Ours     & 96.74             & 95.66             & 95.52             & 94.46             & 93.58             & 96.06             & 95.81             & 95.97             \\
                     &           & FIESTA   & \underline{97.63} & \uwave{96.69}     & \uwave{96.20}     & \uwave{95.08}     & \uwave{95.81}     & \uwave{97.57}     & \underline{97.67} & \underline{97.09} \\
                     &           & Voxblox  & 93.83             & 96.05             & 94.07             & 94.86             & 95.20             & 91.24             & 90.56             & 91.43             \\
                     &           & iSDF     & \uwave{97.15}     & \underline{98.63} & \underline{97.73} & \underline{98.12} & \underline{97.77} & \underline{97.96} & \uwave{96.39}     & \uwave{96.58}     \\
                     &           & VDB-GPDF & \textbf{100.00}   & \textbf{100.00}   & \textbf{100.00}   & \textbf{100.00}   & \textbf{100.00}   & \textbf{100.00}   & \textbf{100.00}   & \textbf{100.00}   \\
    \hhline{|~|-|-|-|-|-|-|-|-|-|-|}
                     & \nearcell & Ours     & 90.10             & 88.10             & 87.61             & 85.38             & 84.73             & 88.73             & 88.12             & 87.56             \\
                     &           & FIESTA   & \underline{92.70} & \uwave{90.85}     & \uwave{89.28}     & 86.89             & 89.89             & \uwave{92.84}     & \underline{93.09} & \underline{90.85} \\
                     &           & Voxblox  & 83.24             & 89.45             & 84.69             & \uwave{88.80}     & \uwave{90.02}     & 80.52             & 76.27             & 77.72             \\
                     &           & iSDF     & \uwave{91.54}     & \underline{96.31} & \underline{93.86} & \underline{95.05} & \underline{94.74} & \underline{94.36} & \uwave{90.06}     & \uwave{90.05}     \\
                     &           & VDB-GPDF & \textbf{100.00}   & \textbf{100.00}   & \textbf{100.00}   & \textbf{100.00}   & \textbf{100.00}   & \textbf{100.00}   & \textbf{100.00}   & \textbf{100.00}   \\
    \hhline{|~|-|-|-|-|-|-|-|-|-|-|}
                     & \farcell  & Ours     & \textbf{100.00}   & \textbf{100.00}   & \textbf{100.00}   & \textbf{100.00}   & \textbf{100.00}   & \underline{99.97} & \underline{99.89} & \textbf{100.00}   \\
                     &           & FIESTA   & \textbf{100.00}   & \textbf{100.00}   & \textbf{100.00}   & \textbf{100.00}   & \textbf{100.00}   & \textbf{100.00}   & \textbf{100.00}   & \textbf{100.00}   \\
                     &           & Voxblox  & \uwave{99.17}     & \uwave{99.87}     & \uwave{99.51}     & \underline{98.58} & \uwave{98.98}     & 97.13             & 98.38             & \uwave{98.46}     \\
                     &           & iSDF     & \underline{99.98} & \underline{99.98} & \underline{99.98} & \textbf{100.00}   & \underline{99.99} & \uwave{99.93}     & \uwave{99.86}     & \underline{99.92} \\
                     &           & VDB-GPDF & \textbf{100.00}   & \textbf{100.00}   & \textbf{100.00}   & \textbf{100.00}   & \textbf{100.00}   & \textbf{100.00}   & \textbf{100.00}   & \textbf{100.00}   \\
    \hline
    \signacccell     & \allcell  & Ours     & \textbf{97.07}    & 96.36             & \uwave{95.82}     & \uwave{95.17}     & \uwave{95.27}     & \uwave{96.47}     & \underline{96.70} & \underline{96.42} \\
                     &           & FIESTA   & \uwave{96.23}     & \uwave{96.37}     & \underline{95.86} & 94.77             & 95.26             & \underline{96.99} & \textbf{96.98}    & \uwave{96.31}     \\
                     &           & Voxblox  & 94.62             & \underline{96.68} & 94.31             & \underline{95.55} & \underline{96.37} & 92.16             & 92.61             & 92.24             \\
                     &           & iSDF     & \underline{96.47} & \textbf{97.47}    & \textbf{96.82}    & \textbf{97.70}    & \textbf{97.34}    & \textbf{97.09}    & \uwave{96.39}     & \textbf{96.73}    \\
                     &           & VDB-GPDF & 90.55             & 89.90             & 93.30             & 88.29             & 90.38             & 90.79             & 90.43             & 90.98             \\
    \hhline{|~|-|-|-|-|-|-|-|-|-|-|}
                     & \nearcell & Ours     & \textbf{92.13}    & \uwave{90.42}     & \underline{89.45} & \uwave{88.16}     & \uwave{87.57}     & \underline{91.01} & \underline{91.05} & \underline{90.32} \\
                     &           & FIESTA   & \uwave{88.24}     & 89.17             & \uwave{88.61}     & 85.27             & 85.62             & \uwave{90.81}     & \uwave{89.68}     & \uwave{88.61}     \\
                     &           & Voxblox  & 87.52             & \underline{92.00} & 86.75             & \underline{91.40} & \underline{91.95} & 84.85             & 83.34             & 82.78             \\
                     &           & iSDF     & \underline{91.06} & \textbf{93.78}    & \textbf{92.22}    & \textbf{94.69}    & \textbf{93.34}    & \textbf{93.01}    & \textbf{91.06}    & \textbf{91.89}    \\
                     &           & VDB-GPDF & 73.54             & 73.47             & 82.24             & 72.11             & 74.70             & 75.22             & 69.14             & 75.24             \\
    \hhline{|~|-|-|-|-|-|-|-|-|-|-|}
                     & \farcell  & Ours     & \textbf{100.00}   & \textbf{100.00}   & \textbf{100.00}   & \textbf{100.00}   & \textbf{100.00}   & \underline{99.97} & \underline{99.89} & \textbf{100.00}   \\
                     &           & FIESTA   & \textbf{100.00}   & \textbf{100.00}   & \textbf{100.00}   & \textbf{100.00}   & \textbf{100.00}   & \textbf{100.00}   & \textbf{100.00}   & \textbf{100.00}   \\
                     &           & Voxblox  & \uwave{99.17}     & \uwave{99.87}     & \uwave{99.51}     & \underline{98.58} & \uwave{98.98}     & 97.13             & 98.38             & \uwave{98.46}     \\
                     &           & iSDF     & \underline{99.98} & \underline{99.98} & \underline{99.98} & \textbf{100.00}   & \underline{99.99} & \uwave{99.93}     & \uwave{99.86}     & \underline{99.92} \\
                     &           & VDB-GPDF & \textbf{100.00}   & \textbf{100.00}   & \textbf{100.00}   & \textbf{100.00}   & \textbf{100.00}   & \textbf{100.00}   & \textbf{100.00}   & \textbf{100.00}   \\
    \hline
\end{tabular}
}
\end{table*}

\balance
{\small\putbib}
\end{bibunit}

\end{document}